\documentclass{article}

 \usepackage[preprint]{neurips_2026}

\usepackage[utf8]{inputenc} 
\usepackage[T1]{fontenc}    
\usepackage{hyperref}       
\usepackage{url}            
\usepackage{booktabs}       
\usepackage{amsfonts}       
\usepackage{nicefrac}       
\usepackage{microtype}      
\usepackage{xcolor}         

\usepackage{booktabs} 
\usepackage{kotex}
\usepackage{setspace}

\usepackage{rotating}

\usepackage{amsmath}
\usepackage{amssymb}
\usepackage{mathtools}
\usepackage{amsthm}
\usepackage[ruled,vlined]{algorithm2e}
\usepackage{thmtools,thm-restate}
\usepackage{cancel}
\usepackage{pifont}

\usepackage{algorithmic}
\usepackage{multicol,multirow}
\usepackage{subcaption}

\usepackage{float}
\usepackage{wrapfig}
\usepackage{thm-restate}
\usepackage{titletoc}

\usepackage{makecell}

\usepackage{graphicx}
\usepackage{svg}
\usepackage{soul}
\theoremstyle{plain}
\newtheorem{theorem}{Theorem}[section]
\newtheorem{proposition}[theorem]{Proposition}

\theoremstyle{definition}
\newtheorem{definition}[theorem]{Definition}

\theoremstyle{remark}
\newtheorem{remark}[theorem]{Remark}

\usepackage[capitalize,noabbrev]{cleveref}
\crefname{algocf}{Algorithm}{Algorithms}
\title{IPPRO: Importance-based Pruning with PRojective Offset for Magnitude-indifferent Structural Pruning}

\author{%
  Jaeheun Jung\thanks{equal contribution},\\
  Department of Mathematics\\
  Korea University\\
  Seoul, Republic of Korea \\
  \texttt{wodsos@korea.ac.kr} \\
  \And
  Jaehyuk Lee\footnotemark[1],\\
  Department of Mathematics\\
  Korea University\\
  Seoul, Republic of Korea \\
  \texttt{jaehyeokbear@korea.ac.kr} \\
  \AND
  Yeajin Lee,\\
  Program in Actuarial Science \\
  and Financial Engineering\\
  Korea University\\
  Seoul, Republic of Korea \\
  \texttt{gin1214@korea.ac.kr} \\  
  \And
  Donghun Lee\thanks{corresponding author}\\
  Department of Mathematics\\
  Korea University\\
  Seoul, Republic of Korea \\
  \texttt{holy@korea.ac.kr} \\
}

\begin{document}

\maketitle

\begin{abstract}
Importance-based structured pruning overwhelmingly relies on filter magnitude. This proxy is fundamentally flawed: due to scale invariance, functionally identical filters can receive arbitrarily different importance scores under rescaling. We propose IPPRO (Importance-based Pruning with PROjective Offset), a scale-invariant pruning framework grounded in projective geometry.
By embedding filters into real projective space ($\mathbb{RP}^N$), IPPRO resolves the singularity at the origin, placing all filters at an equal angular distance from the zero filter. We define PROscore, which captures functional importance by measuring a filter's angular displacement toward zero under a single gradient step (directional collapse). We further connect PROscore to exact $L_0$ relaxation, proving this one-shot criterion reliably predicts multi-step pruning dynamics.
Extensive experiments across CNNs, Vision Transformers, and LLMs (e.g., ResNet, DeiT, LLaMA) demonstrate that IPPRO consistently outperforms existing methods, yielding particularly striking gains under high compression and no-fine-tuning regimes, IPPRO establishes a robust, architecture-agnostic paradigm for neural network compression.
\end{abstract}

\section{Introduction}

Deep neural networks achieve remarkable performance, yet their growing scale demands practical compression. Structured pruning---removing entire filters---directly translates to hardware speedups and remains the most widely deployed paradigm. Among these, importance-based approaches dominate: they score each filter, prune the least important ones, and recover performance via fine-tuning.

The central question is deceptively simple: \emph{what makes a filter important?}

\paragraph{The magnitude assumption and its failure.}

\begin{figure}[t] 
    \centering
        \begin{minipage}{0.8\textwidth}
        \centering
        \begin{subfigure}{0.22\linewidth}
            \centering
            \includegraphics[width=\textwidth]{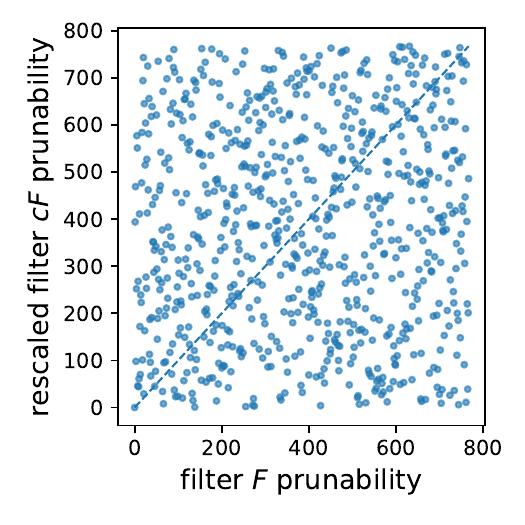}
            \caption{Magnitude}\label{fig:cf_l1}  
        \end{subfigure}
        \hfill
        \begin{subfigure}{0.22\linewidth}
            \centering
            \includegraphics[width=\textwidth]{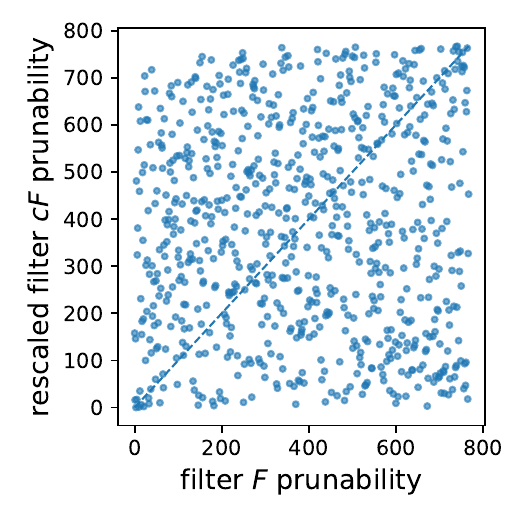}
            \caption{Taylor}\label{fig:cf_taylor}   
        \end{subfigure}
        \hfill
        \begin{subfigure}{0.22\linewidth}
            \centering
            \includegraphics[width=\textwidth]{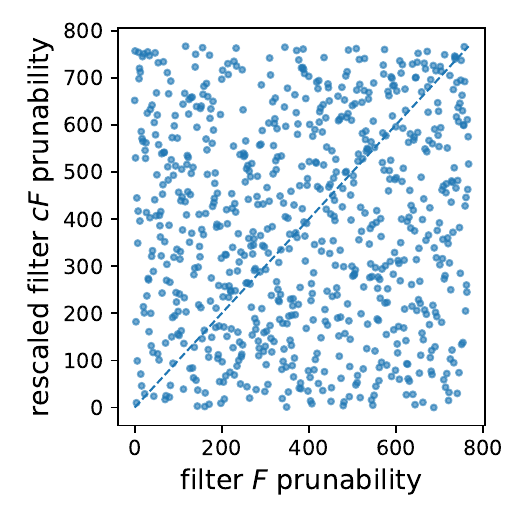}
            \caption{VBP}\label{fig:cf_vbp}  
        \end{subfigure}
        \hfill
        \begin{subfigure}{0.22\linewidth}
            \centering
            \includegraphics[width=\textwidth]{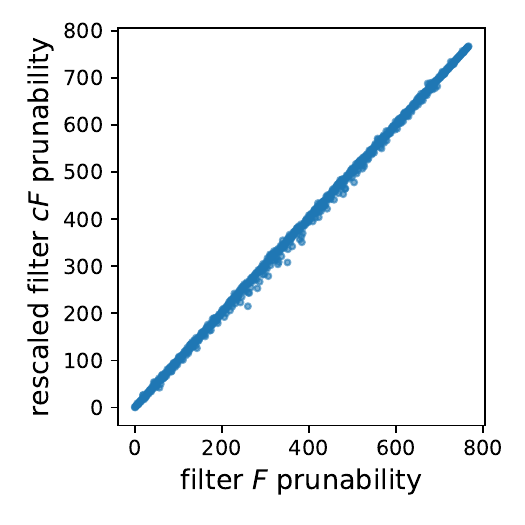}
            \caption{IPPRO (ours)}\label{fig:cf_ippro}   
        \end{subfigure}
        \end{minipage}
    \caption{Sensitivity of pruning criteria under filter rescaling on a 
DeiT-Tiny MLP block. Each plot compares the importance rank of 
$F$ against its rescaled counterpart $cF$ (subsequent layer compensated 
by $1/c$). }\label{fig:f_cf_compare}
\end{figure}

Magnitude-based pruning~\cite{filters2016pruning} remains the dominant paradigm~\cite{fang2024isomorphic, fang2023depgraph} due to its simplicity, yet importance scoring depends strongly on filter scale. For ReLU networks, rescaling a filter by 
$c>0$ and dividing the next layer by $1/c$ preserves the network function; magnitude-based criteria nonetheless assign such filters arbitrarily different scores (Figure~\ref{fig:cf_l1}).

\begin{figure*}[t]
    \centering
    \includegraphics[width=\linewidth, height=2.3cm]{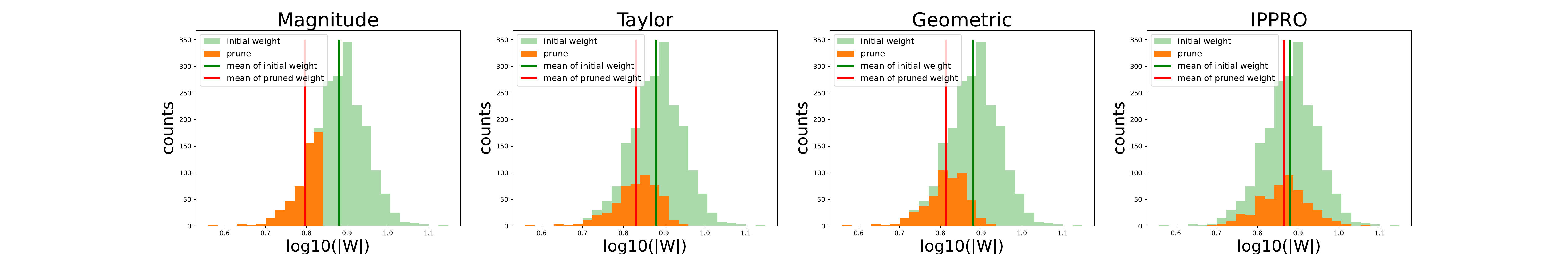}
    \caption{Visualization of magnitude of pruning filters obtained by four different criteria(Magnitude, Taylor, Geometric Median \cite{he2019filter}, IPPRO), on DeepLabV3-ResNet50.}\label{fig:main_pruning_index}
    \vspace{-4mm}
\end{figure*}

Gradient-based criteria such as Taylor~\cite{molchanov2019importance} $|F\cdot\nabla_F\mathcal{L}|$ are theoretically scale-invariant for ReLU networks. However, modern architectures employ non-homogeneous activations (GELU in ViTs, SiLU in LLMs), where exact equivalence no longer holds. Empirically, Taylor and variance-based (\cref{fig:cf_taylor,fig:cf_vbp}) criteria exhibit severe rank distortion; even advanced methods
preferentially prune small-magnitude filters (Figure~\ref{fig:main_pruning_index}), revealing a persistent ``size matters'' bias. A criterion that loosens this magnitude dependence and remains robust across architectures is desirable.

\paragraph{The geometric obstacle.}
A natural remedy---normalizing filters to compare their angular distance to the zero filter---encounters a geometric obstacle: normalization is \emph{ill-defined at the origin}. Distance metrics like cosine similarity collapse at this singularity. Ironically, structured pruning requires comparing non-zero filters to the zero filter---precisely where conventional geometry fails.

\paragraph{Our approach: projective geometry.}
We resolve this by reformulating pruning through \emph{projective geometry}. Projective space $\mathbb{RP}^N$ consists of equivalence classes of vectors under positive scaling, discarding magnitude by construction. Embedding filters into $\mathbb{RP}^N$ achieves: (i)~invariance to positive rescaling, (ii)~a well-defined zero filter at the ``point at infinity'' $[1:0:\cdots:0]$, and (iii)~a meaningful angular distance to zero. Here, all embedded filters reside at an equal angular distance ($\pi/4$) from the origin, ensuring pruning depends primarily on \emph{directional dynamics under gradient descent}.

\paragraph{Contributions.}
Based on this perspective, we make the following contributions:
\begin{enumerate}
    \item We introduce \textbf{PROscore}, a novel importance criterion defined as the angular displacement toward the zero filter via one gradient step in projective space. The embedding is exactly scale-invariant by construction, and PROscore preserves filter rankings robustly across realistic reparameterizations, even with non-homogeneous activations.
    \item We establish a \textbf{formal connection to exact $L_0$ relaxation}~(\cref{sec:l0_connection}), showing PROscore captures the initial bifurcation dynamics of the Catalyst framework~\cite{catalyst}.
    \item We propose \textbf{IPPRO}, a practical, architecture- and task-agnostic pruning algorithm.
    \item \textbf{Extensive experiments} across CNNs, ViTs, and LLMs demonstrate IPPRO consistently outperforms state-of-the-art methods, with striking advantages under high compression and no-fine-tuning settings.
\end{enumerate}
We present the theoretical foundations in \cref{sec:methods}, the IPPRO algorithm in \cref{sec:algorithm}, and empirical evaluations in \cref{sec:experiment}.

\section{Related Work}
\subsection{Importance-Based Structured Pruning}

Deep Neural Networks (DNNs) are often overparameterized, and pruning mitigates the resulting computational costs by removing redundant filters based on their estimated network contribution. 

Early approaches primarily rely on filter magnitude (e.g., $L_1$- or $L_2$-norm~\cite{han2015deep, he2018soft}), implicitly assuming that larger filters are more important. While gradient-based methods~\cite{blalock2020state, molchanov2019importance} improve upon this by evaluating loss sensitivity, and other works exploit statistical redundancy~\cite{he2019filter, wang2019cop} or structural metrics~\cite{wang2021convolutional, 10484493}, most criteria remain implicitly entangled with magnitude information (as shown in \cref{fig:main_pruning_index}). This entanglement makes it difficult to decouple a filter's true functional importance from its parameter scale. 

In contrast, IPPRO departs from both magnitude- and statistics-based paradigms by formulating pruning in a projective geometric space, deriving a provably scale-invariant importance score directly from directional gradient flow.

\subsection{Structured Pruning for Transformer Architectures}

\begin{figure*}[t]
    \centering
    \begin{subfigure}[b]{0.36\textwidth}
        \includegraphics[width=\textwidth]{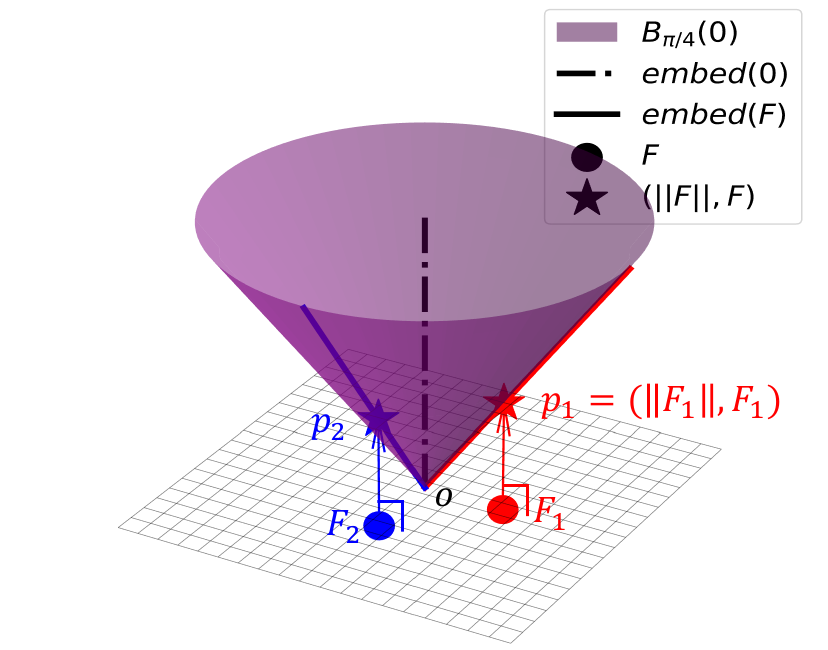}
        \caption{Visualization of $B_{\pi/4}(0)$ and $embed(F)$ in $\mathbb{R}^{N+1}$}
        \label{fig:concept_embed}
    \end{subfigure}
    \begin{subfigure}[b]{0.48\textwidth}
        \includegraphics[width=\textwidth]{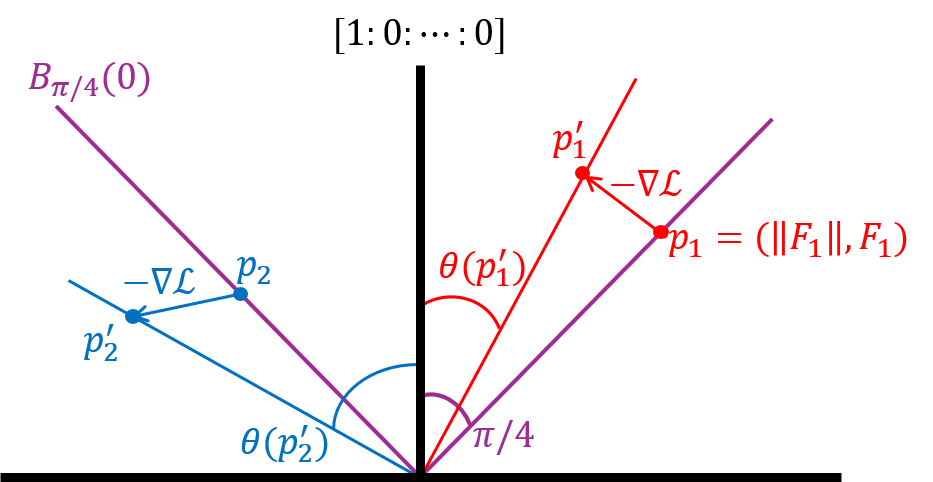}
        \begin{minipage}{.1cm}
            \vfill
            \end{minipage}
        \caption{Intuitive section plot for $\theta(p)$ explanation}
        \label{fig:concept_theta}
    \end{subfigure}
    \caption{Illustrations for conceptual understanding of PROscore for $N$-dimensional filter $F$.}
    \label{fig:intuitive_projective}
    \vspace{-4mm}
\end{figure*}

Transformer pruning poses unique challenges due to the coupled structure of multi-head self-attention (MHA), where pruning decisions must be coordinated across related components. Prior work has mainly explored two directions: \emph{head pruning}, which removes entire attention heads to reduce redundancy \cite{michel2019sixteen,yu2022width,yu2023x}, and \emph{neuron-level pruning}, which reduces the dimensionality within each head by pruning individual neurons or projection components \cite{shim2024snp,yu2022unified,zhu2021visiontransformerpruning}. Some approaches further extend pruning to embedding dimensions or entire attention blocks \cite{yang2023global,fang2024isomorphic,yu2022unified}.

Although IPPRO can be applied to both head- and neuron-level pruning by grouping related neurons, this work focuses on neuron-level pruning, which generalizes CNN filter pruning and provides finer granularity, making it well suited for scale-invariant importance scoring such as PROscore.

\section{Methods}\label{sec:methods}

Let $C$ be the number of prunable channels in a neural network model, and $F_1,\cdots,F_C\in \mathbb{R}^N$ be filters which correspond to the channels.
We aim to design a novel pruning methodology that compares each filter with the zero filter while treating functionally equivalent (positively rescaled) filters identically.
To address this, in \cref{subsec:method-projective_space}, we introduce the concept of projective geometry, and define the mapping function $embed$ which maps the filters to the projective space $\mathbb{RP}^N$. 
After mapping the filters into $\mathbb{RP}^N$, in \cref{subsec:method-importance_score} we explain the distance in projective space given by an angular distance, and define the proposed importance score using the angular distance. Finally, in \cref{sec:l0_connection}, we establish a connection between PROscore and exact $L_0$ relaxation.

\subsection{Embedding Filters into Projective Space}\label{subsec:method-projective_space}

To isolate scale-invariant directional information and enable comparison with the zero filter, we map each filter into real projective space via a homogenization embedding.

\paragraph{Projective space.}
The real projective space $\mathbb{RP}^N$ is defined as the quotient of $\mathbb{R}^{N+1} \setminus \{0\}$ under the equivalence relation $x \sim cx$ for all nonzero $c \in \mathbb{R}$, comprising lines through the origin in $\mathbb{R}^{N+1}$. An element of $\mathbb{RP}^N$ is expressed in \emph{homogeneous coordinates} as $[v] = [v_0 : \cdots : v_N]$, satisfying
\begin{equation}
   \forall\, c \neq 0,\quad [v_0 : \cdots : v_N] = [cv_0 : \cdots : cv_N].
\end{equation}

\paragraph{Embedding map.}
For a nonzero filter $F_i = (F_{i1}, \ldots, F_{iN})$, we define the embedding
\begin{equation}\label{eq:embed}
    \mathrm{embed} \colon (F_{i1}, \ldots, F_{iN}) \mapsto [\|F_i\| : F_{i1} : \cdots : F_{iN}].
\end{equation}

This embedding provides two critical properties:

\emph{Scale invariance.}~For any $c > 0$, $\mathrm{embed}(cF) = \mathrm{embed}(F)$ holds by construction, since $[\|cF\| : cF] = [c\|F\| : cF] = [\|F\| : F]$. Functionally equivalent filters are mapped to the \emph{same point} in $\mathbb{RP}^N$.

\emph{Well-defined origin.}~The zero filter receives the projective representative $[1 : 0 : \cdots : 0]$, the point at infinity. This resolves the singularity of filter normalization at the origin and enables direct geometric comparison between any filter and zero via angular distance. As illustrated in \cref{fig:concept_embed} for $N = 2$, the zero filter corresponds to the central axis $[1 : 0 : \cdots : 0]$.

The embedding~\eqref{eq:embed} can be understood as a \emph{homogenization} of the normalized filter: by lifting $F \in \mathbb{R}^N$ to $[\|F\| : F] \in \mathbb{RP}^N$, scale and direction are explicitly separated, yielding positive-rescaling invariance while preserving directional information.

\subsection{PROscore: Angular Importance via Directional Collapse}\label{subsec:method-importance_score}

In $\mathbb{RP}^N$, the natural distance between two points is the angle between the corresponding lines in $\mathbb{R}^{N+1}$. Let $B_r(0)$ denote the set of points in $\mathbb{RP}^N$ at angular distance $r$ from the origin $[1:0:\cdots:0]$. A key observation is that $\mathrm{Im}(\mathrm{embed}) \subseteq B_{\pi/4}(0)$: \emph{every embedded filter lies at angular distance exactly $\pi/4$ from the origin}. This is visualized in \cref{fig:concept_embed}, where $B_{\pi/4}(0)$ forms a cone and all embedded filters reside on this cone.

This geometric setup is the foundation of our approach. By placing all filters at equal angular distance from the origin, pruning decisions are determined \emph{not} by initial scale but by subsequent directional behavior under gradient descent.

\paragraph{One-step angular displacement.}
After embedding, we evaluate each filter's importance by estimating whether a gradient step moves it \emph{toward} the origin (indicating the filter is collapsing to zero and should be pruned) or \emph{away} from it (indicating the filter carries essential information). As depicted in \cref{fig:concept_theta}, if gradient descent moves $p_i = \mathrm{embed}(F_i)$ to $p_i'$, then the angular distance $\theta(p_i')$ from $p_i'$ to the origin quantifies how much the filter resists collapse.

\paragraph{Definition of PROscore.}
Given filter $F_i$, let $D_i$ be an auxiliary variable initialized to $\|F_i\|$. The projective point $p_i = [\|F_i\| : F_i]$ is updated to
\begin{equation}\label{eq:point}
    p_i' = \left[\|F_i\| - \lambda\frac{\partial \mathcal{L}}{\partial D_i} \;:\; F_i - \lambda\nabla_{F_i}\mathcal{L}\right],
\end{equation}
where $\lambda$ is a step-size parameter and $\mathcal{L}$ is the task loss. The \textbf{PROscore} is defined as
\begin{equation}\label{eq:proscore}
    \mathrm{PROscore}_\lambda(i) \;:=\; \tan(\theta(p_i')) \;=\; \frac{\|F_i - \lambda\nabla_{F_i}\mathcal{L}\|}{\left|D_i - \lambda\frac{\partial \mathcal{L}}{\partial D_i}\right|},
\end{equation}
which serves as the importance score for the $i$-th filter. A small PROscore indicates directional collapse toward the zero filter (the filter should be pruned); a large PROscore indicates functional importance (the filter should be preserved).

\paragraph{Scale invariance of PROscore.}
The coupled scaling of $F_i$ and $D_i$ structurally cancels positive rescaling in the ratio of Equation~\eqref{eq:proscore}, yielding exact invariance for ReLU networks. 
For non-homogeneous activations (GELU, SiLU), exact equivalence no longer holds, but PROscore's ratio structure empirically preserves rankings (Figure~\ref{fig:f_cf_compare}, Appendix~\ref{appendix:stability}).

\subsection{Theoretical Justification: Connection to Exact $L_0$ Relaxation.}\label{sec:l0_connection}

The one-shot design of PROscore is not merely a geometric heuristic, but is formally grounded in the exact $L_0$ relaxation framework Catalyst~\cite{catalyst}. By embedding filters at an initial angular distance of $\pi/4$ in $\mathbb{RP}^N$, IPPRO places them precisely on the critical bifurcation boundary ($c=1$) identified in Catalyst. Because optimization dynamics at this boundary are exponentially divergent---filters either rapidly collapse toward zero or are preserved---a single initial gradient step is sufficient to reliably predict the asymptotic multi-step pruning outcome. We provide the formal derivation mapping PROscore to this exact relaxation framework in \cref{appendix:l0_connection}.

\section{The IPPRO Algorithm}\label{sec:algorithm}

We now describe \textbf{IPPRO} (Importance-based Pruning with PRojective Offset), our implementation of structured pruning using PROscore. We first present the parameter injection mechanism that realizes the projective embedding, then give the scoring algorithm, and finally discuss adaptations for transformer architectures.

\subsection{Projective Offsetting via Parameter Injection}\label{subsec:method-implementation}

Computing PROscore (\cref{eq:proscore}) requires gradients with respect to both the filter $F$ and the homogeneous coordinate $\|F\|$. While gradients with respect to $F$ are standard, $\|F\|$ is not an explicit parameter. We therefore introduce an auxiliary parameter $D$ representing the homogeneous coordinate and inject it into the model using the extension method of~\cite{catalyst}. Specifically, we modify the element-wise operation $\sigma$ following the target layer as
\begin{equation}\label{eqn:psi_activation}
    \psi_{D,\overline{D}}(x) = Dx - \overline{D}x + \sigma(x),
\end{equation}
with $D = \overline{D}$ at initialization. Since the additional terms cancel, the forward pass is \emph{exactly preserved}---FLOPs and the parameter increase is less than 0.2\%. During backpropagation, gradients flow independently through $D$ and $F$, enabling computation of $\frac{\partial \mathcal{L}}{\partial D_i}$. The injected parameters are used \emph{only} for importance estimation and are removed before pruning. A detailed schematic and analysis of the associated computational overhead are provided in \cref{app:injection_trick_computational} and \cref{app:injection_trick}.

\subsection{PROscore Computation}

\begin{algorithm}[H]
    \caption{PROscore Computation}
    \begin{algorithmic}[1]
        \STATE \textbf{Input:} number of channels $C$, dataset $\mathcal{D}$, layer $W$, consecutive operation $\sigma$, model $f(W,\sigma)$
        \STATE Compute $D^{\text{init}} = \mathrm{diag}(\|F_1\|, \ldots, \|F_C\|)$
        \STATE Extend model to $f(W, \psi_{D,\overline{D}})$ using \cref{eqn:psi_activation}
        \STATE Initialize $D, \overline{D} \leftarrow D^{\text{init}}$
        \STATE Compute $\nabla\mathcal{L}$ via backpropagation on $\mathcal{D}$
        \STATE Compute $\mathrm{PROscore}_\lambda(i) = \tan(\theta(p_i'))$ for all $i$ via \cref{eq:proscore}
        \STATE \textbf{Return} $\mathrm{PROscore}_\lambda(i)$ for each filter $F_i$
    \end{algorithmic}\label{alg:IPPRO}
\end{algorithm}

As detailed in \cref{alg:IPPRO}, IPPRO computes PROscore \emph{without updating model parameters}: gradients are accumulated across the dataset, and pruning is applied to the original model using the precomputed scores. The choice of target operation $\sigma$ depends on the architecture (e.g., ReLU for CNNs, or an auxiliary identity for transformers).

\subsection{Adaptation for Multi-Head Attention}\label{subsec:attn_pruning}
Three practical modifications enable PROscore on attention layers. 
\emph{(i)} Since MHA lacks an internal activation, we insert an auxiliary 
identity after the QKV projection and extend it via $\psi_D(x) = Dx + x$, 
with $D$ initialized so that filters lie on $B_{\pi/4}(0)$ without 
altering the Q, K, V tensors (\cref{app:mha_adaptation}). 
\emph{(ii)} We group neurons and average PROscores within each group 
for GPU-friendly parallelism. 
\emph{(iii)} Query and key dimensions are pruned at equal ratios to 
preserve dimensional compatibility.

\section{Empirical Validations}\label{sec:experiment}

We evaluate IPPRO across diverse tasks and architectures, including CNNs and ViTs for image classification, DeepLabV3-R50 for semantic segmentation, and LLaMA/LLaMA2-7B for LLMs.

A key practical advantage of IPPRO is its robustness across two axes: (i) the scaling parameter $\lambda$, and (ii) network reparameterization. For (i), we empirically found PROscore's ranking to be invariant to $\lambda$'s value, so we fix $\lambda$ = 1 for all experiments. For (ii), IPPRO maintains stable accuracy across normalization schemes (BatchNorm, LayerNorm, Identity) where magnitude-based 
pruning shows large variance (Appendix~\ref{appendix:stability}).

We benchmark IPPRO against over 30 state-of-the-art pruning methods, including Taylor \cite{molchanov2019importance}, SNP \cite{shim2024snp}, SIRFP \cite{lv2024fgp}, and LLM-Pruner \cite{ma2023llm}. Across all benchmarks, IPPRO consistently achieves superior or highly competitive performance without architecture-specific tuning. Detailed implementation setups and fine-tuning protocols are provided in \cref{sec:implementation,sec:finetune_recipe}.

\begin{table}[t]
    \centering
    
    \begin{minipage}[c]{0.45\linewidth}
        \centering
        \caption{Cityscapes - DeepLabV3-ResNet50}\label{tab:seg_results}
    \resizebox{\linewidth}{!}{
    \begin{tabular}{lccccc}
        \toprule
        \multirow{2.5}{*}{Method} & \multicolumn{3}{c}{mIoU} & \multirow{2.5}{*}{Params$\downarrow$(\%)} & \multirow{2.5}{*}{FLOPs$\downarrow$(\%)} \\
        \cmidrule(lr){2-4}
        & Base. & Prun. & $\Delta$ \\
        \midrule
        Taylor\cite{molchanov2019importance} & 81.6 & 80.3 & -1.3 & 63.7 & 60.1 \\
        DepGraph\cite{fang2023depgraph} & 81.6 & 80.0 & -1.6 & 59.2 & 60.4 \\
        FPGM\cite{he2019filter} & 81.6 & 80.2 & -1.4 & 63.9 & 61.2 \\
        DCFP\cite{wang2023dcfp} & 81.6 & 80.9 & -0.7 & 64.2 & 60.9 \\
        FGP\cite{lv2024fgp} & 79.3 & 79.0 & \underline{-0.3} & \underline{64.4} & 60.4 \\
        SIRFP\cite{wu2025structural} & 81.6 & \underline{81.3} & \underline{-0.3} & \textbf{64.8} & \underline{61.3} \\
        \textbf{IPPRO (ours)} & 81.5 & \textbf{81.5} & \textbf{0.0} & 64.0 & \textbf{61.8} \\
        \midrule
        FPGM\cite{he2019filter} & 81.6 & 79.3 & -2.3 & \textbf{74.5} & 71.0 \\
        DCFP\cite{wang2023dcfp} & 81.6 & 80.2  & -1.4 &  74.3 & 71.3 \\
        SIRFP\cite{wu2025structural} & 81.6 & \underline{80.9} & \underline{-0.7} & \underline{74.4} & \underline{71.9} \\
        \textbf{IPPRO (ours)} & 81.5 & \textbf{81.2} & \textbf{-0.3} & \textbf{74.5} & \textbf{72.2} \\
        \midrule
        FPGM\cite{he2019filter} & 81.6 & 77.9 & -3.7 &  84.8 &  80.7 \\
        DCFP\cite{wang2023dcfp} & 81.6 &  \underline{79.5} &  \underline{-2.1} & 83.9 & 80.2 \\
        SIRFP\cite{wu2025structural} & 81.6 & 79.4 & -2.2 & \textbf{85.7} & \underline{81.6} \\
        \textbf{IPPRO (ours)} & 81.5 & \textbf{80.1} & \textbf{-1.4} & \underline{84.5} & \textbf{81.7} \\
        \bottomrule
    \end{tabular}
    }
    
    \end{minipage}
    \hfill 
    \begin{minipage}[c]{0.5\linewidth}
        \centering
        \caption{DeiT and EfficientFormer pruning performance on ImageNet-1K}\label{tab:vit_results}
        \resizebox{\linewidth}{!}{
            \begin{tabular}{llccccc}
            \toprule
            \multirow{2.5}{*}{Model} & \multirow{2.5}{*}{Method}  & \multicolumn{3}{c}{Acc($\uparrow$\%)} & \multirow{2.5}{*}{Params$\downarrow$(\%)} & \multirow{2.5}{*}{FLOPs$\downarrow$(\%)} \\
            \cmidrule(lr){3-5}
            & & Base. & Prun. & $\Delta$ \\
            \midrule
            \multirow{5}{*}{DeiT-Base}
            & WDPruning \cite{yu2022width} & 81.80 & 80.76 & -1.04 & 36.1 & 43.7 \\
            & X-Pruner \cite{yu2023x} & 81.80 & \underline{81.02} & \underline{-0.78} & - & 51.7 \\
            & UVC \cite{yu2022unified} & 81.80 & 80.57 & -1.23 & - & \underline{54.5} \\
            & SNP \cite{shim2024snp} & 81.80 & 79.63 & -2.17 & \underline{63.5} & \textbf{63.6} \\
            & \textbf{IPPRO (ours)} & 81.80 & \textbf{81.14} & \textbf{-0.66} & \textbf{63.7} & \textbf{63.6} \\
            \midrule
            \multirow{5}{*}{DeiT-Small}
            & WDPruning \cite{yu2022width} & 79.85 & 78.38 & -1.47 & 39.8 & 43.7 \\
            & X-Pruner \cite{yu2023x} & 79.85 & \underline{78.93} & \underline{-0.92} & - & 47.8 \\
            & UVC \cite{yu2022unified} & 79.85 & 78.82 & -1.03 & - & \underline{50.0} \\
            & SNP \cite{shim2024snp} & 79.85 & 78.52 & -1.33 & \underline{54.7} & \textbf{56.5} \\
            & \textbf{IPPRO (ours)} & 79.85 & \textbf{79.13} & \textbf{-0.72} & \textbf{55.4} & \textbf{56.5} \\
            \midrule
            \multirow{5}{*}{DeiT-Tiny}
            & SSViTE \cite{chen2021chasing} & 72.20 & 70.12 & -2.08 & 26.3 & 30.7 \\
            & WDPruning \cite{yu2022width} & 72.20 & 70.34 & -1.86 & 38.5 & \underline{46.1} \\
            & X-Pruner \cite{yu2023x} & 72.20 & 71.10 & -1.10 & - & \textbf{53.8} \\
            & UVC \cite{yu2022unified}  & 72.20 & \underline{71.30} & \underline{-0.90} & - & \textbf{53.8} \\
            & SNP  \cite{shim2024snp} & 72.20 & 70.29 & -1.91 & \textbf{47.3} & \textbf{53.8} \\
            & \textbf{IPPRO (ours)} & 72.20 & \textbf{71.67} & \textbf{-0.53} & \underline{43.3} & \textbf{53.8} \\
            \midrule
            \multirow{2}{*}{EfficientFormer-L1}
            & SNP  \cite{shim2024snp} & 79.20 & \underline{75.53} & \underline{-3.67} & - & \textbf{53.8} \\
            & \textbf{IPPRO (ours)} & 79.20 & \textbf{77.08} & \textbf{-2.12} & 48.8 & \textbf{53.8} \\
        \bottomrule
    \end{tabular}
    }
    \end{minipage}
\end{table}

\begin{figure*}[t]
    \begin{subfigure}{\linewidth}
        \includegraphics[width=\textwidth, height=1.5cm]{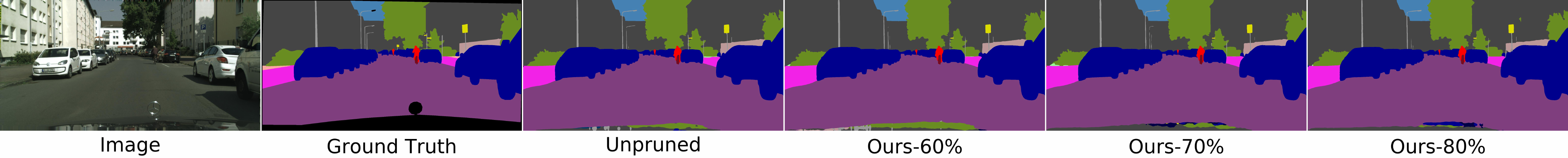}
    \end{subfigure}
    \caption{Qualitative results of IPPRO on the segmentation task: Cityscapes dataset using DeepLabV3. From left to right: input image, ground truth, unpruned model, and IPPRO-pruned models with 60\%, 70\%, and 80\% sparsity, respectively.}\label{fig:seg_results_paper}
\end{figure*}

\subsection{Performance Analysis of IPPRO on CNNs}\label{subsec:cnn_performance}

We first evaluate IPPRO on CNN-based semantic segmentation tasks using DeepLabV3-ResNet50 \cite{chen2017deeplab} on the Cityscapes dataset \cite{cordts2016cityscapes}, following the setup of DCFP \cite{wang2023dcfp}. 

While SIRFP and DCFP are segmentation-specific, IPPRO is task and architecture-agnostic, yet achieves comparable or superior results across pruning levels.
As shown in \cref{tab:seg_results}, IPPRO preserves original mIoU levels even at 60\% parameter reduction and consistently outperforms or matches other methods across pruning levels. Qualitative results (\cref{fig:seg_results_paper,fig:seg_results}) show that semantic structure and object boundaries remain intact, even under high sparsity.

We further validate IPPRO on ResNet-50/MobileNet (ImageNet-1k \cite{deng2009imagenet}) and VGG19/ResNet-56 (CIFAR-10/100 \cite{krizhevsky2009cifar}). Across all settings, IPPRO consistently yields robust compression–accuracy trade-offs. See \cref{appendix:additional_performace} for detailed analyses.

\subsection{Performance Analysis of IPPRO on ViTs}\label{subsec:vit_performance}

To demonstrate the generalizability of IPPRO beyond CNNs, we benchmark it on Transformer-based vision models, including DeiT (Base/Small/Tiny) \cite{touvron2021training}, EfficientFormer-L1 \cite{li2022efficientformer}, and Swin-T \cite{liu2021swin}. As summarized in Table \ref{tab:vit_results}, IPPRO consistently achieves the smallest or highly competitive accuracy drop under comparable FLOPs reductions across all evaluated models. For example, on DeiT-Base, IPPRO reduces FLOPs by 63.6\% with only a 0.66\% accuracy drop, outperforming SNP, which incurs a 2.17\% drop at a similar compression level. Similar trends hold across other architectures.

\begin{table}[t]
    \centering
    
    \begin{minipage}[c]{0.45\linewidth}
        \centering
    \caption{Comparison with VBP under limited fine-tuning budget. pre and post-FT denote Top-1 accuracy before and after fine-tuning, respectively.}
    \resizebox{\linewidth}{!}{
        \label{tab:comparison_vbp}
        \begin{tabular}{lcccc}
        \toprule
        Model & MACs (G) & Param. (M) & \multicolumn{2}{c}{Acc(\%)}  \\
        \cmidrule(lr){4-5}
         &  &  & pre-FT & post-FT \\
        \midrule
        DeiT-T          & 1.26 & 5.72  & --    & 72.02 \\
        Magnitude       & 0.91 & 3.94  & 3.56  & 68.49 \\
        VBP   \cite{berisha2025variance}          & 0.91 & 3.94  & 39.58 & 70.61 \\
        \textbf{IPPRO (ours)}    & 0.91 & 3.87  & \textbf{44.34} & \textbf{71.49} \\
        \midrule
        DeiT-S          & 4.61 & 22.05 & --    & 79.70 \\
        Magnitude       & 3.21 & 14.96 & 4.05  & 76.55 \\
        VBP  \cite{berisha2025variance}           & 3.21 & 14.96 & 64.44 & 78.62 \\
        \textbf{IPPRO (ours)}    & 3.20 & 14.92 & \textbf{67.13} & \textbf{78.88} \\
        \midrule
        Swin-T          & 4.50 & 28.29 & --    & 80.91 \\
        Magnitude       & 3.23 & 21.31 & 34.95  & 75.94 \\
        VBP \cite{berisha2025variance}            & 3.23 & 21.31 & 55.12 & 79.41 \\
        \textbf{IPPRO (ours)}    & 3.22 & 21.37 & \textbf{57.28} & \textbf{80.13} \\
        \bottomrule
        \end{tabular}
     }
    \end{minipage}
    \hfill 
    \begin{minipage}[c]{0.52\linewidth}
        \centering
        \caption{LLM Pruning results on different calibration size}\label{tab:data_sensitivity}
        \resizebox{0.8\linewidth}{!}{
            \begin{tabular}{cccc}
            \toprule
            \ Num samples & Method & Pruning ratio (\%) & Avg($\uparrow$\%) \\
             \midrule
            \multirow{1}{*}{NA}
            & Baseline & 0 & 63.17 \\
            \midrule
            \multirow{2}{*}{10}
            & LLM-Pruner & 20 & 60.06 \\
            & IPPRO (ours) & 20 & \underline{60.94} \\
            \midrule
            \multirow{1}{*}{100}
            & IPPRO (ours) & 20 & 60.59 \\
            \midrule
            \multirow{1}{*}{1000}
            & IPPRO (ours) & 20 & \textbf{60.98} \\
            \bottomrule
            \end{tabular}
        }

        \centering
        \caption{Latency comparison of ViTs}\label{tab:vit_latency}
        \resizebox{\linewidth}{!}{
            \begin{tabular}{lccccc}
            \toprule
            Model & Latency(s) & Params(M) & MACs(G) & Acc & $\Delta$ Acc \\
            \midrule
            DeiT-Tiny & 15.89(1.0x) & 5.72 & 1.26 & 72.20 & 0 \\
            + IPPRO (Ours) & 9.93(1.6x) & 3.3 & 0.6 & 71.67 & -0.53 \\
            \midrule
            DeiT-Small & 39.63(1.0x) & 22.05 & 4.61 & 79.85 & 0 \\
            + IPPRO (Ours) & 29.34(1.35x) & 9.84 & 2.0 & 79.13 & -0.72 \\
            \bottomrule
            \end{tabular}
        }
    \end{minipage}
\end{table}

\begin{table*}[t]
    \small
    \centering
    \caption{LLAMA-7b pruning performance comparison of various pruning method on five common reasoning datasets }\label{tab:llm_results_llama7b}
    \resizebox{0.9\linewidth}{!}{
        \begin{tabular}{llccccccc}
        \toprule
        \multirow{2.5}{*}{\makecell{Remain Param \\ Ratio}} & \multirow{2.5}{*}{Method} & \multicolumn{5}{c}{Acc($\uparrow$\%)} & \multirow{2.5}{*}{Avg($\uparrow$\%)} & \multirow{2.5}{*}{Drop($\downarrow$\%)} \\
         \cmidrule(lr){3-7}
        & & BoolQ & PIQA & ARC-e & ARC-c & OBQA \\
        \midrule
        \multirow{1}{*}{1.0}
        & Baseline & 0.73 & 0.78 & 0.67 & 0.41 & 0.42 & 0.602 & 0.0 \\
        \midrule
        \multirow{4}{*}{0.8}
        & LLM-Pruner \cite{ma2023llm} & \underline{0.69} & \underline{0.76} & 0.63 & 0.37 & \underline{0.40} & 0.570 & 3.0 \\
        & FLAP \cite{an2024fluctuation} & \underline{0.69} & \underline{0.76} & \textbf{0.69} & \textbf{0.39} & 0.39 & 0.584 & 1.6 \\
        & Numerical \cite{shen2025numerical} & 0.68 & 0.75 & 0.59 & 0.37 & 0.39 & 0.556 & 4.4 \\
        & Dobi-SVD \cite{wang2025dobi} & \textbf{0.73} & \textbf{0.77} & 0.65 & 0.37 & \textbf{0.42} & \underline{0.588} & \underline{1.2} \\
        & \textbf{IPPRO (ours)} & \textbf{0.73} & \textbf{0.77} & \underline{0.67} & \underline{0.38} & \underline{0.40} & \textbf{0.590} & \textbf{1.0} \\
        \bottomrule
        \end{tabular}
    }
\end{table*}

\begin{table*}[t]
    \small
    \centering
    \caption{LLAMA2-7b pruning performance comparison of various pruning method on five common reasoning datasets}\label{tab:llm_results_llama2_7b}
    \resizebox{0.9\linewidth}{!}{
        \begin{tabular}{llccccccc}
        \toprule
        \multirow{2.5}{*}{\makecell{Remain Param \\ Ratio}} & \multirow{2.5}{*}{Method} & \multicolumn{5}{c}{Acc($\uparrow$\%)} & \multirow{2.5}{*}{Avg($\uparrow$\%)} & \multirow{2.5}{*}{Drop($\downarrow$\%)} \\
         \cmidrule(lr){3-7}
        & & PIQA & HellaSwag & WinoGrande & ARC-e & ARC-c \\
        \midrule
        \multirow{1}{*}{1.0}
        & Baseline & 0.78 & 0.57 & 0.69 & 0.76 & 0.43 & 0.646 & 0.0 \\
        \midrule
        \multirow{5}{*}{0.5}
        & LLM-Pruner \cite{ma2023llm} & \underline{0.67} & 0.35 & 0.52 & 0.48 & 0.22 & 0.448 & 19.8 \\
        & SliceGPT \cite{ashkboos2024slicegpt} & 0.58 & \textbf{0.46} & \textbf{0.55} & 0.37 & \textbf{0.28} & 0.448 & 19.8 \\
        & Bonsai \cite{dery2024everybody} & 0.66 & 0.40 & \underline{0.54} & \underline{0.49} & \underline{0.26} & \underline{0.470} & \underline{17.6} \\
        & Wanda-sp \cite{sun2023simple} & 0.63 & 0.32 & 0.53 & 0.43 & 0.20 & 0.422 & 22.4 \\
        & Bypass \cite{gao2025bypass} & \underline{0.67} & 0.36 & 0.52 & \textbf{0.50} & 0.24 & 0.458 & 18.8 \\ 
        & \textbf{IPPRO (ours)} & \textbf{0.68} & \underline{0.43} & 0.51 & \textbf{0.50} & \textbf{0.28} & \textbf{0.480} & \textbf{16.6} \\
        \bottomrule
        \end{tabular}
    }
    \vspace{-4mm}
\end{table*}

While fine-tuning typically remains essential for restoring accuracy, performance immediately after pruning provides a useful diagnostic of criterion quality, since it isolates the pruning decision from the recovery procedure. To this end, we evaluated IPPRO under an extremely limited fine-tuning budget (10 epochs) against Variance-Based Pruning (VBP) \cite{berisha2025variance}, a recent state-of-the-art method in this regime. As shown in Table \ref{tab:comparison_vbp}, IPPRO consistently achieves higher pre-fine-tuning accuracy than VBP across DeiT and Swin-T architectures. Because our projective geometric formulation captures intrinsic functional importance through directional collapse rather than magnitude, it establishes a superior baseline immediately after pruning, ultimately yielding better post-fine-tuning results within the same constraint.

\begin{wrapfigure}{r}{0.28\textwidth}
    \vspace{-15pt}
    \centering
    \includegraphics[width=0.25\textwidth]{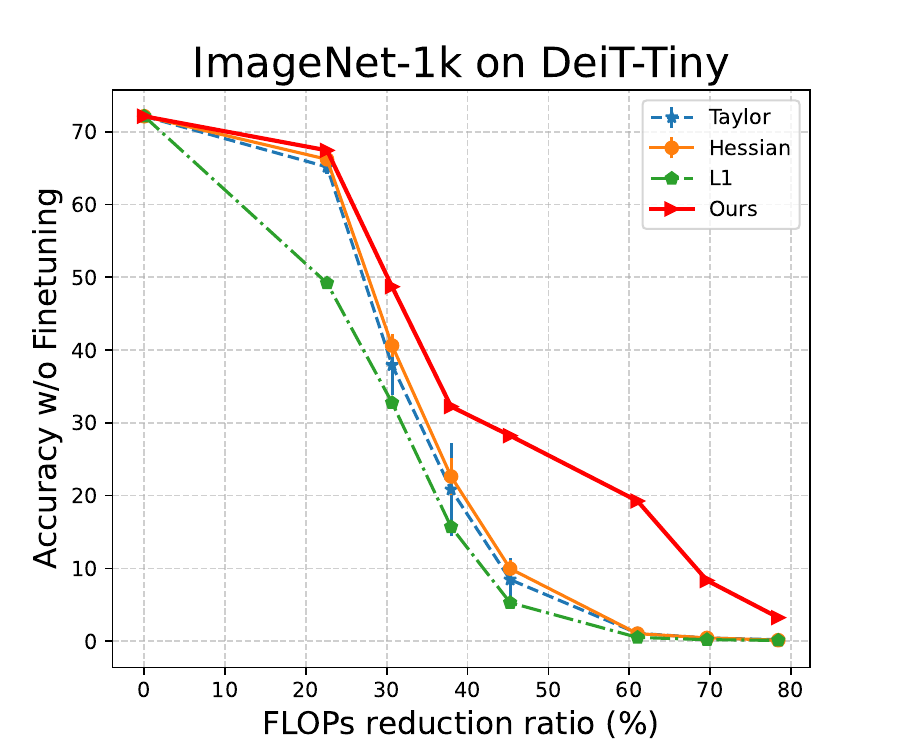}
     \caption{DeiT-Tiny Result}\label{fig:before_finetune_deittiny}  
    \vspace{-15pt}
\end{wrapfigure}
To further verify this robustness without any fine-tuning, we compared IPPRO with classical criteria (Magnitude, Taylor, and Hessian \cite{moosavi2019robustness}) on DeiT-Tiny. IPPRO consistently outperformed all baselines, with its superiority becoming even more pronounced at extreme pruning levels (\cref{fig:before_finetune_deittiny}). This preservation of network functionality demonstrates IPPRO's applicability for low-resource deployment scenarios, such as embedded or on-device settings, where fine-tuning is often impractical. Detailed no-fine-tuning evaluations for ViTs, LLMs, and DeepLabV3 are provided in Appendices~\ref{appendix:wo_ft_vit}--\ref{appendix:wo_ft_deep}.

\subsection{Performance Analysis of IPPRO on LLMs}

To evaluate cross-domain generalizability across a variety of other tasks, we conducted experiments on both the LLaMA-7B \cite{touvron2023llama} and LLaMA2-7B\cite{touvron2023llama2} models using different pruning ratios. Our experimental setup followed the configuration of LLM-Pruner \cite{ma2023llm}, so we used only randomly selected 10 Bookcorpus \cite{zhu2015aligning} dataset for the PROscore calculation, and during the performance recovery phase using the Alpaca dataset \cite{taori2023stanford}. For evaluation, we conducted zero-shot evaluations only.

We compared the results of the LLaMA-7B model on five commonsense reasoning datasets: BoolQ \cite{clark2019boolq}, PIQA \cite{bisk2020piqa}, ARC-Easy/challenge \cite{clark2018think} and OpenbookQA \cite{mihaylov2018can}. As shown in Table \ref{tab:llm_results_llama7b}, IPPRO demonstrates consistent and robust performance on all datasets.
As a result, our average score is higher than other state-of-the-art LLM pruning methods. Similarly, we compared the results of the LLaMA2-7B model on five common sense reasoning datasets: PIQA, HellaSwag \cite{zellers2019hellaswag}, WinoGrande \cite{sakaguchi2021winogrande}, ARC-easy/challenge. As shown in Table \ref{tab:llm_results_llama2_7b}, the performance of IPPRO is robust across multiple datasets, rather than being optimized for a specific task. This indicates that IPPRO ensures strong generalized performance across various tasks.

\subsection{Gradient Sample Estimation Sensitivity}\label{subsec:llm_sampling}

Because PROscore is computed from accumulated gradients, its cost scales linearly with the number of calibration samples---raising a natural question: how many samples are actually needed?

Remarkably few. On LLaMA-7B (\cref{tab:data_sensitivity}), 10 calibration samples yield an average accuracy only 0.04\% below that of 1{,}000 samples, and already surpass LLM-Pruner run on the same 10-sample budget. This rapid convergence is not specific to language models. On CIFAR-10 with ResNet-56 at 72\% FLOPs reduction (\cref{tab:sampling_results}), reducing the calibration set from 100\% to 5\% cuts scoring time by $6.3\times$ (45.4s $\to$ 7.2s) while shifting pruned accuracy by only 0.26 percentage points on average; the best 5\% run matches the full-data result within 0.04\%. The pattern holds under extreme compression: on CIFAR-100 with VGG19 at 88\% FLOPs reduction, the best 5\% run reaches 70.24\% versus 70.47\% with the complete dataset---a gap of 0.23\%.

These results suggest that directional importance in projective space is a low-variance signal: a small random subset already captures the dominant gradient directions that determine which filters collapse toward zero. This property makes IPPRO practical for large-scale deployment and opens a natural path toward stochastic online pruning. Detailed per-task breakdowns appear in \cref{tab:app_fullllm_data_sensitivity,appendix:sampling_analysis}.

\subsection{Latency Test of ViTs}

To further validate efficiency gains, we report end-to-end latency alongside FLOPs. As shown in \cref{tab:vit_latency}, IPPRO consistently improves inference latency across ViTs while incurring only minor accuracy degradation. Latency is measured on an RTX 4090 GPU with 200 warm-up runs and averaged over 1,000 runs (batch size 64). Additional latency results for LLMs are provided in \cref{appendix:latency}.

\subsection{Visual Interpretation: Filtering Spurious Neurons}\label{subsec:spurious}

\begin{figure*}[t]
    \centering
    \begin{subfigure}{0.1\textwidth}
        \includegraphics[width=\textwidth]{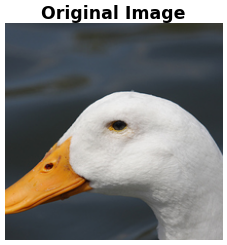}
        \vspace{2mm}
        \caption{Input}
        \label{fig:raw_128694}
    \end{subfigure}
    \begin{subfigure}[b]{0.2\textwidth}
        \includegraphics[width=\textwidth]{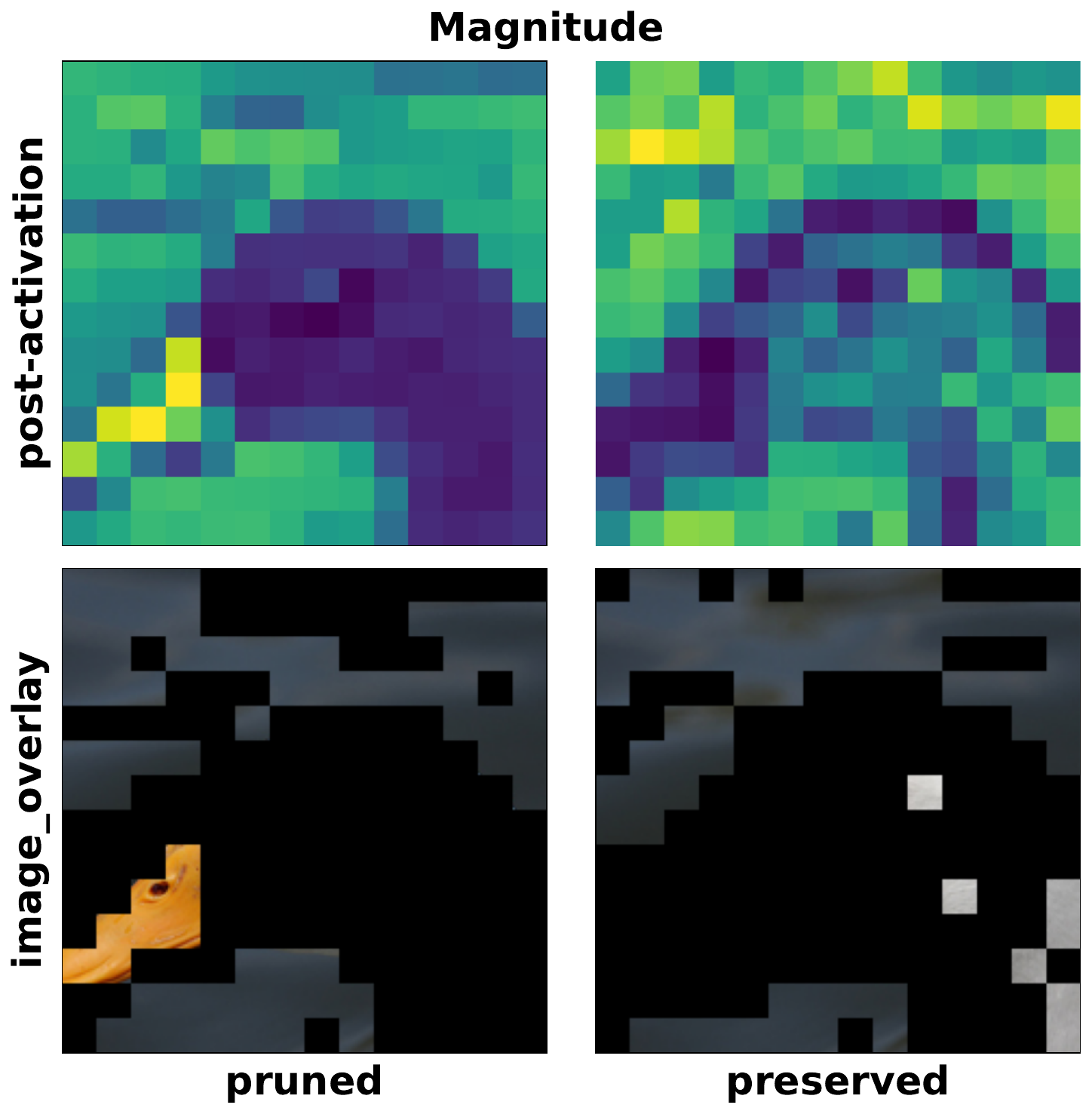}
        \caption{Magnitude}
        \label{fig:spu_mag_128694}
    \end{subfigure}
    \begin{subfigure}[b]{0.2\textwidth}
        \includegraphics[width=\textwidth]{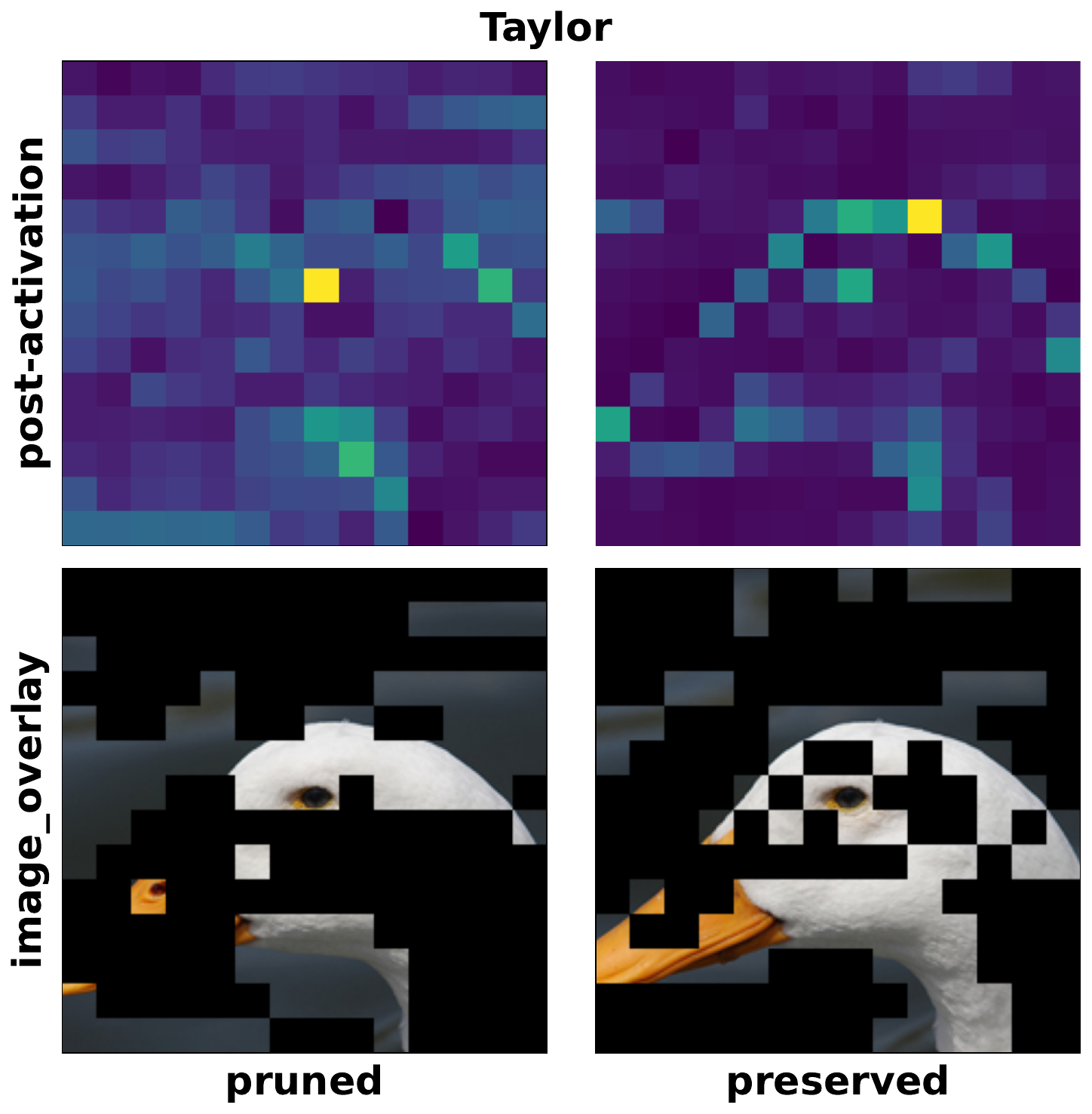}
        \caption{Taylor}
        \label{fig:spu_taylor_128694}
    \end{subfigure}
    \begin{subfigure}[b]{0.2\textwidth}
        \includegraphics[width=\textwidth]{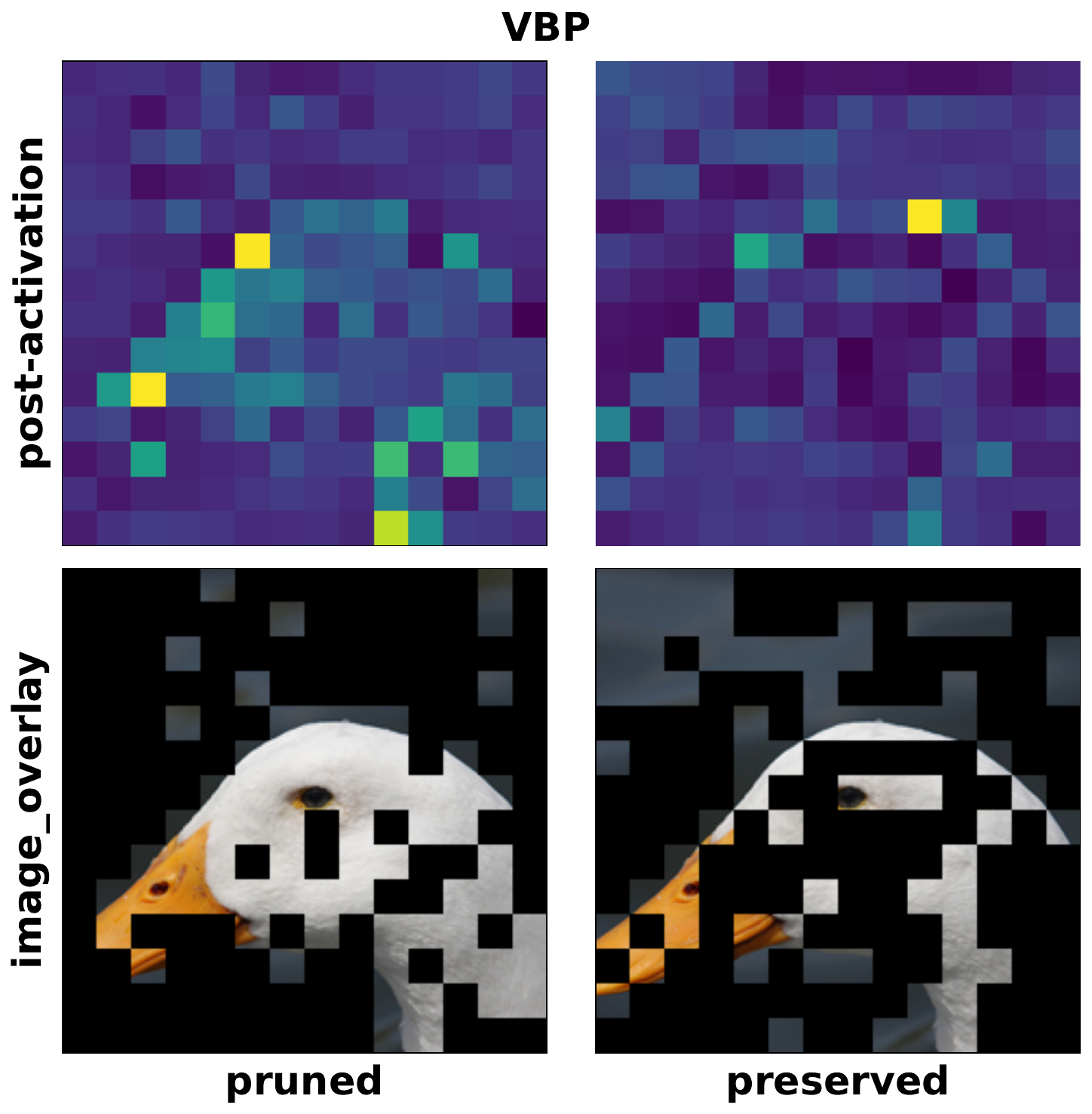}
        \caption{VBP}
        \label{fig:spu_vbp_128694}
    \end{subfigure}
    \begin{subfigure}[b]{0.2\textwidth}
        \includegraphics[width=\textwidth]{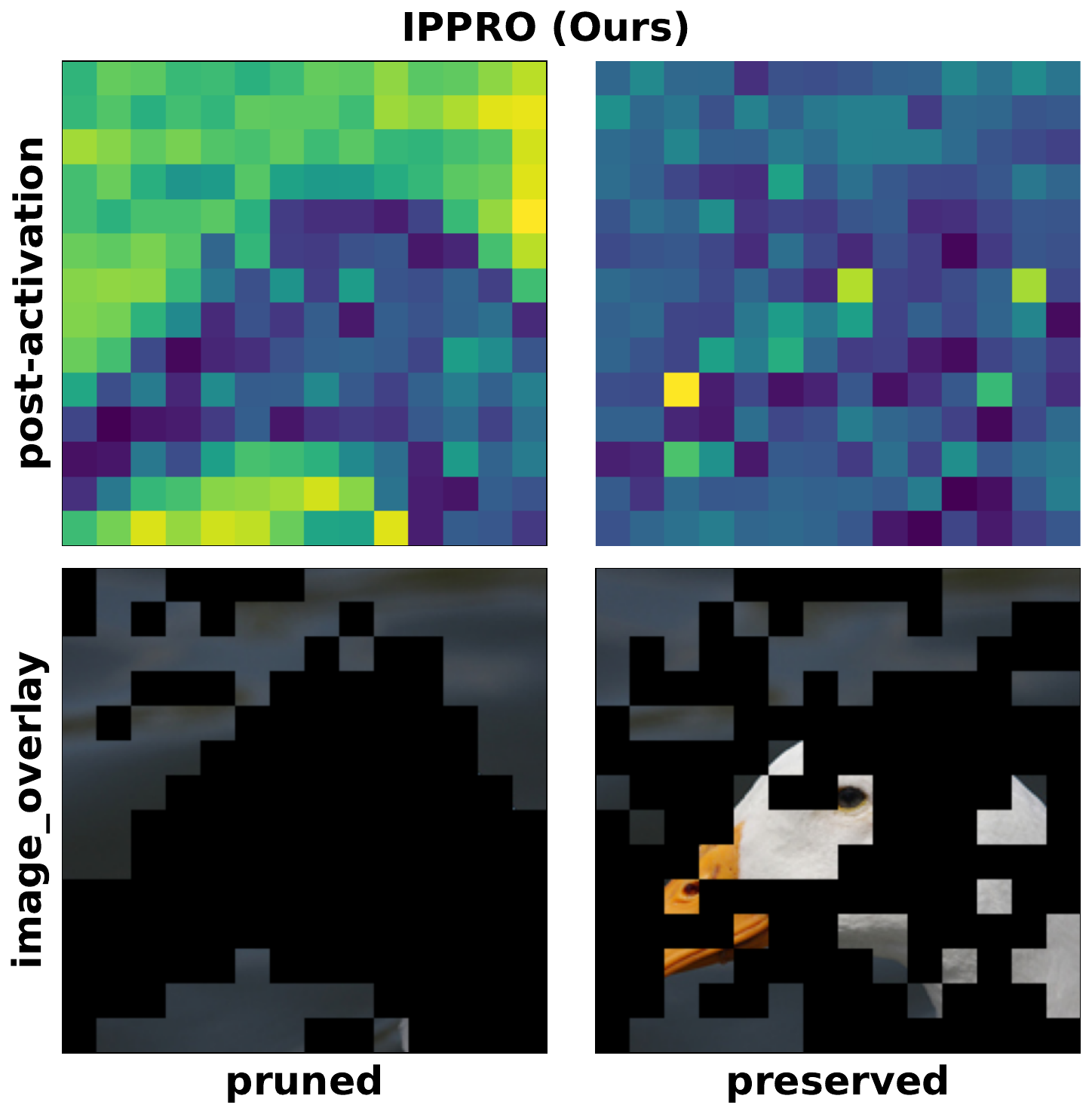}
        \caption{IPPRO (Ours)}
        \label{fig:spu_mag_128694}
    \end{subfigure}    
    \caption{Comparison of activation map and image overlay from top/bottom 10 filters for local and global criteria, on DeiT-T model's first mlp layer (duck input).}
    \label{fig:spurious}
\end{figure*}

We qualitatively analyze the property of IPPRO's pruning decision, focusing on the DeiT-Tiny model, to investigate what filters would be pruned or preserved and explain the empirical success of IPPRO (see Appendix~\ref{appendix:spurious} for setups and results).

Interestingly, IPPRO shows preference to prune spurious filters\cite{geirhos2020shortcut,beery2018recognition} and preserve robust filters\cite{ilyas2019adversarial}, as illustrated in \cref{fig:spurious}. While other baselines (magnitude, Taylor, VBP) are not correlated, filters assigned low PROscores (targeted for removal) tend to activate on background ripples or noise. Conversely, filters preserved by high PROscores concentrate their activations on structurally meaningful features, such as the duck's beak and eye.

This result offers a compelling interpretation of IPPRO: by pruning spurious filters, IPPRO preserves model performance with minimal damage and preserves test performance before the finetuning. Consequently, IPPRO achieves superior post-FT performance on \cref{tab:comparison_vbp}.

\section{Discussion}

\textbf{Pruning as directional collapse.}
IPPRO reframes filter importance through projective dynamics rather than parameter magnitude. The bifurcation boundary $c=1$ in the Catalyst framework corresponds precisely to the angular distance $\pi/4$, explaining why a single gradient step reliably predicts multi-step pruning outcomes (\cref{sec:l0_connection}).

\textbf{Why a task-agnostic IPPRO outperforms task-specific ones.}
A recurring pattern in our experiments is that IPPRO consistently matches or surpasses methods specifically designed for their target domain: SIRFP for segmentation (\cref{tab:seg_results}), SNP for transformers (\cref{tab:vit_results}), and Bonsai for LLMs (\cref{tab:llm_results_llama2_7b}). We attribute this counterintuitive outcome to a fundamental gap in the pruning literature: the absence of a principled way to compare non-zero filters against the zero filter. Task-specific methods invest in sophisticated scoring heuristics or architectural constraints but inherit the same magnitude-entangled criteria at their core (\cref{fig:main_pruning_index}). IPPRO addresses the problem at a deeper level by resolving the geometric singularity at the origin, and this foundational correction propagates across architectures and tasks without architecture-specific adaptation.

\textbf{``No-fine-tuning accuracy'' as a pruning quality diagnostic.}
The no-fine-tuning experiments (\cref{subsec:vit_performance,appendix:wo_ft_vit}) deserve particular emphasis. Post-fine-tuning accuracy---the standard metric in the literature---conflates the quality of the pruning decision with the effectiveness of the recovery procedure. An aggressive fine-tuning schedule can rescue a mediocre pruning criterion, masking real differences between methods. Pre-fine-tuning accuracy strips away this confounder. IPPRO's consistent advantage in this regime---e.g., halving the perplexity of LLM-Pruner on WikiText2 at 70\% retention (\cref{tab:llm_wo_finetune})---demonstrates that projective scoring preserves network functionality at the moment of surgery, not merely after recovery. We believe no-fine-tuning evaluation can serve as a meaningful complementary diagnostic alongside conventional post-recovery metrics.

\textbf{Limitations and future work.}
Like all one-shot methods, IPPRO scores filters independently and does not model inter-filter dependencies; explicitly capturing combinatorial interactions—e.g., via second-order projective curvature—may yield further gains under extreme compression. Because projective scoring is immune to magnitude inflation in wider networks, it may also prove disproportionately valuable at frontier scales. Extending the embedding to other compression paradigms such as quantization-aware training or low-rank factorization is another natural direction. Finally, the preferential removal of spurious filters (Section \ref{subsec:spurious}) opens a promising direction: investigating projective importance as a lens on shortcut learning and model robustness.

\section{Conclusion}

Importance-based structured pruning has long relied on filter magnitude as a proxy for functional contribution. This proxy is fragile: even minor reparameterizations produce arbitrarily different scores. IPPRO resolves this at its geometric root. By embedding filters into real projective space $\mathbb{RP}^N$, we replace magnitude thresholding with a principled geometric primitive: measuring whether a filter's gradient trajectory directionally collapses toward the zero filter. The resulting criterion, PROscore, is structurally invariant under positive rescaling, empirically robust across reparameterizations, and connected to exact $L_0$ relaxation.
The experiments confirm that this geometric foundation translates into practical gains. Across CNNs, Vision Transformers, and Large Language Models, IPPRO consistently outperforms over 30 specialized baselines. This advantage is most pronounced in the regimes where the right pruning decision matters most: high compression, where errors compound, and no-fine-tuning settings, where recovery training cannot mask a poor initial cut. Ultimately, IPPRO shows that strong magnitude dependence is not an inevitable cost of simplicity, but largely a consequence of formulating importance in the wrong geometric space. These results establish IPPRO not merely as an incremental improvement, but as a shift in the geometric language of neural network compression.

\bibliographystyle{unsrt}
\bibliography{aimlk}

@String(ECCV= {Eur. Conf. Comput. Vis.})

@String(IJCAI = {IJCAI})

@String(AAAI = {AAAI})

@String(ECCV  = {ECCV})

@inproceedings{liu2021swin,
  title={Swin transformer: Hierarchical vision transformer using shifted windows},
  author={Liu, Ze and Lin, Yutong and Cao, Yue and Hu, Han and Wei, Yixuan and Zhang, Zheng and Lin, Stephen and Guo, Baining},
  booktitle={Proceedings of the IEEE/CVF international conference on computer vision},
  pages={10012--10022},
  year={2021}
}

@article{ilyas2019adversarial,
  title={Adversarial examples are not bugs, they are features},
  author={Ilyas, Andrew and Santurkar, Shibani and Tsipras, Dimitris and Engstrom, Logan and Tran, Brandon and Madry, Aleksander},
  journal={Advances in neural information processing systems},
  volume={32},
  year={2019}
}

@inproceedings{gao2025bypass,
  title={Bypass back-propagation: Optimization-based structural pruning for large language models via policy gradient},
  author={Gao, Yuan and Liu, Zujing and Zhang, Weizhong and Du, Bo and Xia, Gui-Song},
  booktitle={Proceedings of the 63rd Annual Meeting of the Association for Computational Linguistics (Volume 1: Long Papers)},
  pages={29356--29377},
  year={2025}
}

@article{geirhos2020shortcut,
  title={Shortcut learning in deep neural networks},
  author={Geirhos, Robert and Jacobsen, J{\"o}rn-Henrik and Michaelis, Claudio and Zemel, Richard and Brendel, Wieland and Bethge, Matthias and Wichmann, Felix A},
  journal={Nature Machine Intelligence},
  volume={2},
  number={11},
  pages={665--673},
  year={2020},
  publisher={Nature Publishing Group UK London}
}

@inproceedings{beery2018recognition,
  title={Recognition in terra incognita},
  author={Beery, Sara and Van Horn, Grant and Perona, Pietro},
  booktitle={Proceedings of the European conference on computer vision (ECCV)},
  pages={456--473},
  year={2018}
}

@inproceedings{
Sagawa2020Distributionally,
title={Distributionally Robust Neural Networks},
author={Shiori Sagawa* and Pang Wei Koh* and Tatsunori B. Hashimoto and Percy Liang},
booktitle={International Conference on Learning Representations},
year={2020},
url={https://openreview.net/forum?id=ryxGuJrFvS}
}

@inproceedings{shen2025numerical,
  title={Numerical pruning for efficient autoregressive models},
  author={Shen, Xuan and Song, Zhao and Zhou, Yufa and Chen, Bo and Liu, Jing and Zhang, Ruiyi and Rossi, Ryan A and Tan, Hao and Yu, Tong and Chen, Xiang and others},
  booktitle={Proceedings of the AAAI Conference on Artificial Intelligence},
  volume={39},
  number={19},
  pages={20418--20426},
  year={2025}
}

@article{berisha2025variance,
  title={Variance-Based Pruning for Accelerating and Compressing Trained Networks},
  author={Berisha, Uranik and Mehnert, Jens and Condurache, Alexandru Paul},
  journal={arXiv preprint arXiv:2507.12988},
  year={2025}
}

@inproceedings{moosavi2019robustness,
  title={Robustness via curvature regularization, and vice versa},
  author={Moosavi-Dezfooli, Seyed-Mohsen and Fawzi, Alhussein and Uesato, Jonathan and Frossard, Pascal},
  booktitle={Proceedings of the IEEE/CVF Conference on Computer Vision and Pattern Recognition},
  pages={9078--9086},
  year={2019}
}

@article{michel2019sixteen,
  title={Are sixteen heads really better than one?},
  author={Michel, Paul and Levy, Omer and Neubig, Graham},
  journal={Advances in neural information processing systems},
  volume={32},
  year={2019}
}

@misc{taori2023stanford,
  title={Stanford alpaca: An instruction-following llama model},
  author={Taori, Rohan and Gulrajani, Ishaan and Zhang, Tianyi and Dubois, Yann and Li, Xuechen and Guestrin, Carlos and Liang, Percy and Hashimoto, Tatsunori B},
  year={2023},
  publisher={Stanford, CA, USA}
}

@inproceedings{zhu2015aligning,
  title={Aligning books and movies: Towards story-like visual explanations by watching movies and reading books},
  author={Zhu, Yukun and Kiros, Ryan and Zemel, Rich and Salakhutdinov, Ruslan and Urtasun, Raquel and Torralba, Antonio and Fidler, Sanja},
  booktitle={Proceedings of the IEEE international conference on computer vision},
  pages={19--27},
  year={2015}
}

@article{wang2025dobi,
  title={Dobi-SVD: Differentiable SVD for LLM Compression and Some New Perspectives},
  author={Wang, Qinsi and Ke, Jinghan and Tomizuka, Masayoshi and Chen, Yiran and Keutzer, Kurt and Xu, Chenfeng},
  journal={arXiv preprint arXiv:2502.02723},
  year={2025}
}

@inproceedings{
yu2022unified,
title={Unified Visual Transformer Compression},
author={Shixing Yu and Tianlong Chen and Jiayi Shen and Huan Yuan and Jianchao Tan and Sen Yang and Ji Liu and Zhangyang Wang},
booktitle={International Conference on Learning Representations},
year={2022},
url={https://openreview.net/forum?id=9jsZiUgkCZP}
}

@inproceedings{touvron2021training,
  title={Training data-efficient image transformers \& distillation through attention},
  author={Touvron, Hugo and Cord, Matthieu and Douze, Matthijs and Massa, Francisco and Sablayrolles, Alexandre and J{\'e}gou, Herv{\'e}},
  booktitle={International conference on machine learning},
  pages={10347--10357},
  year={2021},
  organization={PMLR}
}

@inproceedings{yu2023x,
  title={X-pruner: explainable pruning for vision transformers},
  author={Yu, Lu and Xiang, Wei},
  booktitle={Proceedings of the IEEE/CVF conference on computer vision and pattern recognition},
  pages={24355--24363},
  year={2023}
}

@inproceedings{yu2022width,
  title={Width \& depth pruning for vision transformers},
  author={Yu, Fang and Huang, Kun and Wang, Meng and Cheng, Yuan and Chu, Wei and Cui, Li},
  booktitle={Proceedings of the AAAI conference on artificial intelligence},
  volume={36},
  number={3},
  pages={3143--3151},
  year={2022}
}

@inproceedings{yang2023global,
  title={Global vision transformer pruning with hessian-aware saliency},
  author={Yang, Huanrui and Yin, Hongxu and Shen, Maying and Molchanov, Pavlo and Li, Hai and Kautz, Jan},
  booktitle={Proceedings of the IEEE/CVF conference on computer vision and pattern recognition},
  pages={18547--18557},
  year={2023}
}

@inproceedings{an2024fluctuation,
  title={Fluctuation-based adaptive structured pruning for large language models},
  author={An, Yongqi and Zhao, Xu and Yu, Tao and Tang, Ming and Wang, Jinqiao},
  booktitle={Proceedings of the AAAI Conference on Artificial Intelligence},
  volume={38},
  number={10},
  pages={10865--10873},
  year={2024}
}

@misc{zhu2021visiontransformerpruning,
      title={Vision Transformer Pruning}, 
      author={Mingjian Zhu and Yehui Tang and Kai Han},
      year={2021},
      eprint={2104.08500},
      archivePrefix={arXiv},
      primaryClass={cs.CV},
      url={https://arxiv.org/abs/2104.08500}, 
}

@inproceedings{shim2024snp,
  title={SNP: Structured Neuron-level Pruning to Preserve Attention Scores},
  author={Shim, Kyunghwan and Yun, Jaewoong and Choi, Shinkook},
  booktitle={European Conference on Computer Vision},
  pages={90--104},
  year={2024},
  organization={Springer}
}

@article{ma2023llm,
  title={Llm-pruner: On the structural pruning of large language models},
  author={Ma, Xinyin and Fang, Gongfan and Wang, Xinchao},
  journal={Advances in neural information processing systems},
  volume={36},
  pages={21702--21720},
  year={2023}
}

@inproceedings{wang2019eigendamage,
  title={Eigendamage: Structured pruning in the kronecker-factored eigenbasis},
  author={Wang, Chaoqi and Grosse, Roger and Fidler, Sanja and Zhang, Guodong},
  booktitle={International conference on machine learning},
  pages={6566--6575},
  year={2019},
  organization={PMLR}
}

@article{krizhevsky2009cifar,
  title={Learning multiple layers of features from tiny images},
  author={Krizhevsky, Alex and others},
  year={2009}
}

@inproceedings{fang2023depgraph,
  title={Depgraph: Towards any structural pruning},
  author={Fang, Gongfan and Ma, Xinyin and Song, Mingli and Mi, Michael Bi and Wang, Xinchao},
  booktitle={Proceedings of the IEEE/CVF Conference on Computer Vision and Pattern Recognition},
  pages={16091--16101},
  year={2023}
}

@article{wang2020greg,
  title={Neural pruning via growing regularization},
  author={Wang, Huan and Qin, Can and Zhang, Yulun and Fu, Yun},
  journal={arXiv preprint arXiv:2012.09243},
  year={2020}
}

@inproceedings{
catalyst,
title={Catalyst: Structured Pruning with Robust Bifurcation Dynamics},
author={Jaeheun Jung and Donghun Lee},
booktitle={High-dimensional Learning Dynamics 2025},
year={2025},
url={https://openreview.net/forum?id=v0l6QS3Umt}
}

@article{sui2021chip,
  title={Chip: Channel independence-based pruning for compact neural networks},
  author={Sui, Yang and Yin, Miao and Xie, Yi and Phan, Huy and Aliari Zonouz, Saman and Yuan, Bo},
  journal={Advances in Neural Information Processing Systems},
  volume={34},
  pages={24604--24616},
  year={2021}
}

@article{tang2020scop,
  title={Scop: Scientific control for reliable neural network pruning},
  author={Tang, Yehui and Wang, Yunhe and Xu, Yixing and Tao, Dacheng and Xu, Chunjing and Xu, Chao and Xu, Chang},
  journal={Advances in Neural Information Processing Systems},
  volume={33},
  pages={10936--10947},
  year={2020}
}

@inproceedings{lin2020hrank,
  title={Hrank: Filter pruning using high-rank feature map},
  author={Lin, Mingbao and Ji, Rongrong and Wang, Yan and Zhang, Yichen and Zhang, Baochang and Tian, Yonghong and Shao, Ling},
  booktitle={Proceedings of the IEEE/CVF conference on computer vision and pattern recognition},
  pages={1529--1538},
  year={2020}
}

@inproceedings{lin2019towards,
  title={Towards optimal structured cnn pruning via generative adversarial learning},
  author={Lin, Shaohui and Ji, Rongrong and Yan, Chenqian and Zhang, Baochang and Cao, Liujuan and Ye, Qixiang and Huang, Feiyue and Doermann, David},
  booktitle={Proceedings of the IEEE/CVF conference on computer vision and pattern recognition},
  pages={2790--2799},
  year={2019}
}

@inproceedings{molchanov2019importance,
  title={Importance estimation for neural network pruning},
  author={Molchanov, Pavlo and Mallya, Arun and Tyree, Stephen and Frosio, Iuri and Kautz, Jan},
  booktitle={Proceedings of the IEEE/CVF conference on computer vision and pattern recognition},
  pages={11264--11272},
  year={2019}
}

@article{luo2020autopruner,
  title={Autopruner: An end-to-end trainable filter pruning method for efficient deep model inference},
  author={Luo, Jian-Hao and Wu, Jianxin},
  journal={Pattern Recognition},
  volume={107},
  pages={107461},
  year={2020},
  publisher={Elsevier}
}

@inproceedings{he2019filter,
  title={Filter pruning via geometric median for deep convolutional neural networks acceleration},
  author={He, Yang and Liu, Ping and Wang, Ziwei and Hu, Zhilan and Yang, Yi},
  booktitle={Proceedings of the IEEE/CVF conference on computer vision and pattern recognition},
  pages={4340--4349},
  year={2019}
}

@inproceedings{fang2024isomorphic,
  title={Isomorphic pruning for vision models},
  author={Fang, Gongfan and Ma, Xinyin and Mi, Michael Bi and Wang, Xinchao},
  booktitle={European Conference on Computer Vision},
  pages={232--250},
  year={2024},
  organization={Springer}
}

@article{tang2024filter,
  title={Filter Pruning Based on Information Capacity and Independence},
  author={Tang, Xiaolong and Ye, Shuo and Shi, Yufeng and Hu, Tianheng and Peng, Qinmu and You, Xinge},
  journal={IEEE Transactions on Neural Networks and Learning Systems},
  year={2024},
  publisher={IEEE}
}

@INPROCEEDINGS{10484493,
  author={Gupta, Arshita and Bau, Tien and Kim, Joonsoo and Zhu, Zhe and Jha, Sumit and Garud, Hrishikesh},
  booktitle={2024 IEEE/CVF Winter Conference on Applications of Computer Vision (WACV)}, 
  title={Torque based Structured Pruning for Deep Neural Network}, 
  year={2024},
  volume={},
  number={},
  pages={2699-2708},
  doi={10.1109/WACV57701.2024.00269}}

@article{han2015deep,
  title={Deep compression: Compressing deep neural networks with pruning, trained quantization and huffman coding},
  author={Han, Song and Mao, Huizi and Dally, William J},
  journal={arXiv preprint arXiv:1510.00149},
  year={2015}
}

@article{touvron2023llama2,
  title={Llama 2: Open foundation and fine-tuned chat models},
  author={Touvron, Hugo and Martin, Louis and Stone, Kevin and Albert, Peter and Almahairi, Amjad and Babaei, Yasmine and Bashlykov, Nikolay and Batra, Soumya and Bhargava, Prajjwal and Bhosale, Shruti and others},
  journal={arXiv preprint arXiv:2307.09288},
  year={2023}
}

@article{blalock2020state,
  title={What is the state of neural network pruning?},
  author={Blalock, Davis and Gonzalez Ortiz, Jose Javier and Frankle, Jonathan and Guttag, John},
  journal={Proceedings of machine learning and systems},
  volume={2},
  pages={129--146},
  year={2020}
}

@article{chen2021chasing,
  title={Chasing sparsity in vision transformers: An end-to-end exploration},
  author={Chen, Tianlong and Cheng, Yu and Gan, Zhe and Yuan, Lu and Zhang, Lei and Wang, Zhangyang},
  journal={Advances in neural information processing systems},
  volume={34},
  pages={19974--19988},
  year={2021}
}

@inproceedings{
filters2016pruning,
title={Pruning Filters for Efficient ConvNets},
author={Hao Li and Asim Kadav and Igor Durdanovic and Hanan Samet and Hans Peter Graf},
booktitle={International Conference on Learning Representations},
year={2017},
url={https://openreview.net/forum?id=rJqFGTslg}
}

@article{he2018soft,
  title={Soft filter pruning for accelerating deep convolutional neural networks},
  author={He, Yang and Kang, Guoliang and Dong, Xuanyi and Fu, Yanwei and Yang, Yi},
  journal={arXiv preprint arXiv:1808.06866},
  year={2018}
}

@article{wang2019cop,
  title={Cop: Customized deep model compression via regularized correlation-based filter-level pruning},
  author={Wang, Wenxiao and Fu, Cong and Guo, Jishun and Cai, Deng and He, Xiaofei},
  journal={arXiv preprint arXiv:1906.10337},
  year={2019}
}

@inproceedings{wang2021convolutional,
  title={Convolutional neural network pruning with structural redundancy reduction},
  author={Wang, Zi and Li, Chengcheng and Wang, Xiangyang},
  booktitle={Proceedings of the IEEE/CVF conference on computer vision and pattern recognition},
  pages={14913--14922},
  year={2021}
}

@article{basha2024deep,
  title={Deep model compression based on the training history},
  author={Basha, SH Shabbeer and Farazuddin, Mohammad and Pulabaigari, Viswanath and Dubey, Shiv Ram and Mukherjee, Snehasis},
  journal={Neurocomputing},
  volume={573},
  pages={127257},
  year={2024},
  publisher={Elsevier}
}

@inproceedings{liu2021group,
  title={Group fisher pruning for practical network compression},
  author={Liu, Liyang and Zhang, Shilong and Kuang, Zhanghui and Zhou, Aojun and Xue, Jing-Hao and Wang, Xinjiang and Chen, Yimin and Yang, Wenming and Liao, Qingmin and Zhang, Wayne},
  booktitle={International Conference on Machine Learning},
  pages={7021--7032},
  year={2021},
  organization={PMLR}
}

@inproceedings{liu2019metapruning,
  title={Metapruning: Meta learning for automatic neural network channel pruning},
  author={Liu, Zechun and Mu, Haoyuan and Zhang, Xiangyu and Guo, Zichao and Yang, Xin and Cheng, Kwang-Ting and Sun, Jian},
  booktitle={Proceedings of the IEEE/CVF international conference on computer vision},
  pages={3296--3305},
  year={2019}
}

@inproceedings{he2017channel,
  title={Channel pruning for accelerating very deep neural networks},
  author={He, Yihui and Zhang, Xiangyu and Sun, Jian},
  booktitle={Proceedings of the IEEE international conference on computer vision},
  pages={1389--1397},
  year={2017}
}

@inproceedings{wu2025structural,
  title={Structural Pruning via Spatial-aware Information Redundancy for Semantic Segmentation},
  author={Wu, Dongyue and Guo, Zilin and Yu, Li and Sang, Nong and Gao, Changxin},
  booktitle={Proceedings of the AAAI Conference on Artificial Intelligence},
  volume={39},
  number={8},
  pages={8368--8376},
  year={2025}
}

@article{wang2023dcfp,
  title={DCFP: Distribution calibrated filter pruning for lightweight and accurate long-tail semantic segmentation},
  author={Wang, Zixiao and Xie, Hongtao and Wang, Yuxin and Xu, Hai and Jin, Guoqing},
  journal={IEEE Transactions on Circuits and Systems for Video Technology},
  volume={34},
  number={7},
  pages={6063--6076},
  year={2023},
  publisher={IEEE}
}

@article{chen2017deeplab,
  title={Deeplab: Semantic image segmentation with deep convolutional nets, atrous convolution, and fully connected crfs},
  author={Chen, Liang-Chieh and Papandreou, George and Kokkinos, Iasonas and Murphy, Kevin and Yuille, Alan L},
  journal={IEEE transactions on pattern analysis and machine intelligence},
  volume={40},
  number={4},
  pages={834--848},
  year={2017},
  publisher={IEEE}
}

@inproceedings{cordts2016cityscapes,
  title={The cityscapes dataset for semantic urban scene understanding},
  author={Cordts, Marius and Omran, Mohamed and Ramos, Sebastian and Rehfeld, Timo and Enzweiler, Markus and Benenson, Rodrigo and Franke, Uwe and Roth, Stefan and Schiele, Bernt},
  booktitle={Proceedings of the IEEE conference on computer vision and pattern recognition},
  pages={3213--3223},
  year={2016}
}

@article{lin2024gat,
  title={GAT TransPruning: progressive channel pruning strategy combining graph attention network and transformer},
  author={Lin, Yu-Chen and Wang, Chia-Hung and Lin, Yu-Cheng},
  journal={PeerJ Computer Science},
  volume={10},
  pages={e2012},
  year={2024},
  publisher={PeerJ Inc.}
}

@article{yang2023filter,
  title={Filter pruning via attention consistency on feature maps},
  author={Yang, Huoxiang and Liang, Yongsheng and Liu, Wei and Meng, Fanyang},
  journal={Applied Sciences},
  volume={13},
  number={3},
  pages={1964},
  year={2023},
  publisher={MDPI}
}

@inproceedings{sun2024towards,
  title={Towards better structured pruning saliency by reorganizing convolution},
  author={Sun, Xinglong and Shi, Humphrey},
  booktitle={Proceedings of the IEEE/CVF Winter Conference on Applications of Computer Vision},
  pages={2204--2214},
  year={2024}
}

@article{xu2020layer,
  title={Layer pruning via fusible residual convolutional block for deep neural networks},
  author={Xu, Pengtao and Cao, Jian and Shang, Fanhua and Sun, Wenyu and Li, Pu},
  journal={arXiv preprint arXiv:2011.14356},
  year={2020}
}

@article{wang2019dbp,
  title={DBP: Discrimination based block-level pruning for deep model acceleration},
  author={Wang, Wenxiao and Zhao, Shuai and Chen, Minghao and Hu, Jinming and Cai, Deng and Liu, Haifeng},
  journal={arXiv preprint arXiv:1912.10178},
  year={2019}
}

@article{yu2019autoslim,
  title={Autoslim: Towards one-shot architecture search for channel numbers},
  author={Yu, Jiahui and Huang, Thomas},
  journal={arXiv preprint arXiv:1903.11728},
  year={2019}
}

@article{shen2022structural,
  title={Structural pruning via latency-saliency knapsack},
  author={Shen, Maying and Yin, Hongxu and Molchanov, Pavlo and Mao, Lei and Liu, Jianna and Alvarez, Jose M},
  journal={Advances in Neural Information Processing Systems},
  volume={35},
  pages={12894--12908},
  year={2022}
}

@article{you2019gate,
  title={Gate decorator: Global filter pruning method for accelerating deep convolutional neural networks},
  author={You, Zhonghui and Yan, Kun and Ye, Jinmian and Ma, Meng and Wang, Ping},
  journal={Advances in neural information processing systems},
  volume={32},
  year={2019}
}

@inproceedings{duan2022network,
  title={Network pruning via feature shift minimization},
  author={Duan, Yuanzhi and Zhou, Yue and He, Peng and Liu, Qiang and Duan, Shukai and Hu, Xiaofang},
  booktitle={Proceedings of the Asian conference on computer vision},
  pages={4044--4060},
  year={2022}
}

@article{chen2024effective,
  title={An effective information theoretic framework for channel pruning},
  author={Chen, Yihao and Wang, Zefang},
  journal={arXiv preprint arXiv:2408.16772},
  year={2024}
}

@inproceedings{jiang2022channel,
  title={On the Channel Pruning using Graph Convolution Network for Convolutional Neural Network Acceleration.},
  author={Jiang, Di and Cao, Yuan and Yang, Qiang},
  booktitle={IJCAI},
  pages={3107--3113},
  year={2022}
}

@article{wang2022qsfm,
  title={QSFM: Model pruning based on quantified similarity between feature maps for AI on edge},
  author={Wang, Zidu and Liu, Xuexin and Huang, Long and Chen, Yunqing and Zhang, Yufei and Lin, Zhikang and Wang, Rui},
  journal={IEEE Internet of Things Journal},
  volume={9},
  number={23},
  pages={24506--24515},
  year={2022},
  publisher={IEEE}
}

@article{liu2023filter,
  title={Filter pruning by quantifying feature similarity and entropy of feature maps},
  author={Liu, Yajun and Fan, Kefeng and Wu, Dakui and Zhou, Wenju},
  journal={Neurocomputing},
  volume={544},
  pages={126297},
  year={2023},
  publisher={Elsevier}
}

@article{chen2023rgp,
  title={Rgp: Neural network pruning through regular graph with edges swapping},
  author={Chen, Zhuangzhi and Xiang, Jingyang and Lu, Yao and Xuan, Qi and Wang, Zhen and Chen, Guanrong and Yang, Xiaoniu},
  journal={IEEE Transactions on Neural Networks and Learning Systems},
  year={2023},
  publisher={IEEE}
}

@article{lv2024fgp,
  title={FGP: Feature-Gradient-Prune for Efficient Convolutional Layer Pruning},
  author={Lv, Qingsong and Sun, Jiasheng and Zhou, Sheng and Zhang, Xu and Li, Liangcheng and Gao, Yun and Qiao, Sun and Song, Jie and Bu, Jiajun},
  journal={arXiv preprint arXiv:2411.12781},
  year={2024}
}

@article{sun2024multi,
  title={Multi-Dimensional Pruning: Joint Channel, Layer and Block Pruning with Latency Constraint},
  author={Sun, Xinglong and Lakshmanan, Barath and Shen, Maying and Lan, Shiyi and Chen, Jingde and Alvarez, Jose},
  journal={arXiv preprint arXiv:2406.12079},
  year={2024}
}

@article{li2022efficientformer,
  title={Efficientformer: Vision transformers at mobilenet speed},
  author={Li, Yanyu and Yuan, Geng and Wen, Yang and Hu, Ju and Evangelidis, Georgios and Tulyakov, Sergey and Wang, Yanzhi and Ren, Jian},
  journal={Advances in Neural Information Processing Systems},
  volume={35},
  pages={12934--12949},
  year={2022}
}

@inproceedings{deng2009imagenet,
  title={Imagenet: A large-scale hierarchical image database},
  author={Deng, Jia and Dong, Wei and Socher, Richard and Li, Li-Jia and Li, Kai and Fei-Fei, Li},
  booktitle={2009 IEEE conference on computer vision and pattern recognition},
  pages={248--255},
  year={2009},
  organization={Ieee}
}

@article{touvron2023llama,
  title={Llama: Open and efficient foundation language models},
  author={Touvron, Hugo and Lavril, Thibaut and Izacard, Gautier and Martinet, Xavier and Lachaux, Marie-Anne and Lacroix, Timoth{\'e}e and Rozi{\`e}re, Baptiste and Goyal, Naman and Hambro, Eric and Azhar, Faisal and others},
  journal={arXiv preprint arXiv:2302.13971},
  year={2023}
}

@article{clark2019boolq,
  title={Boolq: Exploring the surprising difficulty of natural yes/no questions},
  author={Clark, Christopher and Lee, Kenton and Chang, Ming-Wei and Kwiatkowski, Tom and Collins, Michael and Toutanova, Kristina},
  journal={arXiv preprint arXiv:1905.10044},
  year={2019}
}

@inproceedings{bisk2020piqa,
  title={Piqa: Reasoning about physical commonsense in natural language},
  author={Bisk, Yonatan and Zellers, Rowan and Gao, Jianfeng and Choi, Yejin and others},
  booktitle={Proceedings of the AAAI conference on artificial intelligence},
  volume={34},
  number={05},
  pages={7432--7439},
  year={2020}
}

@article{clark2018think,
  title={Think you have solved question answering? try arc, the ai2 reasoning challenge},
  author={Clark, Peter and Cowhey, Isaac and Etzioni, Oren and Khot, Tushar and Sabharwal, Ashish and Schoenick, Carissa and Tafjord, Oyvind},
  journal={arXiv preprint arXiv:1803.05457},
  year={2018}
}

@article{zellers2019hellaswag,
  title={Hellaswag: Can a machine really finish your sentence?},
  author={Zellers, Rowan and Holtzman, Ari and Bisk, Yonatan and Farhadi, Ali and Choi, Yejin},
  journal={arXiv preprint arXiv:1905.07830},
  year={2019}
}

@article{sakaguchi2021winogrande,
  title={Winogrande: An adversarial winograd schema challenge at scale},
  author={Sakaguchi, Keisuke and Bras, Ronan Le and Bhagavatula, Chandra and Choi, Yejin},
  journal={Communications of the ACM},
  volume={64},
  number={9},
  pages={99--106},
  year={2021},
  publisher={ACM New York, NY, USA}
}

@article{ashkboos2024slicegpt,
  title={Slicegpt: Compress large language models by deleting rows and columns},
  author={Ashkboos, Saleh and Croci, Maximilian L and Nascimento, Marcelo Gennari do and Hoefler, Torsten and Hensman, James},
  journal={arXiv preprint arXiv:2401.15024},
  year={2024}
}

@article{dery2024everybody,
  title={Everybody prune now: Structured pruning of llms with only forward passes},
  author={Dery, Lucio and Kolawole, Steven and Kagy, Jean-Fran{\c{c}}ois and Smith, Virginia and Neubig, Graham and Talwalkar, Ameet},
  journal={arXiv preprint arXiv:2402.05406},
  year={2024}
}

@article{sun2023simple,
  title={A simple and effective pruning approach for large language models},
  author={Sun, Mingjie and Liu, Zhuang and Bair, Anna and Kolter, J Zico},
  journal={arXiv preprint arXiv:2306.11695},
  year={2023}
}

@article{mihaylov2018can,
  title={Can a suit of armor conduct electricity? a new dataset for open book question answering},
  author={Mihaylov, Todor and Clark, Peter and Khot, Tushar and Sabharwal, Ashish},
  journal={arXiv preprint arXiv:1809.02789},
  year={2018}
}

\newpage
\appendix
\onecolumn

\section{Implementation Detail}\label{sec:implementation}
Note that the model parameters are not updated during the PROscore computation. 
The extended model is reverted to the original model, after which we prune the filters according to the obtained PROscore. 

To reduce the randomness of the empirical results of our implementation, we fixed the pre-trained weights $W$ and the parameter $D$ obtained from the extended projective space computed using the pretrained weights. 
In addition, we accumulate the gradients $\mathcal\nabla_{F_i}\mathcal{L}$ and $\frac{\partial \mathcal{L}}{\partial D_i}$ from each batch during backpropagation process using the entire dataset $\mathcal{D}$, unless otherwise mentioned. 

Using the accumulated gradient values, along with the original model weights and the expanded parameter $D$, the layer-wise PROscore is computed according to \cref{eq:proscore}. 
Using the PROscore, layer-wise pruning is performed on the original base model, which does not include the extended parameter $D$, using the previously computed importance scores. 

\setlength{\textfloatsep}{10pt}
\setlength{\floatsep}{10pt}
\setlength{\intextsep}{10pt}
\setlength{\dbltextfloatsep}{10pt}
\setlength{\dblfloatsep}{10pt}
\begin{table*}[t]
\centering
\caption{Fine-tuning configurations for models.}\label{tab:finetuning_recipe}
\label{tab:train-configs}
\resizebox{0.8\textwidth}{!}{%
\begin{tabular}{lccccc}
\toprule
\textbf{Fine-tuning Configs} & \textbf{ResNet-50} & \textbf{MobileNetv2} & \textbf{ResNet-56} & \textbf{VGG-19} & \textbf{DeiT} \\
\midrule
dataset  &  ImageNet-1K &  ImageNet-1K & CIFAR-10 &  CIFAR-100 & ImageNet-1K \\
epochs  & 180 & 300 & 400 & 400 & 300 \\
batch size  & 128 & 256 & 256 & 256 & 512 \\
optimizer                & SGD         & SGD         & SGD        & SGD        & AdamW \\
learning rate scheduler   & step  & cosine        & step   & cosine     & cosine \\
step size & 40  & -        & 30   & -     & - \\
base learning rate       & 0.01        & 0.01         & 0.025       & 0.017      & 0.0005 \\
weight decay             & 1e-4          & 1e-4          & 0.0005       & 0.005       & 0.01 \\
optimizer momentum       & 0.9  & 0.9  & 0.9        & 0.9        & (0.9, 0.999) \\
\bottomrule
\end{tabular}%
}
\end{table*}

\section{Finetune Recipe}\label{sec:finetune_recipe}

In this section, we detail the hyperparameters used for fine-tuning as shown in \cref{tab:finetuning_recipe}. All or some of the models were fine-tuned using the PyTorch 1.13 framework on an NVIDIA RTX 4090 GPU. 
Additionally, we emphasize that we did not use external data augmentation skills such as Color-jitter or Mix-up, which are often employed to improve the performance in other pruning methods, since it raises ethical issues regarding fair comparison, as the pretrained models did not use them, and results obtained with additional augmentation may not reflect the benefit of the pruning method alone.

\section{Formal Connection to Exact $L_0$ Relaxation}\label{appendix:l0_connection}

The PROscore defined in Equation~(\ref{eq:proscore}) is intrinsically connected to the exact $L_0$ relaxation framework recently established by Catalyst~\cite{catalyst}.

Conventional magnitude-based pruning can be viewed as an approximate convex relaxation of the $L_0$ sparsity constraint, analogous to Lasso. However, this relaxation suffers from a geometric mismatch: the $\epsilon$-closure of $L_1$ regularization forms a ball centered at zero, whereas the ideal pruning-invariant set $\mathcal{X}_{\text{tgt}}$ is a union of coordinate hyperplanes. Catalyst resolves this by introducing auxiliary variables $D_i$ and minimizing $\|DW\|_{2,1}$ in an extended parameter space, achieving an \emph{exact} relaxation whose nontrivial global minima lie on $\mathcal{X}_{\text{tgt}}$. When initialized at $D_i = \|F_i\|$, the ratio $c_i = D_i / \|F_i\|$ undergoes provable exponential bifurcation at the critical boundary $c = 1$: filters with $c_i > 1$ collapse toward zero while others are preserved. Crucially, this fate is determined in the very early stages of optimization.

IPPRO provides a natural geometric interpretation of this bifurcation. By embedding filters into $\mathbb{RP}^N$ via $\mathrm{embed}(F_i) = [\|F_i\| : F_i]$, all filters are placed at an angular distance $\pi/4$ from the origin---which corresponds exactly to the initial condition $c = 1$ in the Catalyst framework. The PROscore then measures the one-step angular displacement from this boundary. 

Precisely, PROscore computes $\tan(\theta(p_i'))$, which is mathematically equivalent to the reciprocal of the one-step updated ratio: $\mathrm{PROscore}_\lambda(i) = 1/c_i'$ (we use the tangent instead of the cotangent so that smaller values indicate less important filters). Thus, $\mathrm{PROscore}(i) < 1$ directly indicates movement toward the zero filter $[1:0:\cdots:0]$ (meaning the filter will be pruned under continued dynamics), while $\mathrm{PROscore}(i) > 1$ indicates preservation. Since the underlying bifurcation is exponential, this initial displacement captured by a single gradient step in projective space reliably predicts the final multi-step pruning outcome.

\section{Investigation of IPPRO pruning decision}

IPPRO prunes filters with small $PROscore$, the one-step updated angular displacement in extended space. Regarding the consistent success of IPPRO across diverse architecture and modality, the property of IPPRO-pruned filters is questionable. 
In this section, we investigate $PROscore$ through backpropagation and find out what filter would be selected to be pruned. Interestingly, IPPRO likely prunes \textbf{spurious filters}, which focus on unimportant information of input, such as background rather than the object. 

\paragraph{Reformulation of $PROscore$}

Recall the filters $F_i$ and input $x$, we denote the $i$th entry of pre-activation as $h_{i,x}=F_i\cdot x$ and post activation $a_{i,x}=\sigma(h_{i,x})$. 

\begin{definition}
We define $I^f_i(\mathcal{D})=\sum_{x\in \mathcal{D}} \frac{\partial L}{\partial a_{i,x}}f'(h_{i,x})h_{i,x}$. 
\end{definition}

\begin{remark}

\begin{enumerate}
    \item $I^f_i(\mathcal{D})$ generalizes Taylor importance $F_i\cdot \nabla_{F_i} \mathcal{L}$. If $f$ is equal to activation function $\sigma$, then $I^\sigma_i(\mathcal{D}) = F_i\cdot \nabla_{F_i} \mathcal{L}$ since $h_{i,x}=F_i\cdot x$ and $\nabla_{F_i}\mathcal{L} =\sum_{x\in\mathcal{D}} \frac{\partial L}{\partial a_{i,x}}\sigma'(h_{i,x})x$. 
    \item If $f$ is linear, $I^{id}_i(\mathcal{D}) = \sum_{x\in\mathcal{D}} \frac{\partial L}{\partial a_{i,x}}h_{i,x}$ measures the alignment of pre-activation and gradient or post-activation, without mask of gradient signal.
\end{enumerate}

\end{remark}

\begin{proposition}\label{prop:IPPRO_justification}
Assume that the step size $\lambda\ll 1$ and the loss function is Lipschitz continuous near pretrained parameter. Then
    \begin{equation}
        PROscore_\lambda (i)\approx1+\lambda\frac{I^{id}_i(\mathcal{D})}{\|F_i\|_2}-\lambda\frac{I_i^\sigma(\mathcal{D})}{\|F_i\|_2^2}+O(\frac{\lambda^2}{\|F_i\|_2^2})
    \end{equation}

\begin{proof}
Recall 
\begin{equation}
PROscore_\lambda (i)    = \frac{\|F_i - \lambda\nabla_{F_i}\mathcal{L}\|_2}{|D_{ii}-\lambda\frac{\partial \mathcal{L}}{\partial D_{ii}}|}. 
\end{equation}

The numerator $\|F_i - \lambda\nabla_{F_i}\mathcal{L}\|_2$ can be approximated by 1st order Taylor approximation:

\begin{equation}
    \|F_i-\lambda\nabla_{F_i}\mathcal{L}\|_2 = \|F_i\|_2-\lambda\frac{F\cdot \nabla_F\mathcal{L}}{\|F_i\|_2}+O(\frac{\lambda^2}{\|F_i\|_2
    })
\end{equation}

From backpropagation, we have $\frac{\partial \mathcal{L}}{\partial D_{ii}} = \sum_{x\in \mathcal{D}}\frac{\partial L}{\partial a_{i,x}}h_{i,x}$ since $a_{i,x}=D_{ii}h_{i,x}-\overline{D}_{ii}h_{i,x}+\sigma(h_{i,x})$ at the initialization and thus $\frac{da_{i,x}}{dD_{ii}}=h_{i,x}$.

Therefore, $\frac{\partial \mathcal{L}}{\partial D_{ii}}$ is equal to $I_i^{id}$ and we can rewrite PROscore:

\begin{equation}
    PROscore_\lambda(i)\approx \frac{\|F_i\|_2-\lambda\frac{I^\sigma_i(\mathcal{D})}{\|F\|_2}}{\|F_i\|_2-\lambda I^{id}_i(\mathcal{D})}=\frac{1-\lambda\frac{I_i^\sigma}{\|F_i\|_2^2}}{1-\lambda\frac{I_i^{id}}{\|F_i\|_2}}
\end{equation}

Since $\lambda\ll1$, we can further approximate using $\frac{1-\alpha}{1-\beta}\approx 1+\beta-\alpha$:
\begin{equation}
    PROscore_\lambda(i)\approx1+\lambda\frac{I_i^{id}(\mathcal{D})}{\|F_i\|_2}-\lambda\frac{I_i^\sigma(\mathcal{D})}{\|F_i\|_2^2}+O(\frac{\lambda^2}{\|F_i\|_2^2}).
\end{equation}

\end{proof}
\end{proposition}

\begin{remark}[Remarks on \cref{prop:IPPRO_justification}]

    \begin{itemize}
        \item The error term $O(\frac{\lambda^2}{\|F\|_2^2})$ is negligible in practice and thus $\frac{I_i^{id}}{\|F_i\|_2}-\frac{I_i^\sigma}{\|F_i\|_2^2}$ approximates $PROscore$. 
        For example, on DeiT-T with $\lambda=0.1$, the rank correlation between $PROscore_\lambda$ and $\frac{I_i^{id}}{\|F_i\|_2}-\frac{I_i^\sigma}{\|F_i\|_2^2}$ is larger than 0.995 for all mlp layers.
        \item In practice, for example with DeiT-T, $\frac{I_i^{id}}{\|F_i\|_2}$ dominates the pruning decision; since $Var(\frac{I_i^{id}}{\|F_i\|_2})$ is about 11x-3500x times (depend on layers) larger than $Var(\frac{I_i^{\sigma}}{\|F_i\|_2^2})$.
    \end{itemize}
\end{remark}

\paragraph{Spurious filters}\label{appendix:spurious}

In the field of deep representation learning, \textbf{spurious neurons} refer to internal components of a model that encode non-causal, superficial correlations present within a specific training distribution. 
These neurons are the primary drivers of shortcut learning, a phenomenon wherein a neural network achieves high predictive accuracy by exploiting "easy" environmental cues—such as background textures, lighting conditions, or co-occurring context—rather than learning the robust, intrinsic features of the target class\cite{geirhos2020shortcut}. For instance, a model may mistakenly rely on neurons that detect "Vivarium" to identify "snake," a dependency that fails when the subject is encountered in an atypical setting\cite{Sagawa2020Distributionally}.

\begin{figure*}[t]
    \centering
    \begin{subfigure}{0.2\textwidth}
        \includegraphics[width=\textwidth]{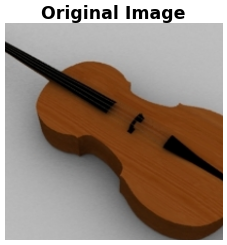}
        \vspace{2mm}
        \caption{Input image}
        \label{fig:raw_624616}
    \end{subfigure}
    \begin{subfigure}[b]{0.38\textwidth}
        \includegraphics[width=\textwidth]{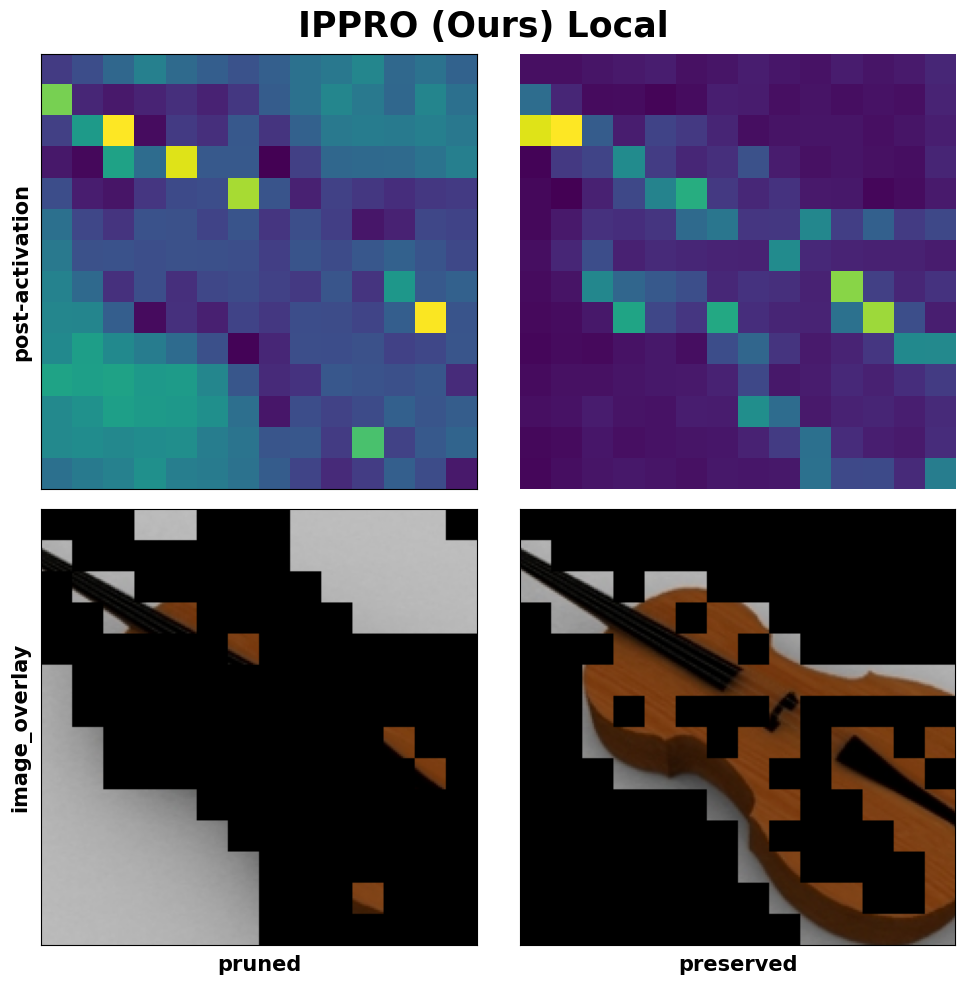}
        \caption{local $l_{ippro}(x)$}
        \label{fig:IPPRO_local_624616}
    \end{subfigure}
    \begin{subfigure}[b]{0.38\textwidth}
        \includegraphics[width=\textwidth]{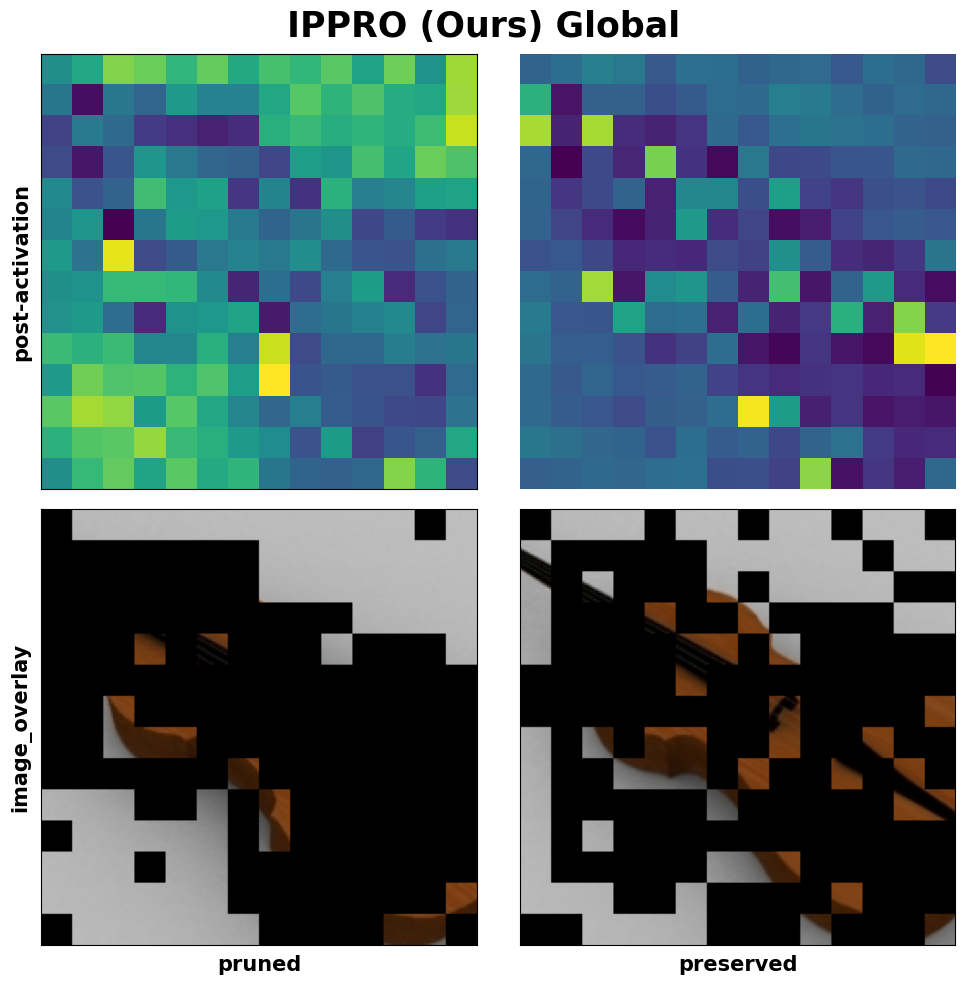}
        \caption{global $PROscore_\lambda$}
        \label{fig:IPPRO_global_624616}
    \end{subfigure}
    \caption{Activation map and image overlay of top/bottom 10 filters for local and global criteria, on DeiT-T model's first mlp layer. \cref{fig:IPPRO_local_624616} shows result under sample-wise local criteria $\frac{\partial L}{\partial a_{i,x}}h_{i,x}$ and \cref{fig:IPPRO_global_624616} shows results under global criteria, $PROscore_\lambda(i)$.}
    \label{fig:spu_local_global_IPPRO}
\end{figure*}

Unlike robust, causal features that capture the fundamental structure of an object, spurious neurons capture "extrinsic" associations that do not necessarily represent the functional contribution of a filter to the model's logic. Consequently, the reliance on these components inherently limits the model's ability to generalize to out-of-distribution (OOD) scenarios and compromises its structural integrity \cite{beery2018recognition}. 

Because these spurious neurons often achieve high activations or possess significant parameter magnitudes on biased training sets, they are frequently misidentified as essential features by traditional importance-based metrics such as \cite{berisha2025variance} or magnitude.

\paragraph{IPPRO is pruning spurious filters}

As stated in \cref{prop:IPPRO_justification}, IPPRO decision is closely related to $\frac{I_i^{id}(\mathcal{D})}{\|F_i\|_2} = \sum_{x\in \mathcal{D}}\frac{F_i}{\|F_i\|}\cdot\frac{\partial L}{a_{i,x}}x$. 
This term represents the alignment between the filter and the "linearized" error signal of the network.

The sample-wise local signal $l_{ippro}(x)=\frac{\partial L}{a_{i,x}}h_{i,x}$ serves as a strict measure of directional alignment between a neuron's linear projection and the network's optimization trajectory.

Robust core features, which capture the intrinsic shape of an object, consistently align with the loss-minimizing direction across diverse samples, thereby accumulating high importance scores. 
In contrast, spurious features that exploit coincidental environmental shortcuts—such as background textures—often induce destructive interference with the loss gradient, especially when the spurious condition is disrupted, resulting in heavily penalized (lowest) scores. 

To empirically validate this theoretical connection, we visualized the spatial activations of neurons ranked by this metric in \cref{fig:spu_local_global_IPPRO}. 
As illustrated in \cref{fig:IPPRO_local_624616}, the bottom 10 neurons designated for pruning (lowest scores) exhibit diffuse activation patterns that explicitly highlight task-irrelevant background regions. 
Conversely, the top 10 neurons selected for preservation tightly localize on the core foreground object (e.g., the structural details of the violin).

Inherently, global $PROscore$ (the accumulation of $l_{ippro}(x)$) also captures the spurious filter and select to prune it as visualized in \cref{fig:IPPRO_global_624616}

These results clearly demonstrate that our metric inherently disentangles spurious contextual biases from robust semantic representations without requiring any explicit spatial annotations or bounding boxes.

\paragraph{Comparison to other methods}

\begin{figure*}[t]
    \centering
    \begin{subfigure}{0.1\textwidth}
        \includegraphics[width=\textwidth]{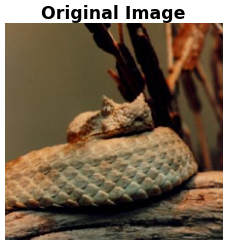}
        \vspace{2mm}
        \caption{Input}
        \label{fig:raw_128694}
    \end{subfigure}
    \begin{subfigure}[b]{0.2\textwidth}
        \includegraphics[width=\textwidth]{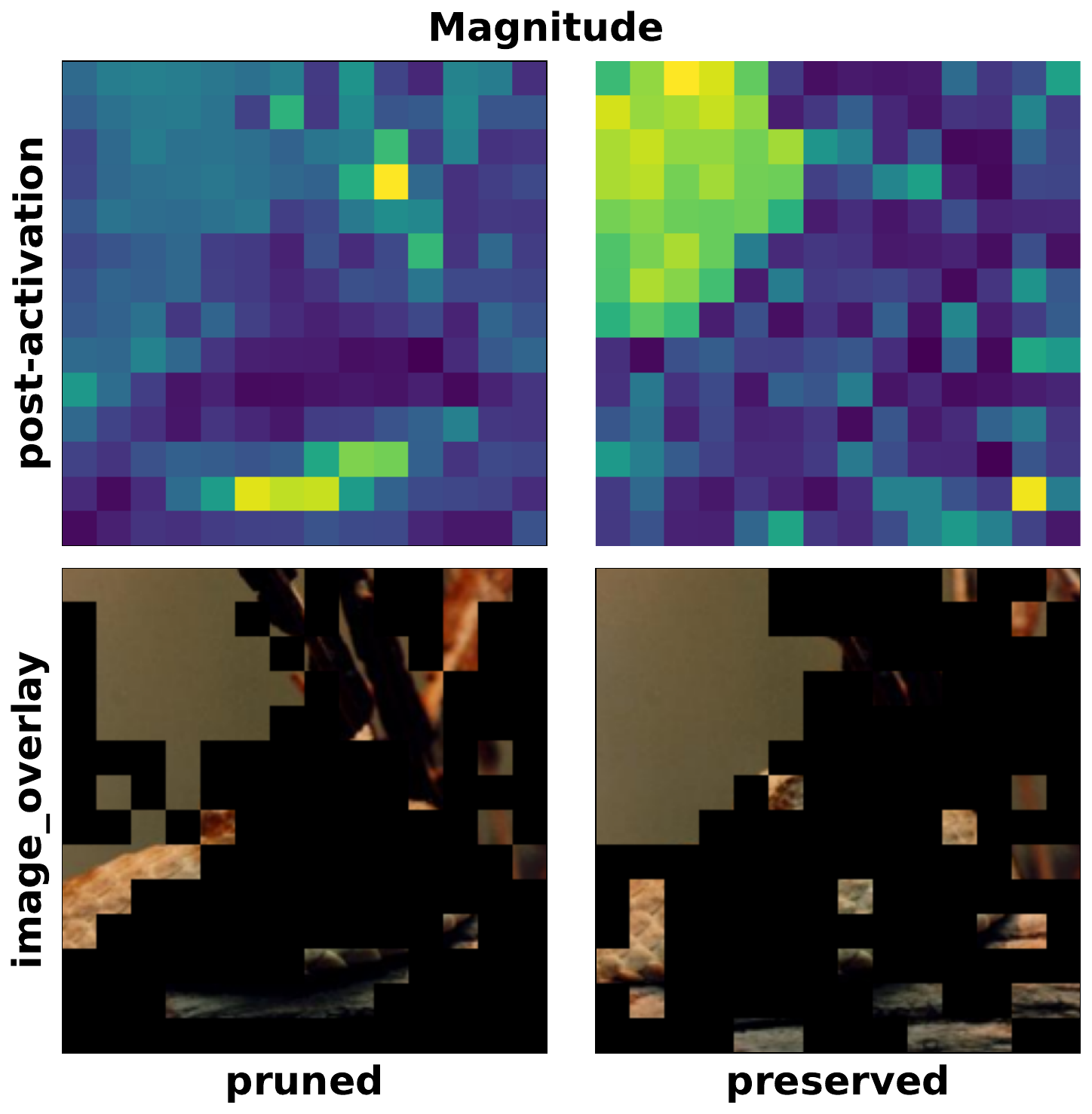}
        \caption{Magnitude}
        \label{fig:spu_mag_128694}
    \end{subfigure}
    \begin{subfigure}[b]{0.2\textwidth}
        \includegraphics[width=\textwidth]{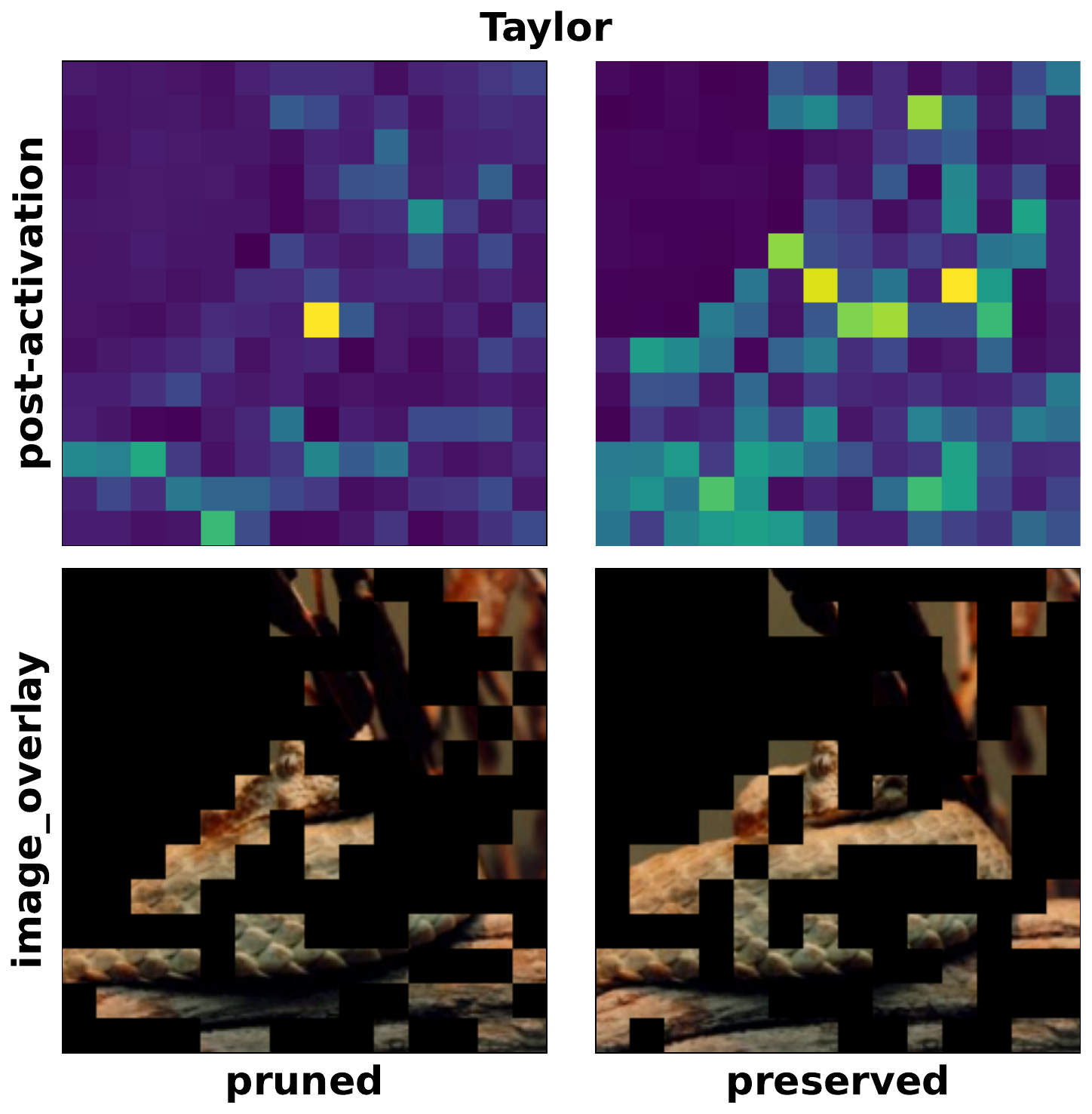}
        \caption{Taylor}
        \label{fig:spu_taylor_128694}
    \end{subfigure}
    \begin{subfigure}[b]{0.2\textwidth}
        \includegraphics[width=\textwidth]{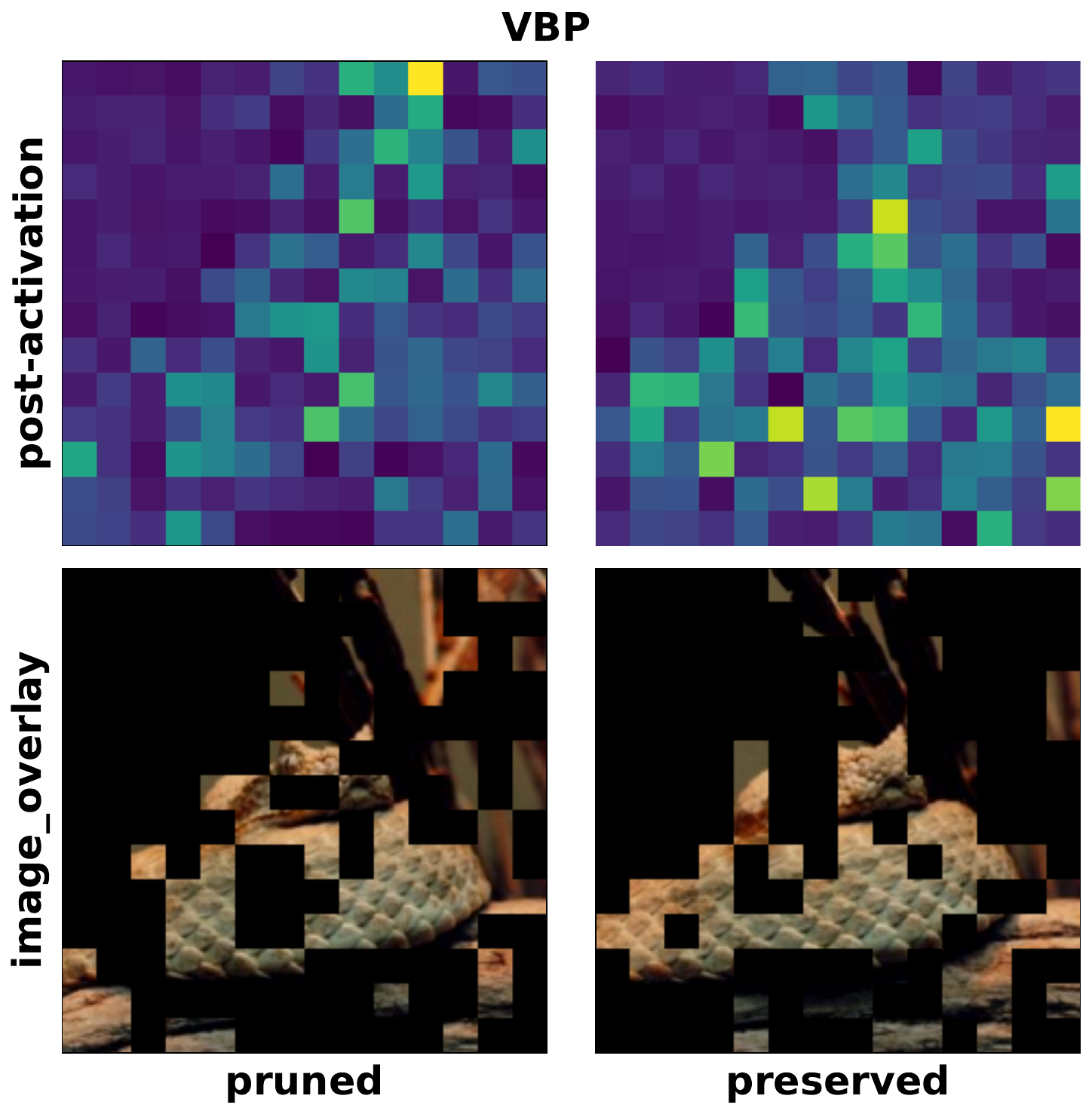}
        \caption{VBP}
        \label{fig:spu_vbp_128694}
    \end{subfigure}
    \begin{subfigure}[b]{0.2\textwidth}
        \includegraphics[width=\textwidth]{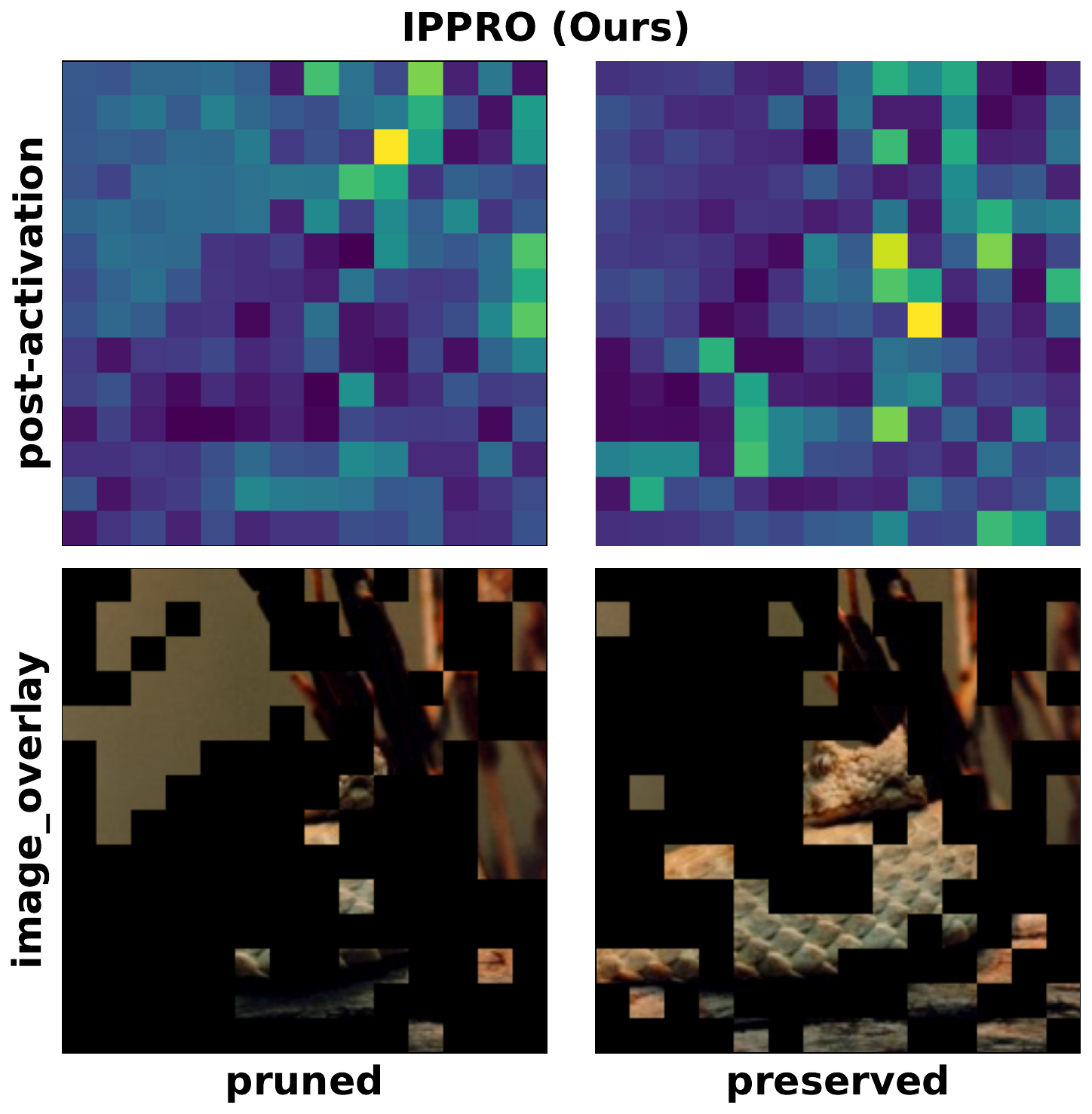}
        \caption{IPPRO}
        \label{fig:spu_mag_128694}
    \end{subfigure}    
    \caption{Comparison of activation map and image overlay from top/bottom 10 filters for local and global criteria, on DeiT-T model's first mlp layer (snake input).}
    \label{fig:spu_local_global_IPPRO}
\end{figure*}

\begin{figure*}[t]
    \centering
    \begin{subfigure}{0.1\textwidth}
        \includegraphics[width=\textwidth]{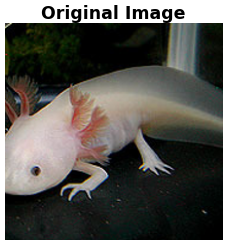}
        \vspace{2mm}
        \caption{Input}
        \label{fig:raw_38465}
    \end{subfigure}
    \begin{subfigure}[b]{0.2\textwidth}
        \includegraphics[width=\textwidth]{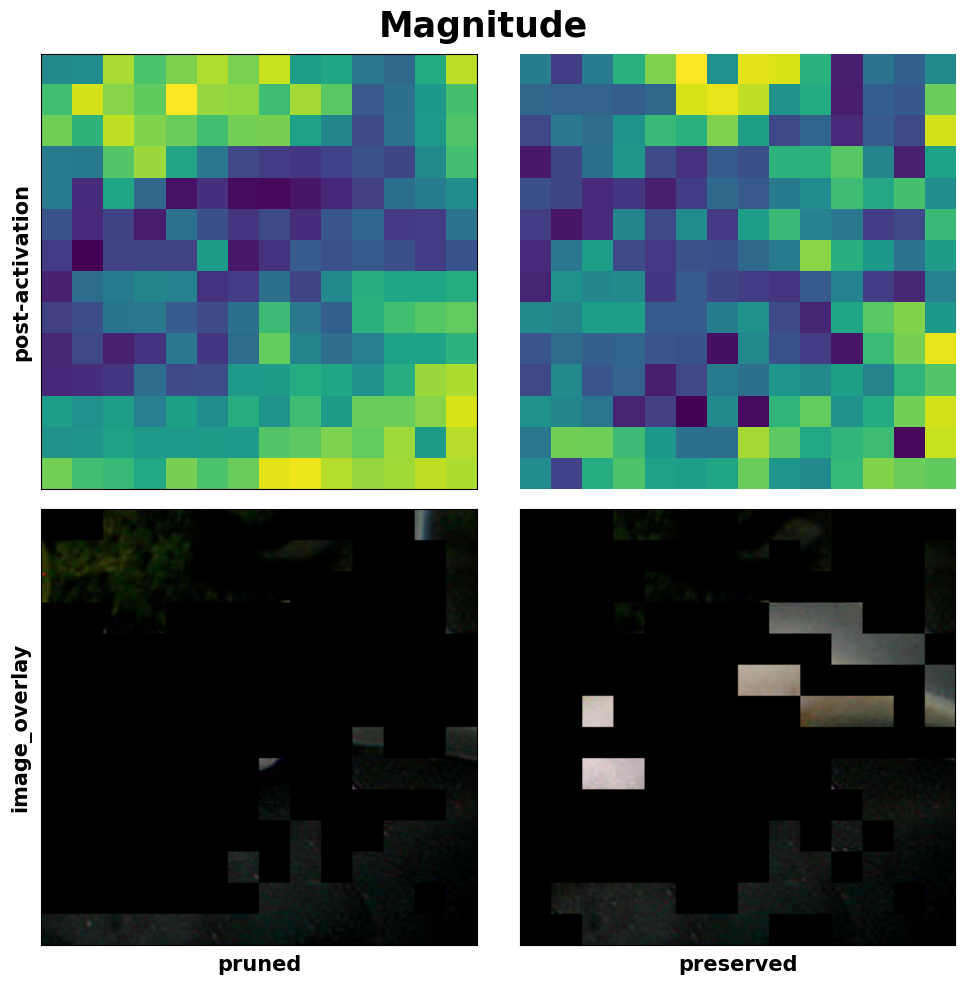}
        \caption{Magnitude}
        \label{fig:spu_mag_38465}
    \end{subfigure}
    \begin{subfigure}[b]{0.2\textwidth}
        \includegraphics[width=\textwidth]{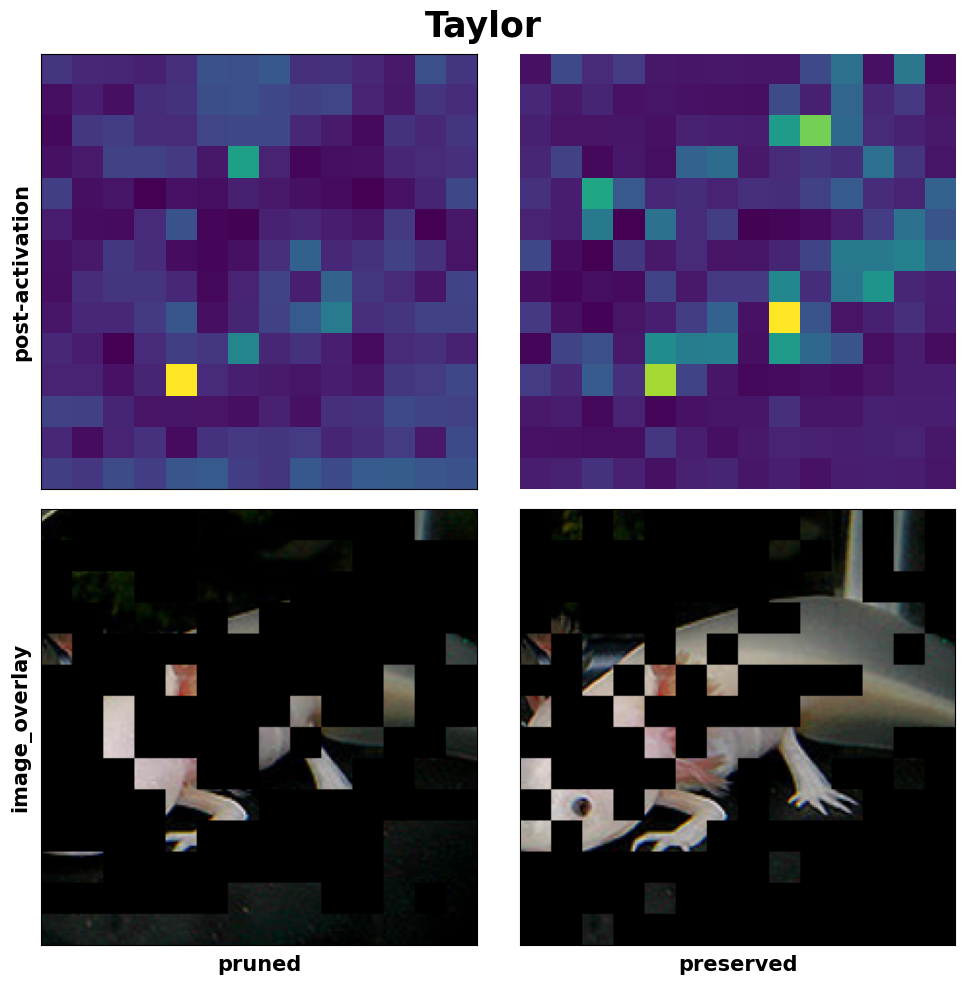}
        \caption{Taylor}
        \label{fig:spu_taylor_38465}
    \end{subfigure}
    \begin{subfigure}[b]{0.2\textwidth}
        \includegraphics[width=\textwidth]{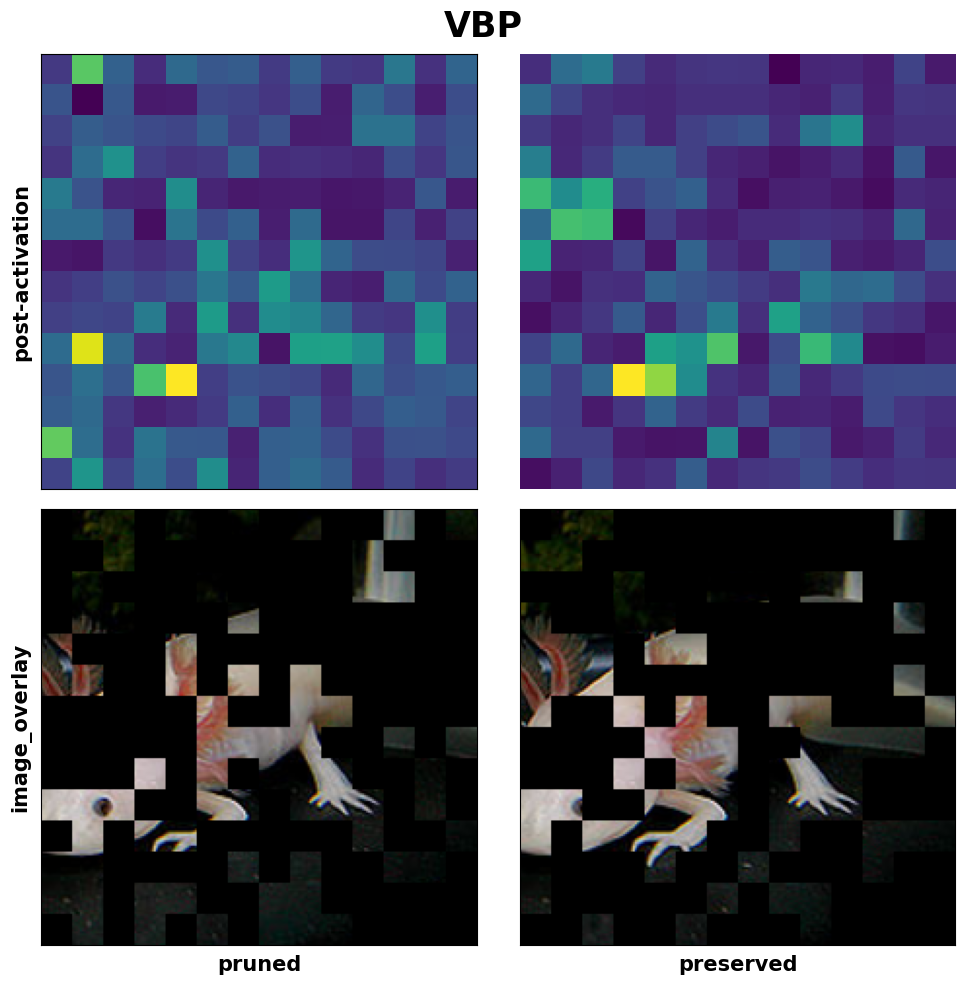}
        \caption{VBP}
        \label{fig:spu_vbp_38465}
    \end{subfigure}
    \begin{subfigure}[b]{0.2\textwidth}
        \includegraphics[width=\textwidth]{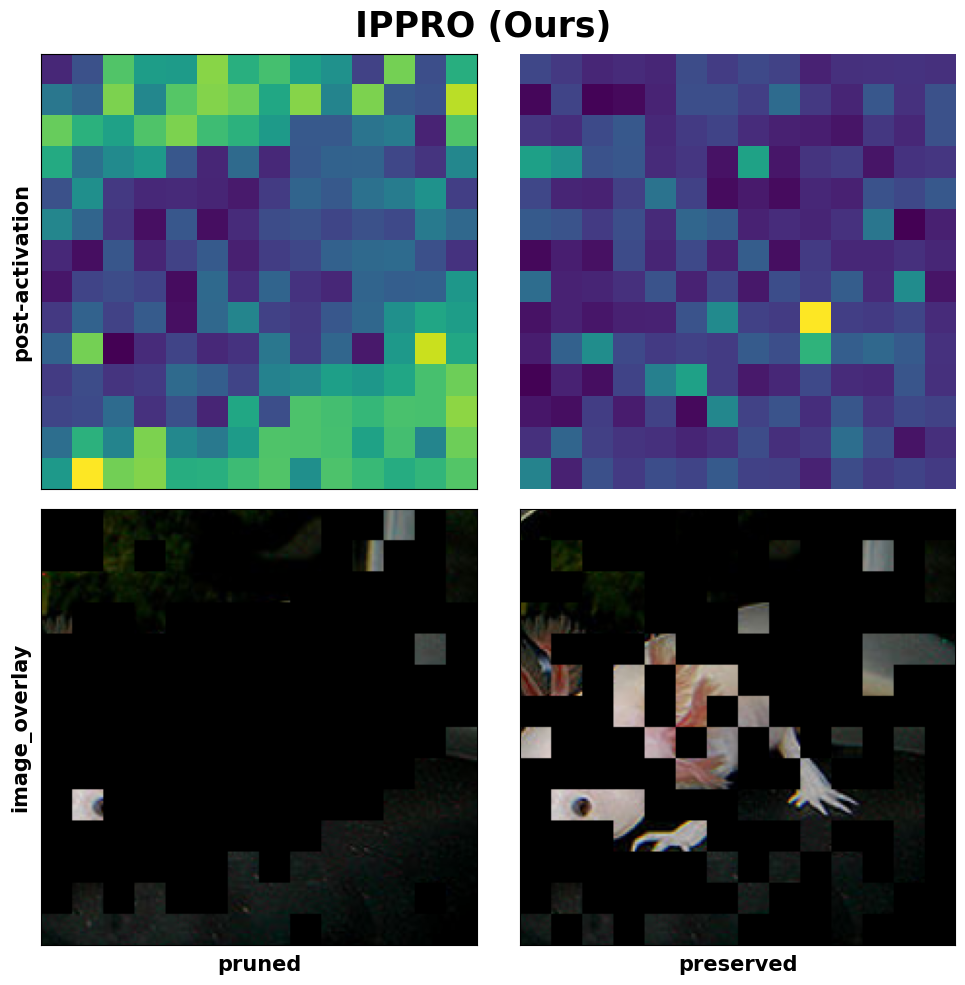}
        \caption{IPPRO}
        \label{fig:spu_mag_38465}
    \end{subfigure}
    
    \caption{Comparison of activation map and image overlay from top/bottom 10 filters for local and global criteria, on DeiT-T model's first mlp layer (axolotl input).}
    \label{fig:spu_local_global_IPPRO}
\end{figure*}

We compare the activation map of pruning decision against the importance-based pruning baselines. While IPPRO prunes spurious filters which are activated on background but retaining filters which activate on important feature, other methods choose pruning filters regardless robustness of the filter. In detail, magnitude pruning prunes spurious filter but also preserves; and other two are preserving robust filter but also prunes them.

Pruning robust filter does not directly implies the model collapse, since there could be another filters with similar role (consider parallel filter but with different magnitude) that are not pruned. 
However, spurious filters are redundant regardless of the existence similar one; as they acts as a noise in forward process. Therefore, IPPRO's preference on pruning spurious filters support the empirical success of IPPRO on various benchmarks. 

\section{MHA Projective Initialization Details}\label{app:mha_adaptation}
As mentioned in \cref{subsec:attn_pruning}, extending the identity operation after the QKV projection requires initializing the auxiliary parameter $D$ such that the augmented filter $(D_i, F_i)$ resides exactly on the projective cone $B_{\pi/4}(0)$, while leaving the output of the Q, K, and V projections functionally unaltered.

To achieve this, we introduce a multiplier $m_i$ for each filter $i$. We seek an $m_i$ that satisfies the condition:
\begin{equation}
    \frac{1}{m_i} - 1 = m_i\|F_i\|
\end{equation}
Solving for $m_i$ allows us to balance the scaling. We then rescale the original filter weights as $F_i \leftarrow m_i F_i$ and initialize the auxiliary parameter as $D_i \leftarrow \frac{1}{m_i} - 1$. This mathematically guarantees the projective embedding property required for valid PROscore computation without modifying the original attention mechanism's forward behavior.

\section{Additional Performance Analysis}\label{appendix:additional_performace}
\subsection{CIFAR-10 and CIFAR-100}\label{appendix:additional_cifar}

We evaluated IPPRO on CIFAR-10 (ResNet-56) and CIFAR-100 (VGG19) using DepGraph pre-trained weights \cite{fang2023depgraph}. As shown in \cref{tab:cifar_results}, IPPRO reduces FLOPs and parameters by up to 49\% on CIFAR-10 with a 0.53\% accuracy gain. Even at higher compression rates (up to 77\%), accuracy drops by only ~1\%, outperforming other methods. On CIFAR-100, IPPRO achieves 87\% FLOPs and 87\% parameter reduction with minimal accuracy loss, highlighting its effectiveness for compact, high-performing models.

\begin{table*}[!ht]
    \centering
        \centering
        \caption{CIFAR-10 and CIFAR-100}\label{tab:cifar_results}
        \resizebox{0.7\linewidth}{!}{
        \begin{tabular}{llccccc}
            \toprule
            \multirow{2.5}{*}{\makecell{Model \\ Dataset}} & \multirow{2.5}{*}{Method} & \multicolumn{3}{c}{Top-1 Acc (\%)} & \multirow{2.5}{*}{\makecell{Params$\downarrow$ \\ (\%)}} & \multirow{2.5}{*}{\makecell{FLOPs$\downarrow$ \\ (\%)}} \\
            \cmidrule(lr){3-5}
            & & Base. & Prun. & $\Delta$ \\
            \midrule
            \multirow{14}{*}{\makecell{ResNet-56 \\ CIFAR-10}} 
            & Depgraph\cite{fang2023depgraph} & 93.53 & 93.77 & +0.24 & - & - \\
            & HBFP\cite{basha2024deep} & 93.26 & 92.42 & -0.84 & \underline{43.9} & 43.6 \\
            & GAL\cite{lin2019towards} & 93.26 & 93.38 & +0.12 & 12.3 & 37.6 \\
            & HRank\cite{lin2020hrank} & 93.26 & 93.52 & +0.26 & 17.0 & 29.3 \\
            & CHIP\cite{sui2021chip} & 93.26 & \textbf{94.16} & \textbf{+0.90} & \underline{43.9} & \underline{47.4} \\
            & FPAC\cite{yang2023filter} & 93.26 & 93.71 & +0.45 & 42.8 & \underline{47.4} \\
        & \textbf{IPPRO (ours)} & 93.53 & \underline{94.06} & \underline{+0.53} & \textbf{49.7} & \textbf{49.3} \\
            \cmidrule{2-7}
            & HRank\cite{lin2020hrank} & 93.26 & 90.72 & -2.54 & 68.4 & \underline{74.1} \\
            & HBFP\cite{basha2024deep} & 93.26 & 91.79 & -1.47 & \underline{75.4} & \textbf{74.9} \\
            & GAL\cite{lin2019towards} & 93.26 & 91.58 & -1.68 & 66.1 & 60.2 \\
            & CHIP\cite{sui2021chip} & 93.26 & \underline{92.05} & \underline{-1.21} & 71.8 & 72.3 \\ 
            & SPSRC\cite{sun2024towards} & 93.59 & 91.65 & -1.94 & 64.7 & 63.9 \\
            & \textbf{IPPRO (ours)} & 93.53 & \textbf{92.47} & \textbf{-1.06} & \textbf{77.7} & 71.8 \\
            \midrule
            \multirow{7}{*}{\makecell{VGG19 \\ CIFAR-100}} 
        & Depgraph\cite{fang2023depgraph} & 73.50 & \underline{70.39} & \underline{-3.11} & - & \underline{88.7} \\
            & Kron-OBS\cite{wang2019eigendamage} & 73.34 & 60.66 & -12.6 & - & 83.5 \\
            & Greg-2\cite{wang2020greg} & 74.02 & 67.75 & -6.27 & - & 88.6 \\
            & EigenD\cite{wang2019eigendamage} & 73.34 & 65.18 & -8.16 & 90.9 & 88.6 \\
            & Torque\cite{10484493} & 73.03 & 65.87 & -7.16 & 90.8 & \underline{88.7} \\
            & GAT Transpruning\cite{lin2024gat} & 73.26 & 66.68 & -6.58 & - & \textbf{89.0} \\
            & \textbf{IPPRO (ours)} & 73.50 & \textbf{70.47} & \textbf{-3.03} & 87.9 & 87.5 \\
            \bottomrule
        \end{tabular}
        }
\end{table*}

\subsection{ImageNet-1K}\label{appendix:additional_resnet50}

\begin{table*}[!ht]
    \centering
    \caption{Comparison of pruning performance on ImageNet dataset}\label{tab:Imagenet_results}
    \resizebox{0.9\linewidth}{!}{
    \begin{tabular}{clcccccccc}
        \toprule
        \multirow{2.5}{*}{Model} & \multirow{2.5}{*}{Method} & \multicolumn{3}{c}{Top-1 Acc (\%)} & \multicolumn{3}{c}{Top-5 Acc (\%)} & \multirow{2.5}{*}{Params$\downarrow$(\%)} & \multirow{2.5}{*}{FLOPs$\downarrow$(\%)} \\ 
        \cmidrule(lr){3-5} \cmidrule(lr){6-8}
        & & Base. & Prun. & $\Delta$ & Base. & Prun. & $\Delta$ & & \\ 
        \midrule
        \multirow{22}{*}{ResNet-50} 
        & SFP \cite{he2018soft} & 76.15 & 74.61 & -1.54 & 92.87 & 92.06 & -0.81 & N/A & 41.8 \\ 
        & Autopruner \cite{luo2020autopruner} & 76.15 & 74.76 & -1.39 & 92.87 & 92.15 & -0.72 & N/A & 48.7 \\ 
        & FPGM \cite{he2019filter} & 76.15 & 75.59 & -0.56 & 92.87 & 92.63 & -0.24 & 37.5 & 42.2 \\ 
        & Taylor \cite{molchanov2019importance} & 76.18 & 74.50 & -1.68 & N/A & N/A & N/A & 44.5 & 44.9 \\  
        & GAL \cite{lin2019towards} & 76.15 & 71.95 & -4.20 & 92.87 & 90.94 & -1.93 & 16.9 & 43.0 \\   
        & HRank \cite{lin2020hrank} & 76.15 & 74.98 & -1.17 & 92.87 & 92.33 & -0.54 & 36.6 & 43.7 \\
        & SCOP \cite{tang2020scop} & 76.15 & 75.95 & -0.20 & 92.87 & 92.79 & -0.08 & 42.8 & 45.3 \\
        & CHIP \cite{sui2021chip} & 76.15 & \underline{76.15} & \underline{0.00} & 92.87 & \underline{92.91} & \underline{+0.04} & 44.2 & \underline{48.7} \\
        & RGP\cite{chen2023rgp} & 76.22 & 75.30 & -0.92 & N/A & N/A & N/A & 43.8 & 43.8 \\
        & FPBICI\cite{tang2024filter} & 76.13 & 76.08 & -0.05 & 92.86 & 92.85 & -0.01 & \underline{45.9} & \textbf{50.4} \\
        & \textbf{IPPRO (ours)} & 76.15 & \textbf{76.21} & \textbf{+0.06} & 92.87 & \textbf{93.02} & \textbf{+0.15 }& \textbf{46.4} & \textbf{50.4} \\
        \cline{2-10}
        & SCOP \cite{tang2020scop} & 76.15 & \underline{75.26} & \underline{-0.89} & 92.87 & 92.53 & -0.34 & 51.8 & 54.6 \\ 
        & SIRFP \cite{wu2025structural} & 76.15 & 75.14 & -1.01 & 92.87 & \textbf{93.12} & \textbf{+0.25} & N/A & 58.7 \\
        & CHIP \cite{sui2021chip} & 76.15 & \underline{75.26} & \underline{-0.89} & 92.87 & 92.53 & -0.34 & 56.7 & \underline{62.8} \\ 
        & Torque \cite{10484493} & 76.07 & 74.58 & -1.49 & N/A & N/A & N/A & \textbf{64.5} & 57.2\\
        & FPBICI\cite{tang2024filter} & 76.13 & 75.01 & -1.12 & 92.86 & 92.30 & -0.56 & 57.7 & \textbf{63.8} \\
        & \textbf{IPPRO (ours)} & 76.15 & \textbf{75.51} & \textbf{-0.64} & 92.87 & \underline{92.67} & \underline{-0.20} & \underline{60.0} & 60.0 \\
        \cline{2-10}
        & HRank \cite{lin2020hrank} & 76.15 & 69.10 & -7.05 & 92.87 & 89.58 & -3.29 & 67.5 & \underline{76.0} \\ 
        & CHIP \cite{sui2021chip} & 76.15 & \underline{73.30} & \underline{-2.85} & 92.87 & \underline{91.48} & \underline{-1.39} & 68.6 & \textbf{76.7} \\
        & RGP \cite{chen2023rgp} & 76.22 & 72.68 & -3.54 & N/A & N/A & N/A & \textbf{75.0} & 75.0 \\
        & \textbf{IPPRO (ours)} & 76.15 & \textbf{73.38} & \textbf{-2.77} & 92.87 & \textbf{91.50} & \textbf{-1.37} & \underline{74.2} & 74.2 \\
        \midrule
        \multirow{3}{*}{MobileNetV2}
        & Meta \cite{liu2019metapruning} & 74.70 & \underline{68.20} & \underline{-6.50} & N/A & N/A & N/A & N/A & \textbf{54.2} \\
        & GFP \cite{liu2021group} & 75.74 & \textbf{69.16} & -6.58 & N/A & N/A & N/A & N/A & \underline{50.5} \\
        & \textbf{IPPRO (ours)} & 72.01 & 67.90 & \textbf{-4.11} & 90.62 & 88.04 &  -2.58 &  42.3 & \textbf{54.2} \\
        
        \bottomrule
    \end{tabular}
    }
\end{table*}

On ImageNet-1K, we evaluate IPPRO with ResNet-50 under three compression levels (50.4\%, 60\%, and 74.2\% FLOPs reduction). As shown in \cref{tab:Imagenet_results}, IPPRO improves Top-1 accuracy by 0.06\% at moderate pruning and remains competitive at higher compression, outperforming state-of-the-art methods. Furthermore, for other FLOP reductions on 60\% and 74\%, IPPRO exhibited significantly higher accuracy compared to other superior methods.

We further validate IPPRO on MobileNetV2, a compact model without residual connections, where it shows the smallest accuracy drop at comparable FLOPs, demonstrating architectural versatility.

\subsection{Global Pruning}\label{sec:global}

We define our importance score using angular deviation in the projective space, measuring how much each filter responds to data relative to a reference value of $\tan(\frac{\pi}{4})$. Although the absolute score can vary with the hyperparameter $\lambda$, PROscore enables global comparability without explicit normalization, unlike magnitude-based methods. This property is essential for global pruning, where importance scores must reside on a comparable scale to ensure fair competition among filters across layers. Without this, layers with intrinsically larger scores may be unjustly favored. To verify that score scaling induced by \cref{eq:proscore} $\lambda$ does not undermine pruning behavior, we analyze its effect and confirm the stability of our method in \cref{lamda}.

By leveraging this robust scoring scheme, our global pruning method consistently outperforms conventional magnitude-based global pruning. As shown in \cref{tab:global_pruning}, it also achieves performance on par with other advanced approaches specifically designed to address the shortcomings of norm-based criteria. 

\setlength{\textfloatsep}{10pt}
\setlength{\floatsep}{10pt}
\setlength{\intextsep}{10pt}
\setlength{\dbltextfloatsep}{10pt}
\setlength{\dblfloatsep}{10pt}
\begin{table*}[h]
    \caption{Global Pruning Performance results of ImageNet and CIFAR-10 datasets. }\label{tab:global_pruning}
    \centering
    \begin{subtable}[t]{0.49\linewidth}
        \centering
        \caption{ImageNet-1K on ResNet-50}\label{tab:global_imagenet_results}
        \resizebox{\linewidth}{!}{
        \begin{tabular}{lcccc}
            \toprule
            \multirow{2.5}{*}{Method} & \multicolumn{3}{c}{Top-1 Acc (\%)} & \multirow{2.5}{*}{\makecell{Remain \\ FLOPs(G)}} \\
            \cmidrule(lr){2-4}
            & Base. & Prun. & $\Delta$ \\
            \midrule
            ResConv-Prune\cite{xu2020layer} & 76.2 & 70.0 & -6.2 & 1.6  \\
            DBP-0.5\cite{wang2019dbp} & 76.2 & 72.4 & -3.8 & N/A  \\
            Meta\cite{liu2019metapruning} & 76.2 & 73.4 & -2.8 & \textbf{1.0} \\
            AutoSlim\cite{yu2019autoslim} & 76.2 & 74.0 & -2.2 & \textbf{1.0} \\
            GReg-2\cite{wang2020greg} & 76.2 & 73.9 & -2.3 & 1.3 \\
            HALP\cite{shen2022structural} & 76.2 & \underline{74.5} & \underline{-1.7} & \underline{1.2} \\
            \textbf{IPPRO (ours)} & 76.2 & \textbf{74.6} & \textbf{-1.6} & 1.3 \\
            \midrule
            HALP\cite{shen2022structural} & 76.2 & 68.1 & -8.1 & \underline{0.6} \\
            MDP \cite{sun2024multi} & 76.2 & \textbf{70.0} &\textbf{ -6.2} & \textbf{0.5} \\
            \textbf{IPPRO (ours)} & 76.2 & \underline{69.9} & \underline{-6.3} & \underline{0.6} \\
            \bottomrule
        \end{tabular}
        }
    \end{subtable}
    \hfill
    \begin{subtable}[t]{0.49\linewidth}
        \centering
        \caption{CIFAR-10 on ResNet-56}\label{tab:global_cifar_results}
        \resizebox{\linewidth}{!}{
        \begin{tabular}{lcccc}
            \toprule
            \multirow{2.5}{*}{Method} & \multicolumn{3}{c}{Top-1 Acc (\%)} & \multirow{2.5}{*}{\makecell{Remain \\ FLOPs(M)}} \\
            \cmidrule(lr){2-4}
            & Base. & Prun. & $\Delta$ \\
            \midrule
            FSM\cite{duan2022network} & 93.26 & 93.63 & \underline{+0.4} & 61.17 \\
            ITFCP\cite{chen2024effective} &93.39 & 93.60 & +0.21 & 60.73 \\
            GCNNA\cite{jiang2022channel} & 93.72 & \underline{93.72} & 0 & \textbf{58.29}\\
            FSIM\cite{liu2023filter} & 93.30 & 93.48 & +0.18 & \underline{59.24}\\
            \textbf{IPPRO (ours)} & 93.53 & \textbf{94.00} & \textbf{+0.47} & 63.66 \\
            \midrule
            QSFM\cite{wang2022qsfm} & 93.21 & 91.88 & \underline{-1.33} & 50.62\\
            GBN\cite{you2019gate} & 93.10 & 91.76 & -1.34 & 40.23\\
            FSIM\cite{liu2023filter} & 93.30 & \underline{91.96} & -1.34 & \textbf{31.08}\\
            \textbf{IPPRO (ours)} & 93.53 & \textbf{92.43} & \textbf{-1.10} & \underline{37.49} \\
            \bottomrule
        \end{tabular}
        }
    \end{subtable}
\end{table*}

\subsection{Latency Test of Large Language Models}\label{appendix:latency}

\cref{tab:latency_llm} reports inference latency for LLaMA-7B and LLaMA2-7B with and without IPPRO. Latency is measured on an RTX 4090 GPU with 200 warm-up runs and averaged over 1,000 runs (batch size 64). As shown in \cref{tab:latency_llm}, IPPRO consistently reduces inference latency and computational cost for both models, at the expense of reduced average accuracy, particularly for LLaMA2-7B. These results indicate that IPPRO can extend efficiency gains beyond vision models to large language models.

\begin{table*}[h]
    \caption{Latency comparison of LLMs}\label{tab:latency_llm}
    \centering
    \resizebox{0.7\linewidth}{!}{
        \begin{tabular}{lccccc}
        \toprule
        Model & Latency(s) & Params(B) & MACs(G) & Avg. Acc & Acc Diff \\
        \midrule
        LLAMA-7B & 43.27(1.0x) & 6.74 & 425.1 & 63.17 & 0 \\
        + IPPRO (Ours) & 36.34(1.2x) & 5.12 & 321.3 & 60.94 & -2.23 \\
        \midrule
        LLAMA2-7B & 42.60(1.0x) & 6.74 & 422.8 & 64.6 & 0 \\
        + IPPRO (Ours) & 20.92(2.0x) & 3.28 & 201.5 & 48.0 & -16.6 \\
        \bottomrule
        \end{tabular}
    }
\end{table*}

\subsection{Results Without Fine-tuning of ViTs}\label{appendix:wo_ft_vit}

To examine whether IPPRO relies heavily on fine-tuning for maintaining performance in DeiT models, we compare accuracy before and after fine-tuning against VBP\cite{berisha2025variance} and SNP\cite{shim2024snp}, as reported in \cref{tab:vit_results_wo_ft}. For a fair comparison with VBP, pruning is applied only to the MLP layers of the DeiT models, and we report the performance after the 10-epoch distillation fine-tuning protocol used in the original VBP work. IPPRO achieves better accuracy than VBP both with and without fine-tuning; notably, on DeiT-Tiny, IPPRO without fine-tuning outperforms VBP by approximately 5\%.

For the comparison with SNP, pruning is applied to both the Attention and MLP layers, and fine-tuning follows the original training recipe without distillation for 300 epochs. In this setting, IPPRO shows a substantial advantage, with the gap reaching nearly 30× in the no fine-tuning scenario. These results indicate that our method preserves model performance significantly better than prior approaches, even immediately after pruning without any fine-tuning.

\begin{table}[H]
    \centering
    \caption{DeiT pruning performance without fine-tuning on ImageNet-1K. Methods are grouped by pruning scope: MLP-only (vs VBP) or MLP + Attention (vs SNP).}\label{tab:vit_results_wo_ft}
    \resizebox{0.9\linewidth}{!}{
        \begin{tabular}{llccccc}
        \toprule
        \multirow{2.5}{*}{Model} & \multirow{2.5}{*}{Method}  & \multicolumn{3}{c}{Acc($\uparrow$\%)} & \multirow{2.5}{*}{Params$\downarrow$(\%)} & \multirow{2.5}{*}{FLOPs$\downarrow$(\%)} \\
        \cmidrule(lr){3-5}
        & & Base. & w/o FT Prun. & w/ FT Prun. \\
        \midrule
        \multirow{4}{*}{DeiT-Small}
        & VBP \cite{berisha2025variance} \scriptsize{(MLP)} & 79.70 & 64.44 & 78.62 & 32.2 & 30.4 \\
        & \textbf{IPPRO (ours)} \scriptsize{(MLP)} & 79.85 & \textbf{67.13} & \textbf{78.88} & \textbf{32.2} & \textbf{30.8} \\
        \cline{2-7}
        & SNP \cite{shim2024snp} \scriptsize{(MLP+Attn)} & 79.85 & 0.66 & 78.52 & 54.7 & \textbf{56.5} \\
        & \textbf{IPPRO (ours)} \scriptsize{(MLP+Attn)} & 79.85 & \textbf{23.17} & \textbf{79.13} & \textbf{55.4} & \textbf{56.5} \\
        \midrule
        \multirow{4}{*}{DeiT-Tiny}
        & VBP \cite{berisha2025variance} \scriptsize{(MLP)} & 72.20 & 39.58 & 70.61 & \textbf{28.0} & 25.2 \\
        & \textbf{IPPRO (ours)} \scriptsize{(MLP)} & 72.20 & \textbf{44.34} & \textbf{70.84} & 27.6 & \textbf{25.7} \\
        \cline{2-7}
        & SNP \cite{shim2024snp} \scriptsize{(MLP+Attn)} & 72.20 & 0.68 & 70.29 & \textbf{47.3} & \textbf{53.8} \\
        & \textbf{IPPRO (ours)} \scriptsize{(MLP+Attn)} & 72.20 & \textbf{21.43} & \textbf{71.67} & 43.3 & \textbf{53.8} \\
    \bottomrule
\end{tabular}
}
\end{table}

\subsection{Results Without Fine-tuning of LLMs}\label{appendix:wo_ft_llm}

To assess the impact of fine-tuning on LLMs pruning, we evaluated IPPRO without any fine-tuning and compared it against $L_1$-norm \cite{filters2016pruning} and Taylor \cite{molchanov2019importance} pruning methods using LLM-Pruner. We pruned the models to retain 80\% and 70\% of the parameters and reported the results on nine datasets, as shown in \cref{tab:llm_wo_finetune}. At the 20\% retention level, the performance of IPPRO was generally comparable to Taylor, with slight variations across some datasets. In contrast, at the 30\% retention level, IPPRO consistently outperformed all other methods across every dataset. These results indicate that, even without fine-tuning, IPPRO provides stable and reliable performance when applied to LLMs.

\begin{table}[H]
    \centering
    \caption{LLAMA-7b without finetune results}\label{tab:llm_wo_finetune}
    \resizebox{\linewidth}{!}{
        \begin{tabular}{l | l | cc | ccccccc |c | c}
        \toprule
        \makecell{Remain Param \\ Ratio} & Method & WikiText2 ($\downarrow$) & PTB ($\downarrow$) & BoolQ & PIQA & HellaSwag & WinoGrande & ARC-e & ARC-c & OBQA  & Avg & Drop($\downarrow$\%) \\
        \midrule
        \multirow{1}{*}{1.0}
        & Baseline &  12.62 & 22.14 & 73.1 & 78.3 & 72.9 & 66.8 & 67.3 & 41.4 & 42.4 & 63.17 & 0.0 \\
        \midrule
        \multirow{3}{*}{0.8}
        & LLM-Pruner (L1) & 236.23 & 446.55 & 50.52 & 57.89 & 40.42 & 51.46 & 35.86 & 27.99 & 27.80 & 41.70 & 21.47 \\
        & LLM-Pruner (Taylor) & \underline{19.77} & \underline{36.66} & \underline{59.39} & \textbf{75.57} & \textbf{65.34} & \underline{61.33} & \textbf{59.18} & \textbf{37.12} & \textbf{39.80} & \underline{56.82} & \underline{6.35} \\
        & IPPRO (ours) & \textbf{17.89} & \textbf{29.39} & \textbf{65.26} & \underline{74.54} & \underline{64.06} & \textbf{63.22} & \underline{58.26} & \underline{34.56} & \underline{38.20} & \textbf{56.87} & \textbf{6.3} \\
        \midrule
        \multirow{3}{*}{0.7}
        & LLM-Pruner (L1) & 294.00 & 446.55 & 41.47 & 56.58 & 35.62 & 50.91 & 32.11 & 25.51 & 29.80 & 38.86 & 24.31 \\
        & LLM-Pruner (Taylor) & \underline{32.85} & \underline{74.33} & \underline{62.17} & \underline{68.77} & \underline{59.49} & \underline{53.28} & \underline{45.83} & \underline{30.38} & \underline{35.40} & \underline{50.76} & \underline{12.41} \\
        & IPPRO (ours) & \textbf{18.89} & \textbf{31.63} & \textbf{65.26} & \textbf{73.75} & \textbf{61.67} & \textbf{62.12} & \textbf{54.24} & \textbf{33.79} & \textbf{36.60} & \textbf{55.35} & \textbf{7.82} \\
        \bottomrule
        \end{tabular}
    }
\end{table}

\subsection{Results Without Fine-tuning of DeepLabV3}\label{appendix:wo_ft_deep}

We report additional no fine-tuning pruning results on DeepLabV3-ResNet50 for completeness. Following the same setup as \cref{subsec:vit_performance}, we apply uniform structured pruning across layers and evaluate the pruned models without any fine-tuning.

\begin{figure}[H]
    \centering
    \begin{subfigure}[b]{0.46\textwidth}
        \includegraphics[width=0.9\textwidth, height=5cm]{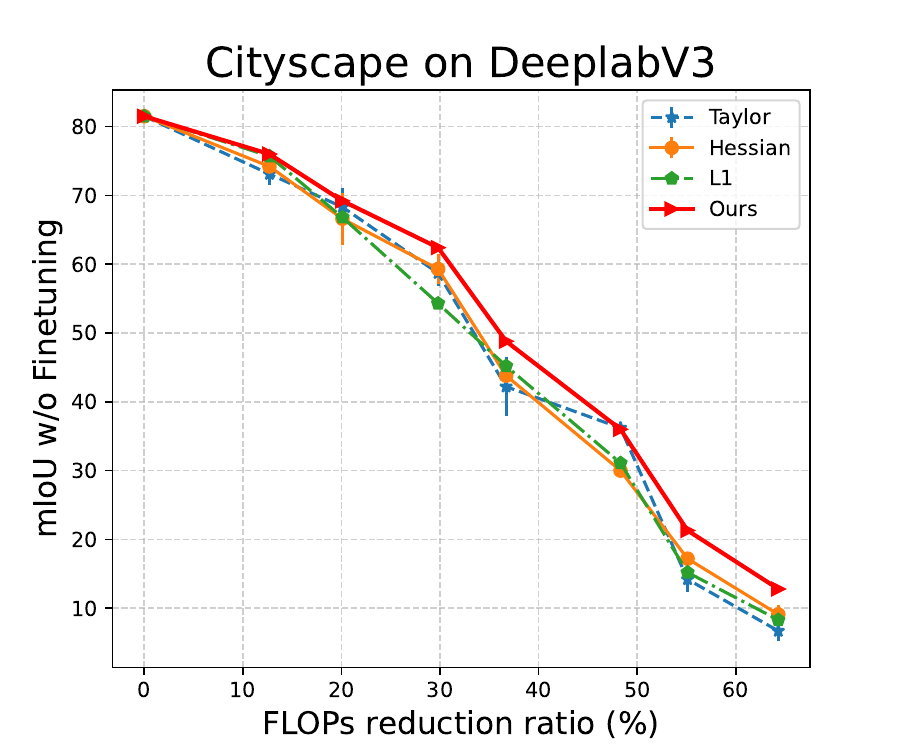}  
        \caption{Without Fine-tuning DeepLabV3-ResNet50 mIoU result}\label{fig:wo_ft_deeplab}
   \end{subfigure}
   \hfill
   \begin{subfigure}[b]{0.46\textwidth}
        \includegraphics[width=0.8\textwidth, height=5cm]{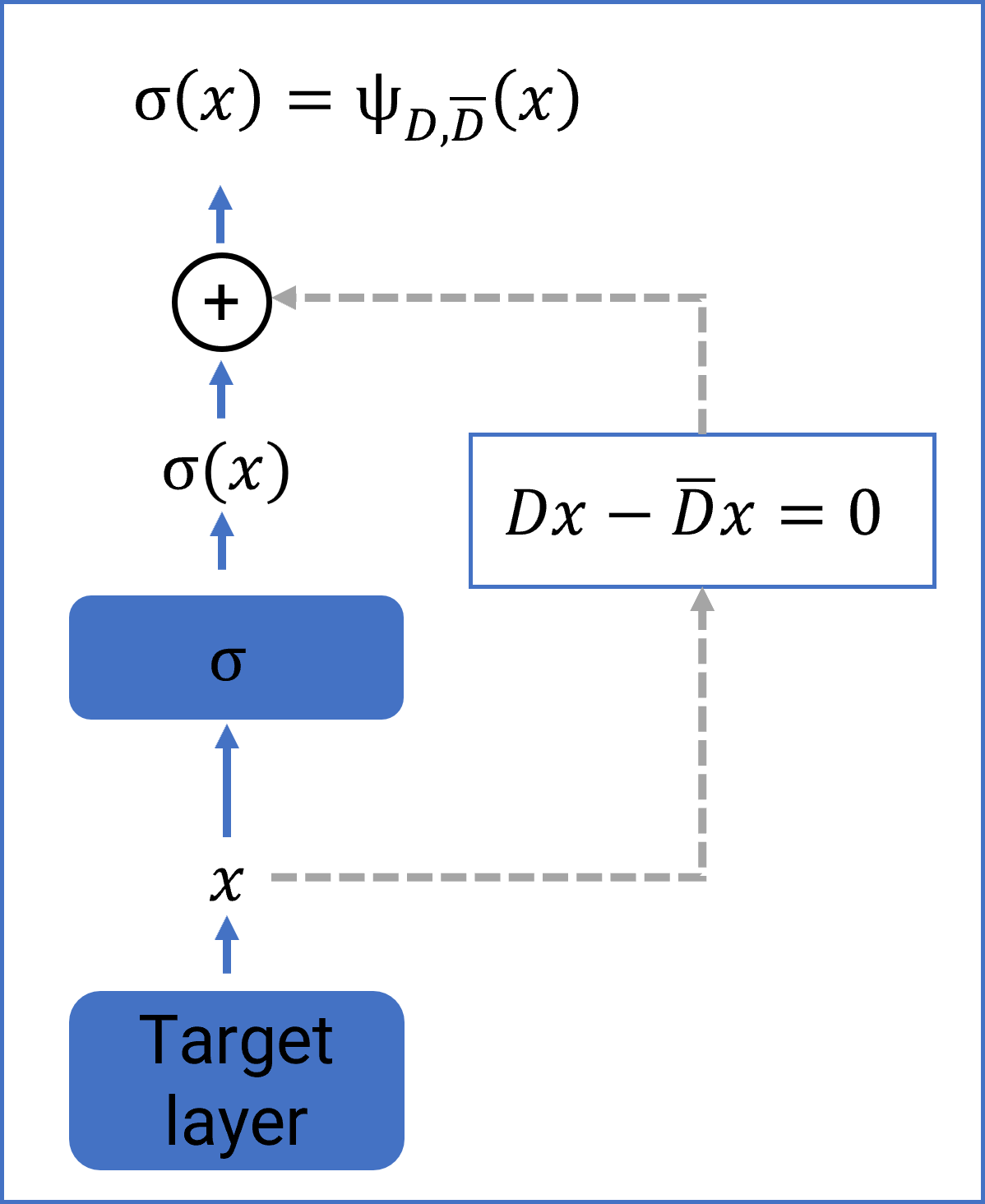}
        \caption{Illustration of the Parameter Injection Trick for PROscore Computation.}\label{fig:injection_trick}
   \end{subfigure}
\end{figure}

As shown in \cref{fig:wo_ft_deeplab}, IPPRO consistently outperforms L1-norm, Taylor, and Hessian-based pruning in terms of mIoU under increasing FLOPs reduction. In particular, IPPRO shows a slower performance degradation at higher pruning ratios, indicating improved robustness when no recovery through fine-tuning is allowed. These results further support that IPPRO captures functionally important channels even in no-finetuning settings.

\subsection{Stability Under Reparameterization}\label{appendix:stability}

\begin{wrapfigure}{r}{0.35\textwidth}
    \vspace{-5pt}
    \centering
    \includegraphics[width=0.25\textwidth]{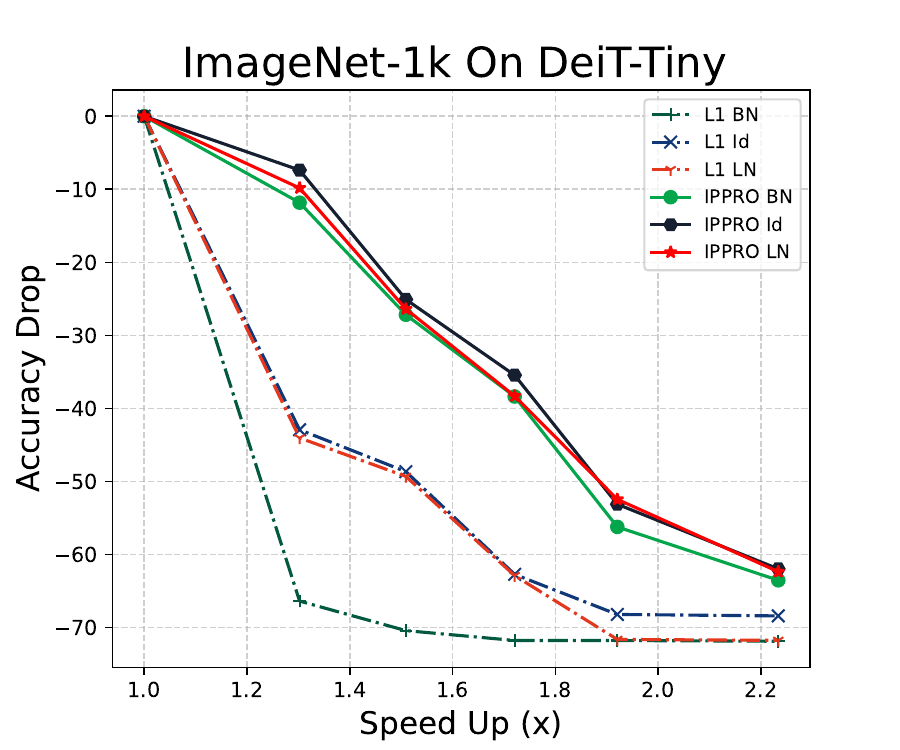}
    \caption{Reparameterization robustness}\label{fig:diff_norm_type}
    \vspace{-15pt}
\end{wrapfigure}

To evaluate robustness under realistic reparameterization, we compare magnitude-based pruning and IPPRO across three normalization settings (BatchNorm, LayerNorm, Identity) starting from the same pretrained weights. 
As shown in \cref{fig:diff_norm_type}, magnitude-based pruning exhibits large, inconsistent accuracy degradation across normalization variants---for example, at $\sim$40\% FLOPs reduction on DeiT-Tiny, the accuracy gap between the best and worst normalization schemes exceeds 4 percentage points. 
In contrast, IPPRO maintains a stable accuracy--speedup trade-off regardless of the normalization scheme, with inter-scheme variance consistently below 0.5 percentage points. 
This empirical stability complements the theoretical scale invariance established in \cref{subsec:method-importance_score} and aligns with the minimal correlation between PROscore and filter magnitude observed in \cref{appendix:importance_criteria}.

\section{Computational Overhead of Parameter Injection}\label{app:injection_trick_computational}

To evaluate the practical efficiency of the proposed method, we compare the architectural statistics and computational costs between the original models and those integrated with the parameter injection trick. As summarized in Table 15, the overhead introduced by IPPRO is negligible across various architectures, including Vision Transformers (DeiTs) and Large Language Models (LLaMA2-7B). 
\begin{itemize}
\item Parameter and FLOPs Efficiency: For DeiT variants, the injection trick increases the total parameter count by less than 0.55\%, as it only adds small scaling parameters proportional to the number of channels. In the case of LLaMA2-7B, the parameter increase is even more marginal, rising from 6.738B to only 6.739B. The MACs (FLOPs) remain identical because the additional element-wise operations are mathematically insignificant compared to the dense matrix multiplications within the blocks. 

\item Latency and Scoring Time: When measuring the actual time required to calculate PROscores using 10 calibration samples, LLaMA2-7B with the injection trick shows a latency of 103.4s, which is nearly identical to the original model's 102.1s and the Taylor-based method's 102.3s. The specific scoring time for IPPRO (2.40s) is comparable to the Taylor importance estimation (2.15s), demonstrating that our geometric approach provides superior scale-invariance without a substantial increase in wall-clock time. 

\end{itemize}

\begin{table}[h]
    \centering
    \caption{Comparison of architectural statistics between the original pretrained model and the model with parameter injection trick.}\label{tab:compare_injection_trick}
    \begin{subtable}[t]{0.49\linewidth}
        \centering
        \caption{DeiT Injection trick analysis}\label{tab:deit_injection}
        \resizebox{\linewidth}{!}{
        \begin{tabular}{lcccc}
            \toprule
            Model & Params(M) & MACs(G) & Latency (s) & Acc \\
            \midrule
            DeiT-Tiny & 5.72 & 1.26 & 15.89 & 72.20 \\
            + Injection Trick & 5.75 & 1.26 & 17.32 & 72.20 \\
            \midrule
            DeiT-Small & 22.05 & 4.61 & 39.63 & 79.70 \\
            + Injection Trick & 22.11 & 4.61 & 40.41 & 79.70 \\
            \midrule
            DeiT-Base & 86.57 & 17.58 & 40.75 & 81.73 \\
            + Injection Trick & 86.69 & 17.58 & 41.39 & 81.73 \\
            \bottomrule
            \end{tabular}
        }
    \end{subtable}
    \hfill
    \begin{subtable}[t]{0.49\linewidth}
    \centering
    \caption{LLaMA2-7B Injection trick analysis}\label{tab:llama_injection}
    \resizebox{\linewidth}{!}{
    \begin{tabular}{lcccc}
        \toprule
        Model & Params(B) & MACs(G) & Latency (s) & Scoring Time (s) \\
        \midrule
        LLaMA2-7B & 6.738 & 422.8 & 102.1 & - \\
        + Taylor & 6.738 & 422.8 & 102.3 & 2.15 \\
        \midrule
        LLaMA2-7B & 6.738 & 422.8 & 102.1 & - \\
        + Injection Trick & 6.739 & 422.8 & 103.4 & 2.40 \\
        \bottomrule
        \end{tabular}
    }
    \end{subtable}
\end{table}

\section{Schematic Analysis of Parameter Injection}\label{app:injection_trick}

\begin{itemize}
    \item Forward Invariance: At initialization, we set $D = \overline{D}$. As shown in \cref{fig:injection_trick}, causing $Dx$ and $-\overline{D}x$ to perfectly cancel out. Consequently, the forward pass remains mathematically identical to the original model, preserving all pretrained features. (\cref{tab:compare_injection_trick})
    \item Transient Nature: These parameters are used solely to ``tap into" the gradient flow. Once PROscores are accumulated, the injected terms are removed, restoring the original architecture before pruning.
\end{itemize}

\section{Sensitivity Analysis of $\lambda$ value}\label{lamda}

The absolute value of PROscore is affected by the hyperparameter $\lambda$, which controls the step size of the gradient movement in real projective space $\mathbb{RP}^N$, since the angle $\theta(p_i')$ is related to the length $\overline{p_ip_i'}$. In extreme case, if $\lambda\to 0$ then $\theta(p_i')\to\frac{\pi}{4}$ and thus the PROscores would distribute close to one.

However, the pruning decision are invariant under selection of isotropic scaling on $\lambda>0$, since we are choosing filters with respect to their relative order which are invariant: if the $p_i''$ is new point under gradient movement with different step size $\lambda$, then following holds: if $\theta(p_i')<\theta(p_j')$ then $\theta(p_i'')<\theta(p_j'')$.

\begin{figure}[H]
    \centering
    \begin{subfigure}[b]{0.46\textwidth}
        \includegraphics[width=\textwidth]{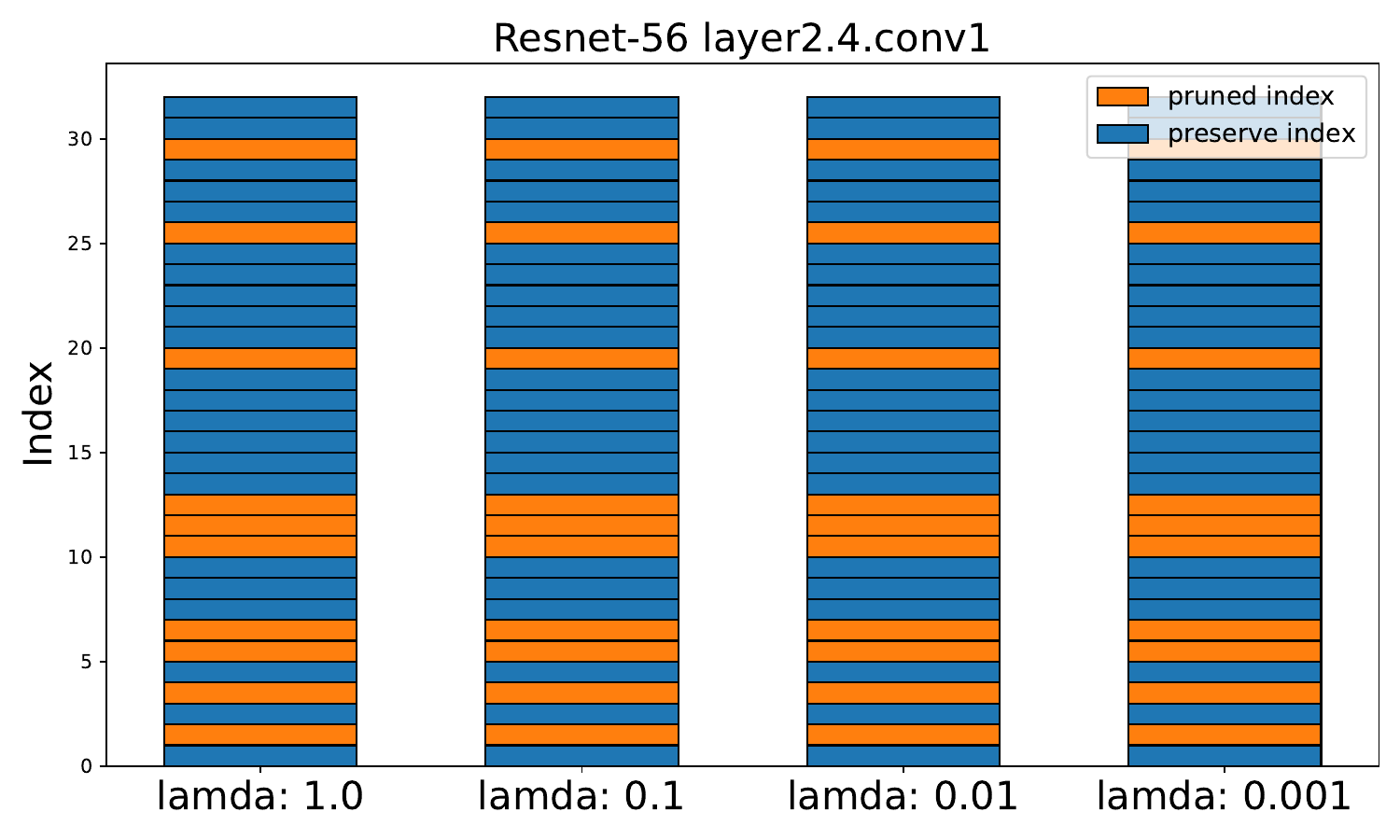}  
   \end{subfigure}
    \begin{subfigure}[b]{0.46\textwidth}
        \includegraphics[width=\textwidth]{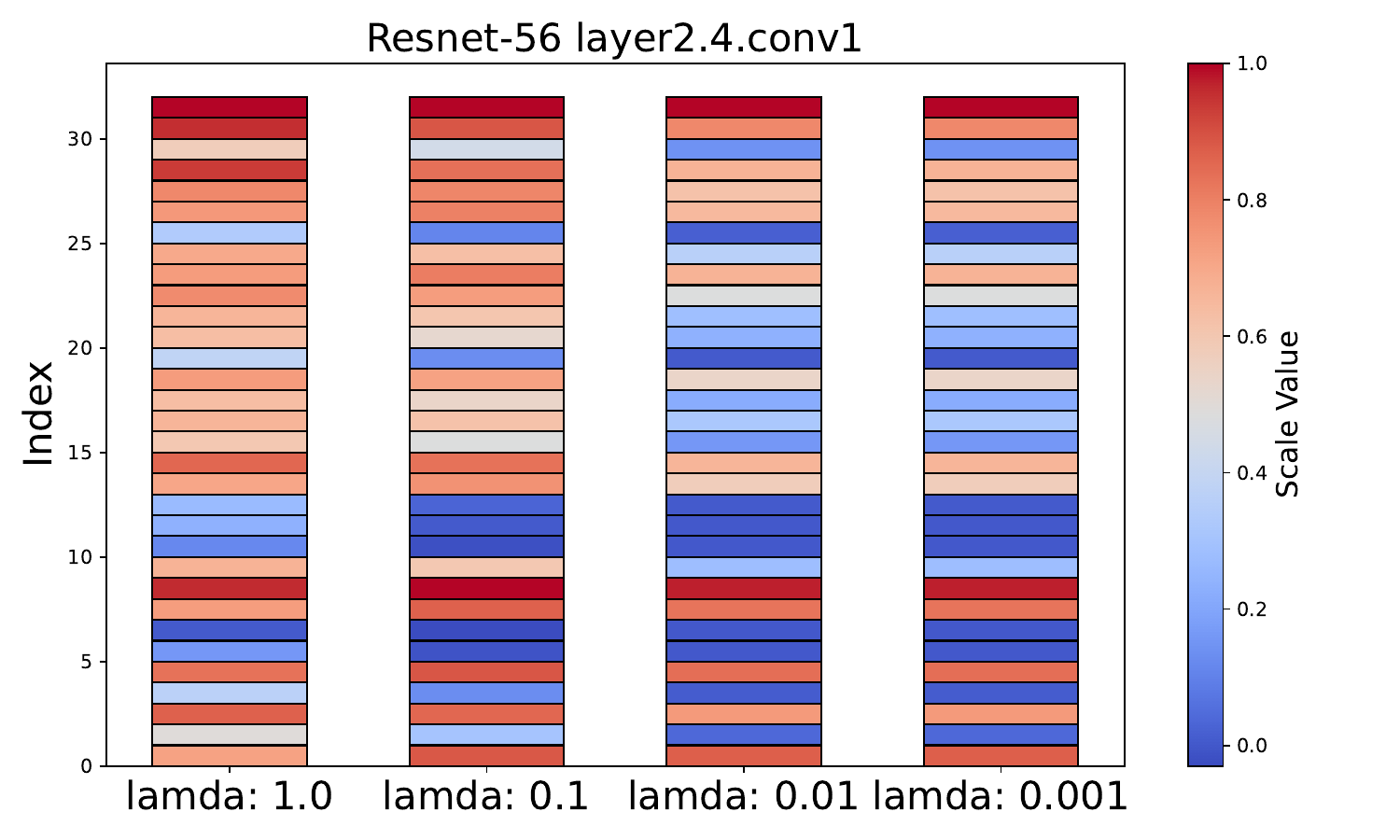}
    \end{subfigure}
    \begin{subfigure}[b]{0.46\textwidth}
        \includegraphics[width=\textwidth]{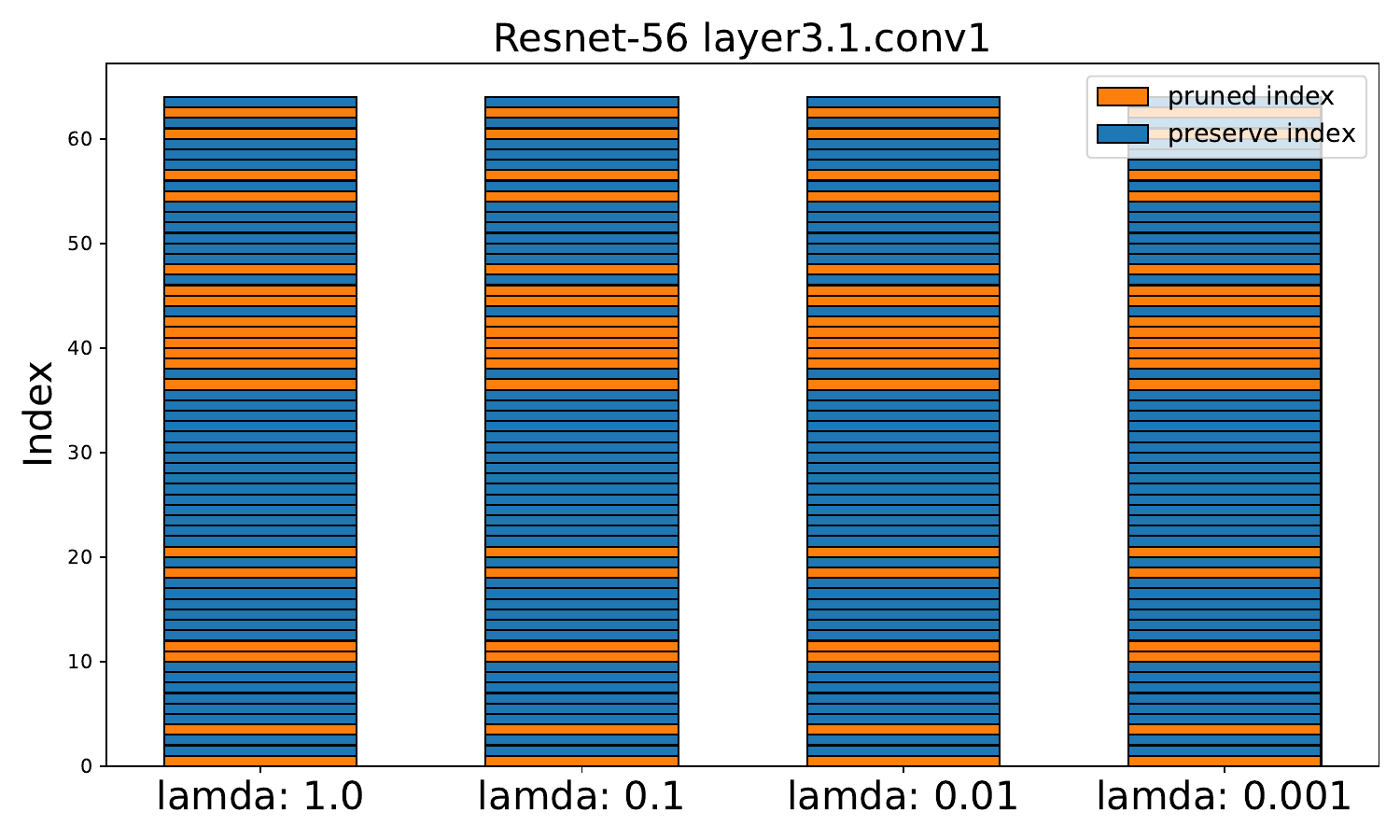}\caption{Preserve Pruning Index of various $\lambda$ value}\label{fig:lambda_preserve}
   \end{subfigure}
    \begin{subfigure}[b]{0.46\textwidth}
        \includegraphics[width=\textwidth]{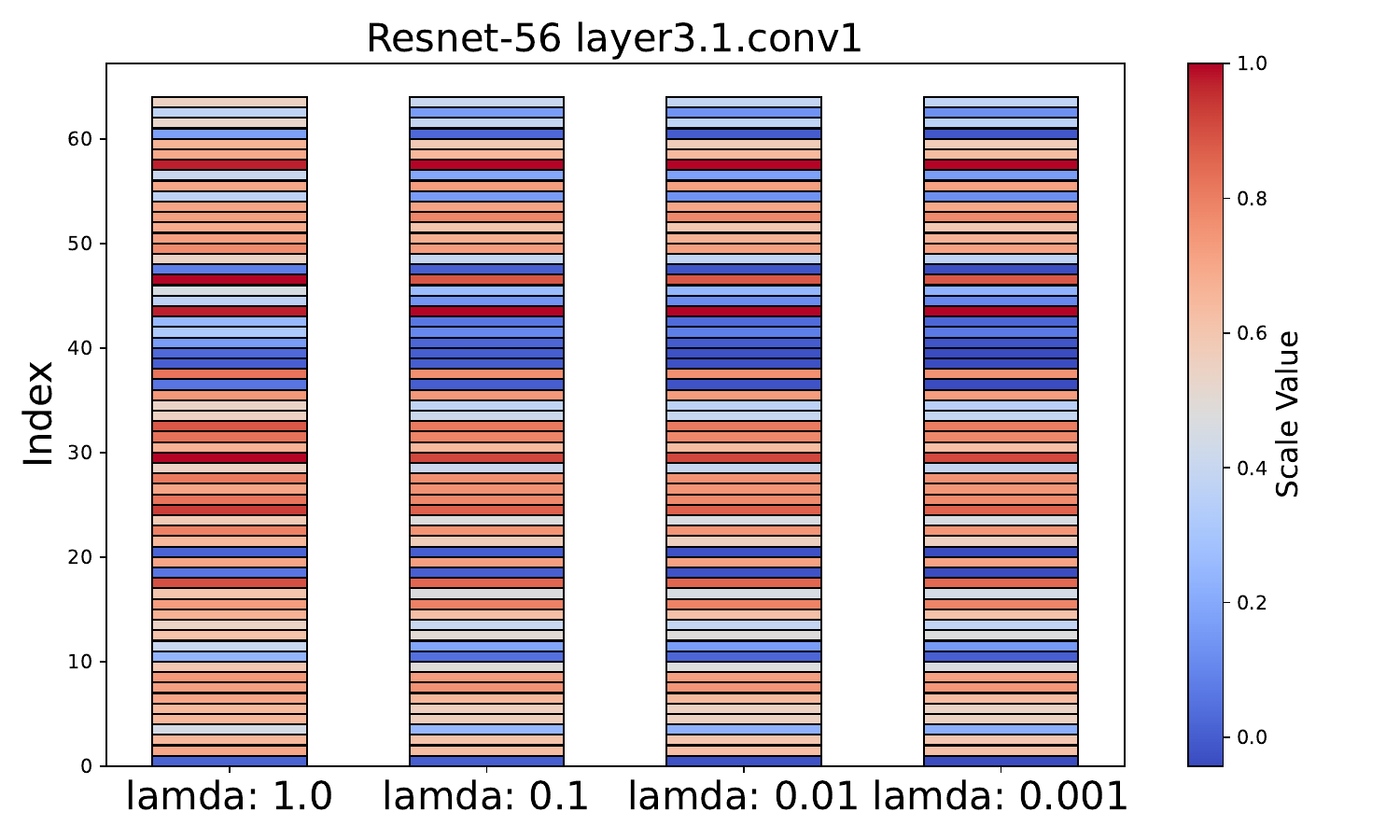}\caption{PROscore scale of different $\lambda$ value}\label{fig:lambda_scale}
    \end{subfigure}
    \caption{Preserve pruning index of $\lambda$ changes}
\end{figure}

To validate this, we conduct experiments on Resnet56 model with $\lambda$ set to 1, 0.1, 0.01, and 0.001, examining both the preservation of pruning indices and the consistency of PROscore scales across layers.
As shown in \cref{fig:lambda_preserve}, the pruning indices remained stable regardless of the $\lambda$ value, and \cref{fig:lambda_scale} shows that the relative importance measured by PROscore is invariant under $\lambda$ selection.

\section{Qualitative Results of DeepLabV3}
\begin{figure*}[!th]
    \begin{subfigure}{\linewidth}
        \includegraphics[width=\textwidth, height=2cm]{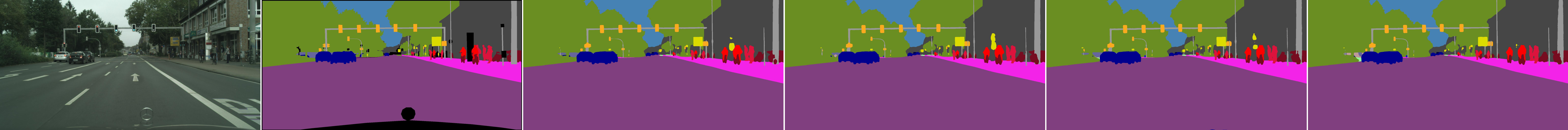}
    \end{subfigure}
    \begin{subfigure}{\linewidth}
        \includegraphics[width=\textwidth, height=2cm]{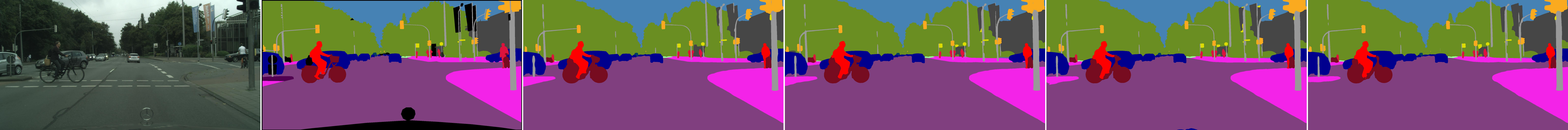}
    \end{subfigure}
    \begin{subfigure}{\linewidth}
        \includegraphics[width=\textwidth, height=2cm]{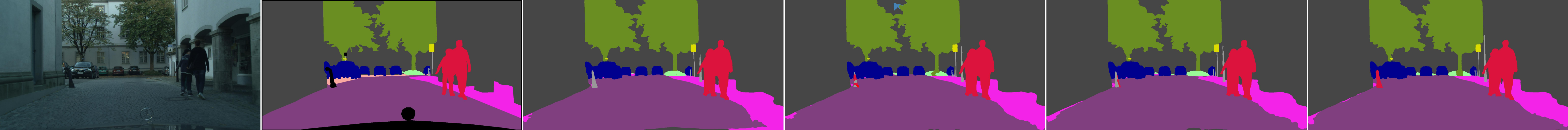}
    \end{subfigure}
    \begin{subfigure}{\linewidth}
        \includegraphics[width=\textwidth, height=2cm]{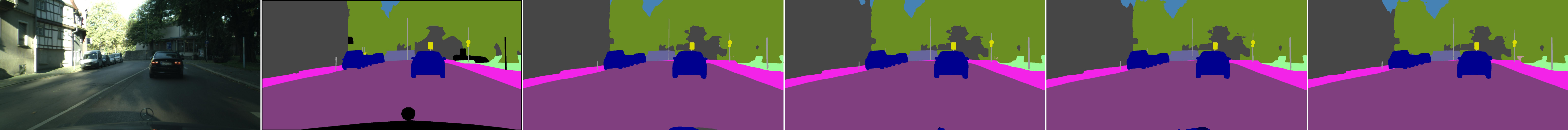}
    \end{subfigure}
    \begin{subfigure}{\linewidth}
        \includegraphics[width=\textwidth, height=2cm]{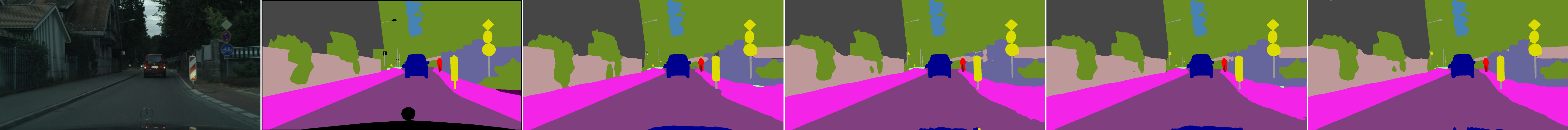}
    \end{subfigure}
    \begin{subfigure}{\linewidth}
        \includegraphics[width=\textwidth, height=2cm]{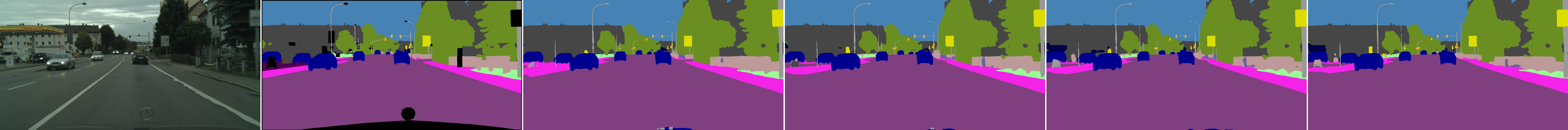}
    \end{subfigure}
    \begin{subfigure}{\linewidth}
        \includegraphics[width=\textwidth, height=2cm]{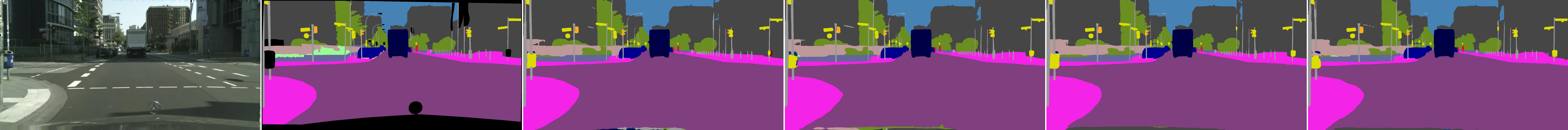}
    \end{subfigure}
    \begin{subfigure}{\linewidth}
        \includegraphics[width=\textwidth, height=2cm]{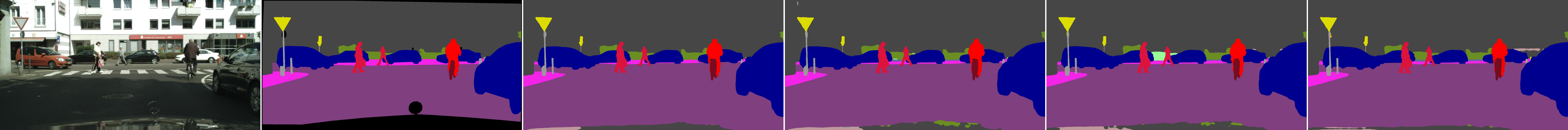}
    \end{subfigure}
    \begin{subfigure}{\linewidth}
        \includegraphics[width=\textwidth, height=2cm]{plots/appendix_seg2/seg_img1.pdf}
    \end{subfigure}
  \caption{Visualization Cityscapes dataset Our Pruning results of using DeepLabV3-ResNet50 models}\label{fig:seg_results}
\end{figure*}

In this section, we present qualitative results of the segmentation task, as shown in \cref{fig:seg_results}. 
We compare the original image, ground truth segmentation, segmentation result from the unpruned model, and the segmentation result after pruning using our PROscore.

\section{Comparison of Importance Criteria}\label{appendix:importance_criteria}

\begin{figure}[!ht]
    \centering
    \begin{subfigure}{\linewidth}
        \includegraphics[width=\textwidth, height=2.5cm]{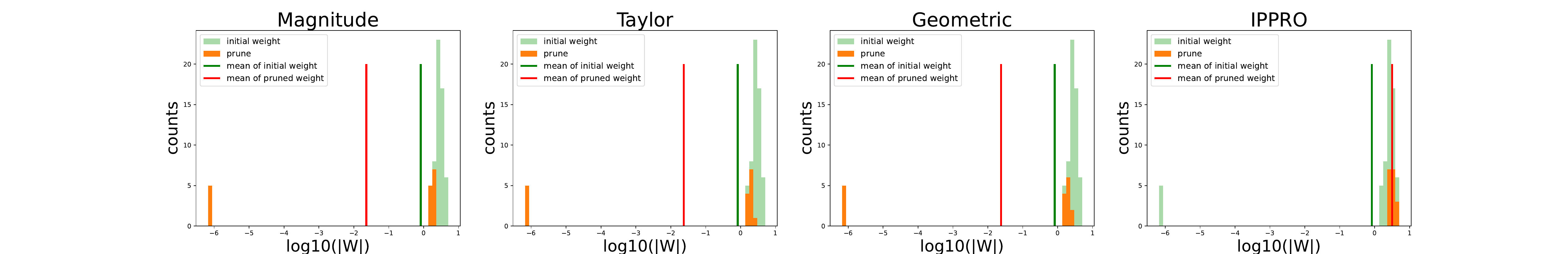}
        \caption{ResNet50 - layer1.2.conv2}
    \end{subfigure}
    \begin{subfigure}[b]{\textwidth}
        \centering
        \includegraphics[width=\textwidth, height=2.5cm]{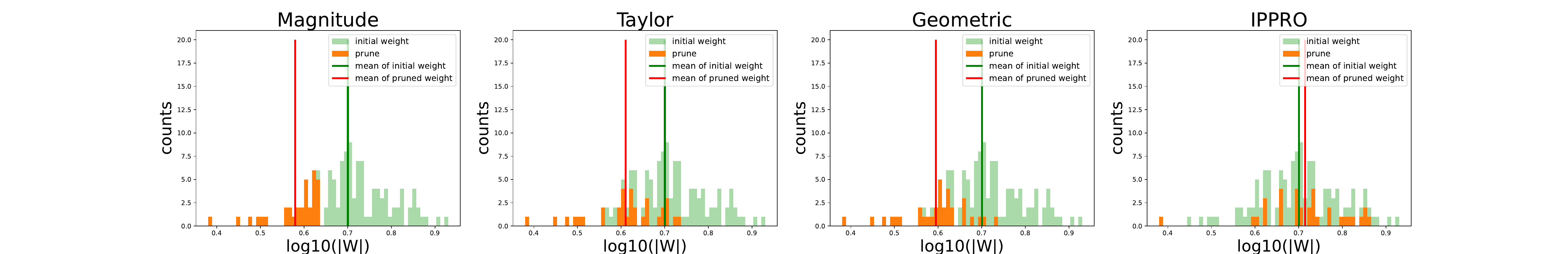}
        \caption{ResNet50 - layer2.1.conv2}
    \end{subfigure}
    \begin{subfigure}[b]{\textwidth}
        \centering
        \includegraphics[width=\textwidth, height=2.5cm]{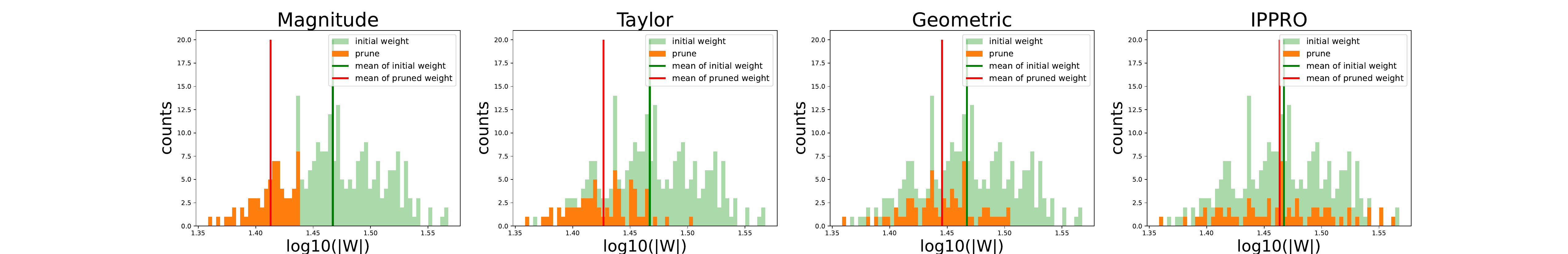}
        \caption{ResNet50 - layer3.0.conv2}
    \end{subfigure}
  \caption{Visualization of magnitude of pruning filters obtained by four different criteria, on DeepLabV3-Resnet50 with Cityscapes dataset. }\label{fig:pruning_index}
\end{figure}

We compare IPPRO with other magnitude- and gradient-based pruning methods such as $L_1$-norm \cite{he2017channel}, Taylor \cite{molchanov2019importance}, and geometric \cite{he2019filter}. For fairness, we compute gradients across the entire dataset, even for methods like Taylor that support randomized subsets. As shown in \cref{fig:pruning_index}, traditional methods heavily depend on pre-trained weights and produce highly similar pruning patterns. In contrast, IPPRO demonstrates reduced reliance on initial weights, yielding distinct importance distributions. Histogram visualizations of ResNet-50 layers further confirm that IPPRO prunes differently, supporting its robustness against weight initialization bias.

\section{Sensitivity Analysis}\label{appendix:sampling_analysis}

We provide full sampling sensitivity results for LLM pruning to supplement the summarized results in \cref{tab:data_sensitivity} of the main paper. Specifically, this section reports per-task performance across different calibration sizes for LLAMA-7B, whereas the main paper focuses on averaged results for clarity. The detailed results are described in \cref{tab:app_fullllm_data_sensitivity}. 

\begin{table}[H]
    \small
    \centering
    \caption{LLM Pruning results on different calibration size}\label{tab:app_fullllm_data_sensitivity}
    \resizebox{\linewidth}{!}{
        \begin{tabular}{llcccccccc}
        \toprule
        \multirow{2.5}{*}{\makecell{Calibration Set \\ Num samples}} & \multirow{2.5}{*}{Method} & \multicolumn{7}{c}{Acc($\uparrow$\%)} & \multirow{2.5}{*}{Avg($\uparrow$\%)} \\
         \cmidrule(lr){3-9}
        & & BoolQ & PIQA & HellaSwag & WinoGrande & ARC-e & ARC-c & OBQA \\
        \midrule
        \multirow{1}{*}{NA}
        & Baseline & 73.1 & 78.3 & 72.9 & 66.8 & 67.3 & 41.4 & 42.4 & 63.17 \\
        \midrule
        \multirow{2}{*}{10}
        & LLM-Pruner & 69.5 & \underline{76.4} & \textbf{68.1} & \underline{65.1} & 63.4 & \underline{37.9} & \underline{40.0} & 60.06 \\
        & IPPRO (ours) & \underline{72.8} & \textbf{76.8} & \underline{67.5} & \underline{65.1} & \underline{67.2} & 37.6 & 39.6 & \underline{60.94} \\
        \midrule
        \multirow{1}{*}{100}
        & IPPRO (ours) & \textbf{72.9} & 75.9 & 66.7 & \textbf{66.3} & 65.6 & 37.7 & 39.0 & 60.59 \\
        \midrule
        \multirow{1}{*}{1000}
        & IPPRO (ours) & 72.1 & \textbf{76.8} & 67.1 & 62.8 & \textbf{68.5} & \textbf{38.8} & \textbf{40.8} & \textbf{60.98} \\
        \bottomrule
        \end{tabular}
    }
\end{table}

As shown in \cref{tab:sampling_results}, we conduct experiments on ResNet-56 with CIFAR-10 and VGG-19 with CIFAR-100, varying the number of samples used to 100\%, 50\%, 25\%, and 5\% of the dataset. A balanced label sampler is employed to preserve label distribution across mini-batches. To account for randomness introduced by subset sampling, each configuration is repeated five times, and both the mean (Mu) and the maximum accuracy are reported.

\begin{table}[H]
    \centering
    \caption{Experimental results on different dataset size}\label{tab:sampling_results}
    \begin{subtable}[t]{0.48\linewidth}
    \centering
    \caption{CIFAR10 dataset on Resnet-56 (FLOPs reduction 71.85\%)}\label{tab:sampling_results_cifar10}
    \resizebox{\linewidth}{!}{
    \begin{tabular}{llccc}
    \toprule
        \multirow{2.5}{*}{\makecell{Dataset \\ size}} & \multirow{2.5}{*}{\makecell{Time \\ usage (s)}}  & \multicolumn{3}{c}{Top-1 Acc (\%)}\\ 
        \cmidrule(lr){3-5}
        & & Base. & Prun. Mu (Max) & $\Delta$ Mu (Max)\\
        \midrule
        Full & 45.4 & 93.53 & 92.47 & -1.06 \\  
        50\% & 25.4 & 93.53 & 92.18 (92.44) & -1.35 (-1.09)\\
        25\% & 16.6 & 93.53 & 92.27 (92.46) & -1.26 (-1.07)\\
        5\% & 7.2 & 93.53 & 92.21 (92.43) & -1.32  (-1.10)\\   
    \bottomrule
    \end{tabular}
    }
\end{subtable}
    \hfill
    \begin{subtable}[t]{0.48\linewidth}
    \centering
    \caption{CIFAR100 dataset on VGG19 (FLOPs reduction 87.5\%)}\label{tab:sampling_results_cifar100}
    \resizebox{\linewidth}{!}{
    \begin{tabular}{llccc}
    \toprule
        \multirow{2.5}{*}{\makecell{Dataset \\ size}} & \multirow{2.5}{*}{\makecell{Time \\ usage (s)}}  & \multicolumn{3}{c}{Top-1 Acc (\%)}\\ 
        \cmidrule(lr){3-5}
        & & Base. & Prun. Mu (Max) & $\Delta$ Mu (Max)\\
        \midrule
        Full & 13.2 & 73.5 & 70.47 & -3.03 \\  
        50\% &  6.9 & 73.5 & 69.66(70.09) & -3.84(-3.41)\\
        25\% & 3.5 & 73.5 & 69.55(70.03) & -3.95(-3.47)\\
        5\% & 0.7 & 73.5 & 69.58(70.24) & -3.92(-3.26)\\   
    \bottomrule
    \end{tabular}
    }
\end{subtable}

\end{table}

\newpage

\end{document}